\documentclass[twoside]{article}

\usepackage[accepted]{aistats2026}
\usepackage[utf8]{inputenc} 
\usepackage[
  style=authoryear,
  backend=biber,
  maxcitenames=2,   
  mincitenames=1,   
  uniquename=false, 
  uniquelist=false  
]{biblatex}

\DeclareDelimFormat[textcite]{multinamedelim}{\addcomma\space}

\usepackage[T1]{fontenc}    
\usepackage{hyperref}       
\usepackage{url}            
\usepackage{booktabs}       
\usepackage{amsfonts}       
\usepackage{nicefrac}       
\usepackage{microtype}      
\usepackage{xcolor}         
\usepackage{stfloats} 

\usepackage{cleveref}
\usepackage{graphicx}
\usepackage{subcaption}

\usepackage{algorithm}
\usepackage{algorithmic}

\newcommand{\md}{\mathrm{d}}
\newcommand{\Pac}{\mathcal{P}_{2,ac}(\mathcal{X})}
\newcommand{\WG}{\nabla \hspace{-0.22cm} \nabla}

\usepackage[textwidth=1.8cm, textsize=scriptsize]{todonotes}

\usepackage[acronym]{glossaries-extra}
\setabbreviationstyle[acronym]{long-short}

\newacronym{LVM}{{{\textsc{\small LVM}}}}{latent variable model}
\newacronym{ML}{{{\textsc{\small ML}}}}{maximum likelihood}
\newacronym{MMLE}{{{\textsc{\small MMLE}}}}{maximum marginal likelihood estimation}
\newacronym{MCMC}{{{\textsc{\small MCMC}}}}{Markov chain Monte Carlo}
\newacronym{LMC}{{{\textsc{\small LMC}}}}{Langevin Monte Carlo}
\newacronym{EM}{{{\textsc{\small EM}}}}{Expectation-Maximisation}
\newacronym{SGD}{{{\textsc{\small SGD}}}}{stochastic gradient ascent}
\newacronym{ULA}{{{\textsc{\small ULA}}}}{unadjusted Langevin algorithm}
\newacronym{IS}{{{\textsc{\small IS}}}}{importance sampling}
\newacronym{MSE}{{{\textsc{\small MSE}}}}{mean-squared error}
\newacronym{SVGD}{{{\textsc{\small SVGD}}}}{Stein variational gradient descent}
\newacronym{SVGD-EM}{{{\textsc{\small SVGD-EM}}}}{Stein variational gradient descent-expectation maximisation}
\newacronym{M-SVGD-EM}{{{\textsc{\small M-SVGD-EM}}}}{Momentum SVGD-EM}
\newacronym{PGD}{{{\textsc{\small PGD}}}}{Particle Gradient Descent}
\newacronym{MPGD}{{{\textsc{\small MPGD}}}}{Momentum Particle Gradient Descent}
\newacronym{SVGD-WNes}{{{\textsc{\small SVGD-WNes}}}}{Wasserstein-Nesterov Stein variational gradient descent}
\newacronym{SVGD-RAGD}{{{\textsc{\small SVGD-RAGD}}}}{Riemann-Nesterov Stein variational gradient descent}
\newacronym{SOUL}{{{\textsc{\small SOUL}}}}{Stochastic Optimization via Unadjusted Langevin algorithm}
\newacronym{RAGD}{{{\textsc{\small RAGD}}}}{Riemannian version of Nesterov’s Accelerated Gradient Descent}

\makeatletter
\newcommand{\blindfootnote}[1]{%
  \begingroup
  \renewcommand\thefootnote{\textsuperscript{$\star$}}%
  \renewcommand\@makefntext[1]{%
    \noindent\textsuperscript{$\star$}\,##1%
  }%
  \footnotetext{#1}%
  \endgroup
}
\makeatother

\begin{document}

%

%

\twocolumn[

\aistatstitle{{Momentum SVGD-EM for Accelerated Maximum Marginal Likelihood Estimation}}

\aistatsauthor{Adam Rozzio$^{\star,1}$ \And Rafael Athanasiades$^{\star,2}$ \And O. Deniz Akyildiz$^{2}$}
\runningauthor{Adam Rozzio, Rafael Athanasiades, O. Deniz Akyildiz}

\aistatsaddress{$^{1}$Department of Mathematics, Ecole Normale Supérieure Paris-Saclay, France \\ $^{2}$Department of Mathematics, Imperial College London, UK}
]

\blindfootnote{Equal contribution.} 

\begin{abstract}
Maximum marginal likelihood estimation (MMLE) can be formulated as the optimization of a free energy functional. From this viewpoint, the Expectation–Maximisation (EM) algorithm admits a natural interpretation as a coordinate descent method over the joint space of model parameters and probability measures. Recently, a significant body of work has adopted this perspective, leading to interacting particle algorithms for MMLE. In this paper, we propose an accelerated version of one such procedure, based on Stein variational gradient descent (SVGD), by introducing Nesterov acceleration in both the parameter updates and in the space of probability measures. The resulting method, termed Momentum SVGD-EM, consistently accelerates convergence in terms of required iterations across various tasks of increasing difficulty, demonstrating effectiveness in both low- and high-dimensional settings.
\end{abstract}

\section{Introduction}
\Glspl*{LVM} are ubiquitous in machine learning and statistics, as they provide a powerful framework for modeling complex data by capturing the underlying latent structure \parencite{whiteley2022statistical}. A typical \gls*{LVM} is defined by a joint probability density $p_\theta(x,y): \mathcal{X} \times \Theta \to \mathbb{R}$, where $y$ is the (fixed) observed data, $x$ is the unobserved latent variable, and $\theta \in \Theta$ is the model's parameter. Given some data $y$, the typical goal is to find the parameter that maximizes the marginal likelihood $p_\theta(y)$, which is the probability of observing the data under the model - termed the \gls*{MMLE}. This results in the following optimization problem:
\begin{equation}
    \label{eq:Marginal Likelihood}
     \theta_\star \in \arg \max_{\theta \in \Theta} \log p_{\theta} (y),
\end{equation}
where $p_\theta(y) := \int_{\mathcal{X}} p_\theta(x,y)\mathrm{d}x$. The standard approach for performing \gls*{MMLE} is to use the celebrated \gls*{EM} algorithm \parencite{dempster1977maximum}. At iteration $t$, the \gls*{EM} algorithm alternates between two steps: the E-step, which computes the expectation $Q(\theta, \theta_t) := \mathbb{E}_{p_{\theta_t}(x|y)}[\log p_\theta(x, y)]$, and the M-step, which updates the model parameters by maximising the expected log-likelihood, i.e. $\theta_{t+1} \in \arg \max_{\theta \in \Theta} Q(\theta, \theta_t)$. The \gls*{EM} algorithm is typically intractable (no closed form) to implement, resulting in approximations for its E and M steps, see, e.g. \textcite{wei1990monte,meng1993maximum,liu1994ecme,lange1995gradient}. In recent years, diffusion based approaches gained traction, in particular, algorithms based on \gls*{ULA} have been proposed, see, e.g., \textcite{de2021efficient}. These approaches are computationally expensive, as they require \gls*{MCMC} chains for the E-step, which can be slow to converge and hard to understand theoretically.

An emerging computational paradigm for \gls*{MMLE} is to use the coordinate descent perspective of \gls*{EM} to recast the problem as a minimisation of the so-called free energy \parencite{neal1998view}. In particular, \textcite{neal1998view} showed that the \gls*{EM} algorithm can be interpreted as a coordinate descent scheme on the free energy functional:
\begin{align}\label{eq: Free_energy}
    \mathcal{F}(\theta,q)\hspace{-0.05cm} &:= \hspace{-0.15cm} \int_{\mathcal{X}} \hspace{-0.15cm}   q(x)\log  q(x)\mathrm{d}x \hspace{-0.05cm} - \hspace{-0.15cm}\int_{\mathcal{X}} \hspace{-0.15cm}  q(x)\log p_\theta (x,y) \mathrm{d}x.
\end{align}
Based on this perspective, \textcite{kuntz2023particle} proposed \gls*{PGD}, a particle-based algorithm for \gls*{MMLE} that uses the free energy as an objective function. The \gls*{PGD} runs an optimiser for $\theta$-updates and a particle system for $q$-updates. This is a significant departure from the classical algorithms for \gls*{EM} as the particle-based approach allows for simultaneous updates of the latent variables and the model parameters. This approach was justified theoretically \parencite{caprio2024error} and led to a number of follow-up works, see, e.g., \textcite{lim2023momentum,sharrock2024tuning,akyildiz2025interacting,oliva2024kinetic,encinar2025proximal}.

Among the works that followed \textcite{kuntz2023particle}, there are two particular works that are of special interest to us in this work. In the first work, \textcite{sharrock2024tuning} proposed the \gls*{SVGD-EM} algorithm, which uses the \gls*{SVGD} \parencite{liu2016stein} to evolve the particles for the latent variables. This contrasts with the approach followed in \textcite{kuntz2023particle}, where the particles are evolved using (independent) Langevin dynamics. The second work is \textcite{lim2023momentum}, which proposed a momentum-based approach to \gls*{PGD} called \gls*{MPGD}. The \gls*{MPGD} algorithm uses an alternative (enriched) free energy to develop a momentum-based algorithm for \gls*{MMLE}. \textcite{lim2023momentum} showed that the \gls*{MPGD} algorithm converges faster than the \gls*{PGD} method on a variety of settings, demonstrating the acceleration provided by the momentum-based approach. In this work, we combine these two approaches. In particular, our contributions are summarised as follows.

\noindent \textbf{Contributions.} In this work, we provide an accelerated version of the \gls*{SVGD-EM} algorithm, which we term \gls*{M-SVGD-EM}. Our approach integrates two distinct Nesterov-inspired acceleration schemes: one applied to the sampling phase using the \gls*{SVGD-WNes} as introduced in \textcite{liu2019understanding}, and the other applied to the parameter updates following classical momentum techniques \parencite{nesterov1983method}. The resulting method bears similarities to \gls*{MPGD} and can be seen as an analogue of \gls*{MPGD} for the \gls*{SVGD-EM} algorithm. We derive the \gls*{M-SVGD-EM} algorithm that combines the \gls*{SVGD-WNes} with Nesterov momentum for the parameter updates, which is achieved by leveraging the free energy perspective of \gls*{MMLE} and the connection between \gls*{SVGD-EM} and Wasserstein gradient flows. To demonstrate the effectiveness of our approach, we conduct numerical experiments on a variety of tasks, including a Toy Hierarchical Model, a Bayesian Logistic Regression task on the \textit{Wisconsin Breast Cancer Dataset}, and a Bayesian Neural Network model applied to the MNIST dataset. Our results show that \gls*{M-SVGD-EM} consistently outperforms \gls*{SVGD-EM} in terms of convergence speed across all tasks, hence providing a fast and efficient method for \gls*{MMLE}.

The paper is organized as follows. In Section~\ref{sec:background}, we recall our problem setup and the \gls*{SVGD-EM} algorithm briefly. In Section~\ref{sec:M-SVGD-EM}, we introduce our method, \gls*{M-SVGD-EM}. In Section~\ref{section: Numerical experiments}, we provide a thorough demonstration of acceleration properties of \gls*{M-SVGD-EM}, showing that the new method consistently accelerates the \gls*{SVGD-EM}. Finally, we conclude with Section~\ref{sec:conclusion}.

\noindent\textbf{Notation.} We denote the latent space, the observation space, and the parameter space with $\mathcal{X} :=\mathbb{R}^{d_x}$, $\mathcal{Y}:= \mathbb{R}^{d_y}$, and $\Theta := \mathbb{R}^{d_\theta}$ respectively. For any finite measure $q$ on $\mathcal{X}$ we note $T_{\#} q$ the pushforward measure of $q$ by the measurable map $T:\mathcal{X} \to \mathcal{X}$. The map denoted by $\text{id}:\mathcal{X} \to \mathcal{X}$ refers to the identity map. We note $\Pac$ the set of absolutely continuous measures w.r.t the Lebesgue measure with finite variance and for $q_1, q_2 \in \Pac$, the Wasserstein-2 distance \parencite[Chapter 7.1]{ambrosio2008gradient} between $q_1$ and $q_2$ is defined by $W_2^2(q_1, q_2) = \inf_{\gamma \in \Gamma(q_1, q_2)} \int_{\mathcal{X} \times \mathcal{X}} \| x_1 - x_2 \|^2 \, \gamma(\md x_1, \md x_2)$
where $\Gamma(q_1, q_2)$ is the set of couplings between $q_1$ and $q_2$. We also note $\mathcal{T}_{q_1}\Pac$ the tangent space at $q_1$ and we refer to the metric space \(( \Pac, W_2) \) as the Wasserstein space.  Let $\mathcal{F}: \Theta\times\Pac \mapsto \mathbb{R}\cup \{\infty\}$, we note $\frac{\delta \mathcal{F}}{\delta q}(\theta,q) : \mathcal{X} \to \mathbb{R}$ the first variation of $\mathcal{F}$ w.r.t $q$ defined \parencite[Definition 7.12]{santambrogio2015optimal} as the unique function (up to an additive constant) such that $\frac{\md}{\md t} \mathcal{F}(\theta, q+tm)|_{t=0} = \int_\mathcal{X} \frac{\delta \mathcal{F}}{\delta q}(\theta,q)(x)\md m(x)$ for any $m : \mathcal{X} \to \mathbb{R}$ satisfying $\int_\mathcal{X}m(x)\md x = 0$. The only Reproducing Kernel Hilbert Space (RKHS) associated with the real positive definite kernel $k(\cdot,\cdot):\mathcal{X} \times \mathcal{X} \to \mathbb{R}$ is noted $\mathcal{H}$ and we note $\nabla_1 k$ the gradient of $k(\cdot, \cdot)$ with respect to its first component. In the following, vector-valued RKHS will be spaces of the form $\mathcal{H}^d := \mathcal{H}\times \dots \times \mathcal{H}$.

\section{Background}\label{sec:background}
Let $p_\theta(x, y)$ be a joint probability distribution differentiable in $\theta$ and $x$ of an \gls*{LVM} where $y \in \mathcal{Y}$ is a fixed observation (dataset), $x\in\mathcal{X}$ denotes latent variables, $\theta \in \Theta$ is the parameter. In this paper, we are interested in fitting $\theta$ given the observed data $y$ by solving the problem
\begin{align}\label{eq:mmle}
\theta_\star \in \arg \max_{\theta\in\Theta} \log p_\theta(y),
\end{align}
where $p_\theta(y) := \int_\mathcal{X} p_\theta(x, y) \mathrm{d} x$. We note that, given that $y$ is fixed, $p_\theta(y)$ is only a function of $\theta$, i.e., $p_\theta(y): \Theta \to \mathbb{R}$. The problem in \eqref{eq:mmle} corresponds to the \glsfirst*{MMLE} problem as introduced above, by maximizing the probability of the data under the given model, after integrating out latent variables. 

To provide the background for our method for solving the \gls*{MMLE}, we first introduce the \gls*{EM} algorithm below from a variational perspective, then move on to describe the \gls*{SVGD-EM}. For brevity, let us define $\ell(\theta, x) = \log p_\theta(x, y)$ as the joint log-likelihood function where $y$ is hidden from the notation $\ell(\theta, x)$ throughout as it is fixed.
\subsection{A variational view of EM}
\label{section: EM}
The \gls*{EM} algorithm is the de facto recipe that is applied for fitting \glspl*{LVM} \parencite{dempster1977maximum}. The \gls*{EM} is an iterative scheme that consists of two steps at iteration $t$ (i) E-step is the task of computing the expectation $Q(\theta, \theta_t) = \mathbb{E}_{p_{\theta_t}(x|y)}[ \log p_\theta(x, y)]$ and (ii) M-step is the task of computing the maximizer of $Q(\theta, \theta_t)$ w.r.t. $\theta$ (leading to the next iterate $\theta_{t+1}$). 
\textcite{neal1998view} showed that the \gls*{EM} algorithm can also be seen as a coordinate descent method on free energy 
$\mathcal{F} : \Theta  \times \mathcal{P}(\mathcal{X}) \to \mathbb{R} \cup \{\infty\}$ defined for $(\theta,q) \in \Theta \times \mathcal{P}(\mathcal{X})$ by
\begin{align*}
    \mathcal{F}(\theta,q) &:= \int_{\mathcal{X}} \log q(x) q(x) \md x - \int_{\mathcal{X}} \ell(\theta, x) q(x) \md x.
\end{align*}
Using this free energy, the steps of \gls*{EM} can be reformulated as 

\begin{minipage}{0.50\textwidth}
    \begin{flushleft}
        \begin{itemize}
            \label{eq: EM as coord GD}
            \item E-step : $q_t := \arg\min_{q \in \mathcal{P}(\mathcal{X})} \mathcal{F}(\theta_t,q)$
        \end{itemize}
    \end{flushleft}
\end{minipage}
\hfill
\begin{minipage}{0.65\textwidth}
    \begin{flushright}
        \begin{itemize}   
            \item M-step : $\theta_{t+1} := \arg\min_{\theta \in \Theta} \mathcal{F}(\theta,q_t)$
        \end{itemize}
    \end{flushright}
\end{minipage}

An important implication here is that for any $\theta \in \Theta$ the posterior $p_\theta(\cdot|y)$ minimizes $q \mapsto \mathcal{F}(\theta,q)$ \parencite{kuntz2023particle}. Moreover the marginal likelihood $p_\theta(y)$ has a global maximum at $\theta$ if and only if $\mathcal{F}(\theta,q)$ has a global maximum at $(\theta_\star,p_{\theta_\star}(\cdot|y))$. Thus, the idea is now to minimize $\mathcal{F}$ and this will be done in the Wasserstein space which gives a well studied notion of gradient \parencite{otto2001geometry, ambrosio2008gradient, villani2009optimal}.
\subsection{Euclidean-Wasserstein gradient flow of \texorpdfstring{$\mathcal{F}$ on $\Theta \times \Pac$}{}}

In order to minimize the free energy $\mathcal{F}$, a viable approach is to formalize its gradient flow on $\Theta \times \Pac$ where $\Theta$ is equipped with the standard Euclidean metric and $\Pac$ with the Wasserstein-2 metric. In this scenario, as presented in \textcite[Appendix~B.1]{kuntz2023particle} we have $\nabla \mathcal{F}(\theta, q) = (\nabla_\theta\mathcal{F}(\theta, q) ,  \nabla_q \mathcal{F}(\theta, q))$ with 
\begin{align*}
\nabla_\theta \mathcal{F}(\theta, q) &= - \int_{\mathcal{X}} \nabla_{\theta} \ell(\theta,x) \, q(x) \, \md x \\
\nabla_{q} \mathcal{F}(\theta, q) &= - \text{div}\big( q \WG \mathcal{F}(\theta, q) \big).
\end{align*}
The term $ \WG \mathcal{F}(\theta, q) : \mathcal{X} \to  \mathcal{X}$ is the velocity vector field commonly called the Wasserstein gradient of $\mathcal{F}$. It is defined by  $ \nabla \hspace{-0.22cm} \nabla \mathcal{F}(\theta, q) :=\nabla_x (\frac{\delta \mathcal{F}}{\delta q}(\theta,q))$, using Lemma 1 of Appendix A.3 in \textcite{kuntz2023particle} leads to the equality $\nabla \hspace{-0.22cm} \nabla \mathcal{F}(\theta, q) = \nabla_x \log ({q}/{p_{\theta}(\cdot,y)})$ and thus the \textit{Euclidean-Wasserstein} gradient flow of $\mathcal{F}$ can be written as
\begin{align}
    \partial_t \theta_t &= \int_\mathcal{X} \nabla_\theta \ell(\theta_t,x) q_t(x) \md x \label{eq:gradient_theta}\\
    \partial_t q_t  &=  \text{div} \left( q_t\nabla_x \log\left(\frac{q_t}{p_{\theta_t}(\cdot,y)}\right)\right). \label{eq:gradient_mu}
\end{align}
\textcite{kuntz2023particle} develop a method based on the flow \eqref{eq:gradient_theta}--\eqref{eq:gradient_mu} to perform \gls*{MMLE}. The method is called \gls*{PGD} and is based on discretizing a stochastic differential equation based on the flow \eqref{eq:gradient_theta}--\eqref{eq:gradient_mu} using the Euler scheme with a time step $\gamma > 0$ and approximating $q_t \approx (1/N) \sum_{i=1}^N \delta_{x_t^{(i)}}$ where $\{x_t^{(i)}\}_{0\leq i \leq N}$ are the particles at time $t$. This leads to a particle-based algorithm for \gls*{MMLE} \parencite{kuntz2023particle}, where particles are evolved with respect to a (time-discretized) SDE solving \eqref{eq:gradient_theta}, independently.

Let us consider the flow \eqref{eq:gradient_theta}--\eqref{eq:gradient_mu} discretized by a semi-implicit Euler scheme with a time step $\gamma > 0$ and obtain the following update rules:
\begin{align}
    \theta_{t+1}
    &= \theta_t + \gamma \int_{\mathcal{X}} \nabla_{\theta}\ell(\theta_t,x)\, q_t(x)\,\mathrm{d}x
    \label{eq:theta_update}\\
    q_{t+1}
    &= \Bigl(\mathrm{id} - \gamma \nabla_x \log \Bigl(\frac{q_t}{p_{\theta_{t+1}}(\cdot,y)}\Bigr)\Bigr)_{\#}\, q_t
    \label{eq:mu_pushforward}.
\end{align}

 Now we can make the important observation that after sampling $q_t$ with the particles $\{x_t^{(i)}\}_{0\leq i \leq N}$, \eqref{eq:mu_pushforward} can be seen as the update $x_{t+1}^{(i)} = x_t^{(i)} -\gamma\WG \mathcal{F}(\theta_{t+1}, q_t)(x_t^{(i)})$ which means the particles are pushed according to $-\WG \mathcal{F}(\theta_{t+1}, q_t)$. This reveals the role of  $-\nabla \hspace{-0.22cm} \nabla \mathcal{F}(\theta, q)(\cdot)$ as being the velocity vector fields driving the particles of a density following the Wasserstein Gradient Flow of $\mathcal{F}$ (with a first order approximation when we consider \eqref{eq:mu_pushforward} instead of \eqref{eq:gradient_mu} -- see \textcite[Proposition~8.4.6]{ambrosio2008gradient}). 
\subsection{SVGD-EM}
The \gls*{SVGD-EM} \parencite{sharrock2024tuning} algorithm relies on a closely relevant but different flow compared to \eqref{eq:gradient_theta}--\eqref{eq:gradient_mu}. In particular, the parameter component \eqref{eq:gradient_theta} stays the same but the density $q_t$ in \eqref{eq:gradient_mu} is driven by a different velocity vector field referred to as a \textit{Wasserstein gradient in the RKHS $\mathcal{H}_k$}. This is expressed as $\nabla \hspace{-0.22cm} \nabla_{H_k} \mathcal{F}(\theta, q) :\mathcal{X} \to \mathcal{X}$, defined by $\WG_{H_k} \mathcal{F}(\theta, q)(\cdot) := - \int_\mathcal{X} \left[ \nabla_x \ell(\theta,x) \, k(x, \cdot) + \nabla_1 k(x, \cdot) \right] \, q(x) \, \md x$ and which can be seen as an image of the standard Wasserstein gradient through an integral operator \parencite{sharrock2024tuning}. It is the same velocity vector field that drives the particles in the \gls*{SVGD} method, hence the name of \gls*{SVGD-EM}. The resulting semi-implicit time-discretized flow is then given by
\begin{align*}
    \theta_{t+1} &= \theta_t + \gamma \int_\mathcal{X} \nabla_{\theta} \ell(\theta_t,x)q_{t}(x)\md x \\
    q_{t+1} &= \bigg(\text{id} + \gamma \int_\mathcal{X} \big[k(x, \cdot) \nabla_x \ell(\theta_{t+1},x) \nonumber \\
    &+ \nabla_1 k(x, \cdot) \big] \, q_t(x) \, \md x )\bigg)_{\#}\, q_{t}.
\end{align*}
An implementable scheme can be obtained by using the particle approximation $q_t \approx (1/N) \sum_{i=1}^N \delta_{x_t^{(i)}}$. The resulting method takes the form of update rules:
\begin{align}
\theta_{t+1} &= \theta_t + \frac{\gamma}{N}\sum_{j=1}^{N} \nabla_{\theta} \ell(\theta_t,x_{t}^{j}) \label{eq:svgd-em-theta} \\
x_{t+1}^i &= x_{t}^i + \frac{\gamma}{N}\sum_{j=1}^{N} [k(x_{t}^j,x_{t}^i)\nabla_{x} \ell(\theta_{t+1},x_{t}^{j})  \label{eq:svgd-em-x} \\
&+ \nabla_{x_{t}^{j}} k(x_{t}^j,x_{t}^i)]. \nonumber
\end{align}

This algorithm has a few important differences with respect to \gls*{PGD}. First, the algorithm is deterministic (compared to stochastic latent variable updates of \gls*{PGD}), second, the $x$-particles are interacting through the kernel $k$ (non-local).

\section{Momentum {SVGD-EM}}
\label{sec:M-SVGD-EM}
In this section, we introduce our method, \gls*{M-SVGD-EM}, which is an accelerated version of the \gls*{SVGD-EM} algorithm. This method accelerates both parameter and latent variable updates and in the following sections we derive the accelerated updates first on the parameter space $\Theta$ and then on the space of measures $\Pac$.
\subsection{Acceleration in \texorpdfstring{$\Theta$}{}}
\label{section: acceleration theta}
We first introduce the (easier) acceleration scheme in the parameter space $\Theta$. This is taken from the standard optimization literature of accelerated gradient descent, see, e.g. \textcite{nesterov1983method, nesterov1988approach, nesterov2005smooth}. The motivation behind this choice is that the accelerated gradient descent can achieve $\mathcal{O}(1/t^2)$ convergence rate for smooth convex functions compared to gradient descent which only attains $\mathcal{O}(1/t)$, this acceleration is known to be a trade-off with stability as shown by  \textcite{attia2021algorithmic} but this was carefully monitored in our experiments. We adopt this strategy to replace the $\theta$-update in \eqref{eq:svgd-em-theta}. Recall that the standard gradient update for a loss function $f$, $\theta_{t+1} = \theta_t - \gamma \nabla_\theta f(\theta_t)$, can be accelerated by introducing a momentum term $\tilde{\theta}_t$ in the original scheme $\theta_{t+1} = \tilde{\theta}_t - \gamma \nabla_\theta f(\tilde{\theta}_t)$ where $\tilde{\theta}_t$ is updated by $\tilde{\theta}_{t+1} = \theta_{t+1} + \alpha_\theta (\theta_{t+1} - \theta_t)$ where $\alpha_\theta$ is a momentum parameter \parencite{nesterov2013introductory}.Applying this to our case, given a set of particles $\{x_t^{(i)}\}_{0\leq i \leq N}$ at time $t$ with the parameter update $\theta_t$, we can replace the update \eqref{eq:svgd-em-theta} with the following update:
\begin{align}
    {\theta}_{t+1} &= \tilde{\theta_{t}} + \frac{\gamma}{N}\sum_{i=1}^{N} \nabla_{\theta} \ell(\tilde{\theta_t},x_{t}^{(i)}) \label{eq:accel_theta}, \\
    \tilde{\theta}_{t+1} &= {\theta}_{t+1} + \alpha_\theta ({\theta}_{t+1} - {\theta}_t). \label{eq:accel_theta2}
\end{align}
This is the standard Nesterov scheme, applied to the parameter update of \gls*{SVGD-EM}. Equipped with \eqref{eq:accel_theta}--\eqref{eq:accel_theta2}, we now turn to the task of accelerating \gls*{SVGD-EM} in the space of probability measures.
\subsection{Acceleration in \texorpdfstring{$\Pac$}{}}\label{section: acceleration q}

To accelerate the latent variable update in \eqref{eq:svgd-em-x} on \gls*{SVGD}, we adopt the acceleration scheme \gls*{SVGD-WNes} proposed by \textcite{liu2019understanding} which uses a Nesterov scheme on $\Pac$ to accelerate the interacting particle sampling algorithm \gls*{SVGD} and builds upon the \gls*{RAGD} method of \textcite{zhang2018estimate}. 

The idea of \gls*{RAGD} is similar as in the Euclidean case with accelerated gradient descent, but with the key difference that, instead of linearly combining vectors, each iterate update is performed using exponential maps. The exponential map at $q$ is defined by $\text{Exp}_q(v) := (\text{id} + v)_{\#}\,q$. This map assigns to a vector tangent in $q$ noted $v \in \mathcal{T}_q\Pac$, an associated measure $(\text{id} + v)_{\#}\,q$ by moving along the geodesic with velocity $v$ in $q$ for a unit of time \parencite[Prop.~8.4.6]{ambrosio2008gradient}. This velocity vector field $v(\cdot)$ serves a role analogous to the standard gradient in the accelerated gradient descent. Using the exponential map, applied to the \gls*{SVGD} dynamics driven by $\WG_{H_k}\mathcal{F}(\theta, \cdot)$ with fixed $\theta$, the \gls*{RAGD} method proposed by \textcite{zhang2018estimate} can be written as:
\begin{align}
q_{t+1} &= \text{Exp}_{\tilde{q}_{t}}(-\gamma \WG_{H_k} \mathcal{F}(\theta, \tilde{q}_{t})) \label{eq:RAGD_1}\\
\tilde{q}_{t+1} &= \text{Exp}_{q_{t+1}} \{ c_1 \text{Exp}_{q_{t+1}}^{-1} [ \text{Exp}_{\tilde{q}_{t}} ( \text{Exp}_{\tilde{q}_{t}}^{-1}(q_{t+1}) \nonumber \\
&+ (c_2-1)(\text{Exp}_{\tilde{q}_{t}}^{-1}(q_{t+1}) - \text{Exp}_{\tilde{q}_{t}}^{-1}(q_{t})))] \}, \label{eq:RAGD_2}
\end{align}
where $c_1, c_2 \in \mathbb{R}^+$ are constants. The eq.~\eqref{eq:RAGD_1} is analogous to the first step of the accelerated gradient descent and the eq.~\eqref{eq:RAGD_2} is a momentum update as in the accelerated gradient descent. The main bottleneck of implementing \eqref{eq:RAGD_1}--\eqref{eq:RAGD_2} is the computation of the exponential map and its inverse \parencite{pele2009fast}. To alleviate this issue and manage to have a time complexity inferior to the $O(N^2)$ of the Sinkhorn method \parencite{cuturi2013sinkhorn, xie2020fast}, \textcite{liu2019understanding} introduce an approximation to this exponential map inverse and obtain the algorithm termed \glsfirst*{SVGD-WNes}. Given two empirical measures $q_t \approx \sum_i^N \delta_{x_t^{(i)}}$ and $\tilde{q}_t \approx \sum_i^N \delta_{\tilde{x}_t^{(i)}}$, to provide a computable method, \textcite{liu2019understanding} assumes a closeness relationship between particles and momentum particles and between particles of successive iterations leading to the approximation 
\begin{equation} 
\label{eq: pairwise close}
\left(\text{Exp}_{q_{t}}^{-1}(\tilde{q}_t)\right)(x_t^{(i)}) \approx  \tilde{x}_t^{(i)} - x_t^{(i)}, \quad  1 \leq i \leq N.
\end{equation}
Using the approximation \eqref{eq: pairwise close} results in the \gls*{SVGD-WNes} method which takes the form of update rules:
\begin{align}
x_{t+1}^{(i)} &= \tilde{x}_{t}^{(i)} + \frac{\gamma}{N}\sum_{j=1}^{N} k(\tilde{x}_{t}^{(j)},\tilde{x}_{t}^{(i)})\nabla_x \ell(\theta,\tilde{x}_{t}^{(j)}) \label{eq:svgd-wnes-x}\\
&+ \nabla_{1}k(\tilde{x}_t^{(j)},\tilde{x}_{t}^{(i)})  \nonumber \\
\tilde{x}_{t+1}^{(i)} &= x_{t+1}^{(i)} + c_1 (c_2 - 1) (x_{t+1}^{(i)} - x_{t}^{(i)}) \label{eq:svgd-wnes-x2}
\end{align}
where $c_1, c_2 \in \mathbb{R}^+$ are constants. We provide a self-contained derivation of eqs.~\eqref{eq:svgd-wnes-x}--\eqref{eq:svgd-wnes-x2} from \eqref{eq:RAGD_1}--\eqref{eq:RAGD_2} in Appendix~\ref{appendix: SVGD-WNes}.
\subsection{Momentum SVGD-EM}
\label{section: M-SVGD-EM}
Having derived acceleration schemes for both the parameter and latent variable updates, we can now combine them to obtain our final method. The resulting algorithm is a combination of the two schemes described in Sections~\ref{section: acceleration theta} and \ref{section: acceleration q}. The resulting method is called \gls*{M-SVGD-EM} and is summarized in Algorithm~\ref{alg:svgd-wnes-em-nes}. The main difference with respect to \gls*{SVGD-EM} is that we replace the update \eqref{eq:svgd-em-theta} with the accelerated update \eqref{eq:accel_theta}--\eqref{eq:accel_theta2} and the update \eqref{eq:svgd-em-x} with the accelerated update \eqref{eq:svgd-wnes-x}--\eqref{eq:svgd-wnes-x2}.
\begin{algorithm}[h]
\caption{M-SVGD-EM}\label{alg:svgd-wnes-em-nes}
\begin{algorithmic}[1]
\STATE {\bfseries Input:} momentum constants $\alpha_X, \alpha_\theta \in \mathbb R$, initial particles $\{x_0^{(i)}\}_{i=1}^{N} \sim q_0$, initial parameter $\theta_0$, joint log-likelihood $\ell$, kernel $k$, learning rate $\gamma$.
\STATE {\bfseries Initialize:} $\theta_0 = \tilde{\theta}_0$ and $\{x_0^{(i)}\}_{i=1}^{N} = \{\Tilde{x}_0^{(i)}\}_{i=1}^{N}$
    \FOR{$t = 0,1,\dots, T-1$} 
        \STATE
        $\theta_{t+1} = \tilde{\theta}_t + \frac{\gamma}{N}\sum_{j=1}^{N} \nabla_{\theta} \ell(\tilde{\theta}_t,x_{t}^{(j)})$
        \STATE $\tilde{\theta}_{t+1} = \theta_{t+1} + \alpha_\theta(\theta_{t+1} - \theta_{t})$
        \FOR{$i = 1,2,\dots, N$}             
            \STATE$x_{t+1}^{(i)}\hspace{-0.15cm}=\tilde{x}_{t}^{(i)}\hspace{-0.1cm}+\hspace{-0.05cm}\frac{\gamma}{N}\sum_{j=1}^{N}k(\tilde{x}_{t}^{(j)},\tilde{x}_{t}^{(i)})\nabla_{x} \ell(\theta_{t+1},\tilde{x}_{t}^{(j)})$
            \vspace{-0.3cm}
            \STATE $\hspace{1.44cm} + \nabla_{1}k(\tilde{x}_{t}^{(j)},\tilde{x}_{t}^{(i)})$
            \STATE $\tilde{x}_{t+1}^{(i)} = x_{t+1}^{(i)} + \alpha_X (x_{t+1}^{(i)} - x_{t}^{(i)})$
        \ENDFOR
    \ENDFOR
\STATE \textbf{Output:} $\theta_{T}$ and $\{x_{T}^{i}\}_{i=1}^{N}$
\end{algorithmic}
\end{algorithm}

\section{Numerical experiments}
\label{section: Numerical experiments}
 We now compare the performance of \gls*{M-SVGD-EM} with a number of other algorithms, including \gls*{PGD} \parencite{kuntz2023particle}, \gls*{MPGD} \parencite{lim2023momentum}, \gls*{SOUL} \parencite{de2021efficient}, and \gls*{SVGD-EM} \parencite{sharrock2024tuning}. The code used to run the following experiments is publicly available\footnote{\hyperlink{https://github.com/Farattel/m-svgd-em}{{https://github.com/Farattel/m-svgd-em}}}.  We benchmark these methods on a number of examples, specifically, on a toy hierarchical model, a Bayesian logistic regression model, and a Bayesian neural network. The goal of these experiments is to better understand the acceleration properties of \gls*{M-SVGD-EM} compared to non-accelerated variants. In the original implementation of \gls*{SVGD-EM} \parencite{sharrock2024tuning}, the kernel of choice is Median RBF kernel $k(x,x') = \exp(- h^{-1}||x-x'||^2)$  with the median heuristic introduced by \textcite{liu2016stein}. In the results that follow, we have used AutoRBF provided in the codebase of the original paper \parencite{sharrock2024tuning}. The AutoRBF kernel is defined as $k(x, x') = \exp\left(-2h^{-2} \|{x} - {x'}\|^2\right)$ where we use JAX \parencite{jax2018github} to handle gradient calculations. Here, $h > 0$ is the bandwidth parameter that was chosen to be $h=5.0$ in our experiments. We selected this kernel over MedianRBF as it demonstrated better performance and more pronounced acceleration effects in our empirical tests. Results using MedianRBF can be found in the Appendix~\ref{appendix:medianrbf}. All experiments were run on a customer grade laptop with CPU 12th Gen Intel(R) Core(TM) i7-12700H 2.30 GHz 32 GB RAM.

\subsection{Toy Hierarchical Model}\label{sec:toy}

We first evaluate the algorithms on a toy hierarchical model. We replicate the experimental settings of \textcite{kuntz2023particle} and \textcite{sharrock2024tuning}.
The model under consideration can be described as
\begin{align*}
p_\theta(x,y) = \prod_{i=1}^{d_y} \mathcal{N}(y_i; x_i,1)\,\mathcal{N}(x_i; \theta,\sigma ^2).
\end{align*}
The model assumes that we observe the data $(y_1,\dots,y_{d_y}) \in \mathbb R ^{d_y}$ generated via first sampling the latent variables $x_i \sim \mathcal{N}(x_i;\theta,\sigma  ^2)$ and then $\mathcal{N}(y_i; x_i,1)$.
In this model the marginal likelihood $\theta \mapsto p_\theta (y)$ has a unique maximizer
$\hat{\theta}_{d_y} = d_y ^{-1} \sum_{i=1}^{d_y} y_i$ and one can even find an explicit expression of the posterior
$p_{\hat{\theta}_{d_y}}(.|y_i)$ (see \textcite{kuntz2023particle} Sect E.1).

For the sake of comparison, we use an alternative version of this model that can be found in \textcite{lim2023momentum},  ToyHM$(10,12)$ \emph{i.e.} we fix $\sigma=12$ and generate the data using $\theta=10$, where we generate a single observation of $d_y=20$ dimensions, particle cloud is initialized by $X_0\sim \mathcal{N}(0,1) $ using $N=20$ particles.

\begin{figure*}[ht]
  \centering
  \begin{subfigure}[h]{0.28\textwidth}
    \includegraphics[height=3.5cm]{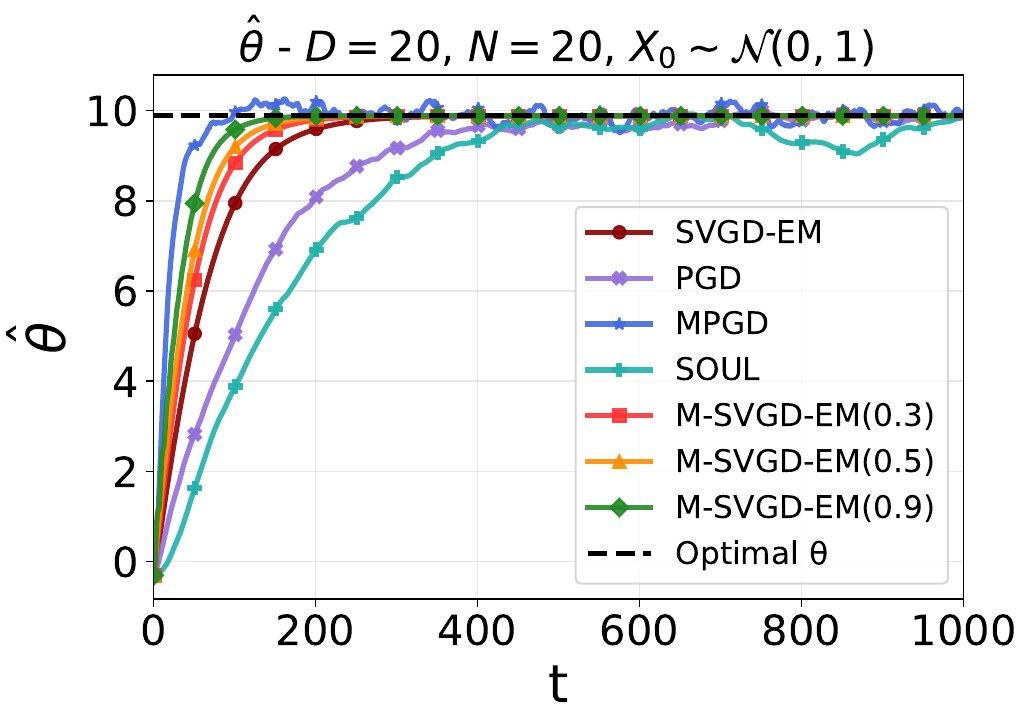}
    \caption{Parameter Estimation}
    \label{fig:toy_param_estimates_paper}
  \end{subfigure}
  \hfill
  \begin{subfigure}[h]{0.28\textwidth}
    \includegraphics[height=3.5cm]{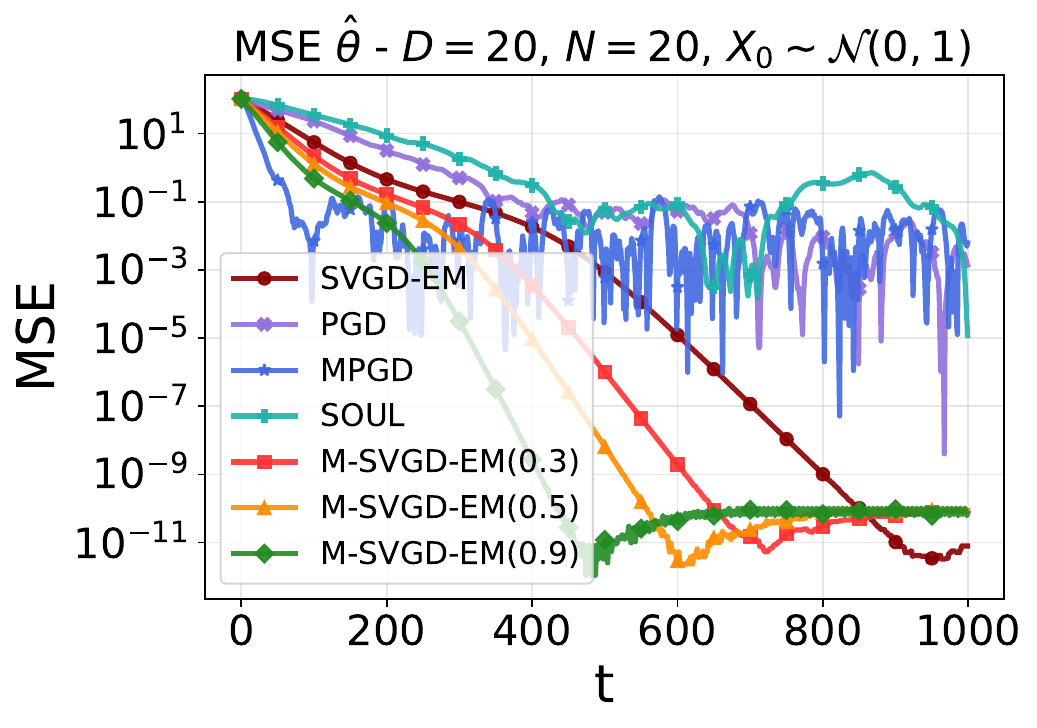}
    \caption{MSE with $\hat{\theta}_{d_y}$}
    \label{fig:toy_mse_paper}
  \end{subfigure}
  \hfill
  \begin{subfigure}[ht]{0.32\textwidth}
    \includegraphics[height=3.5cm]{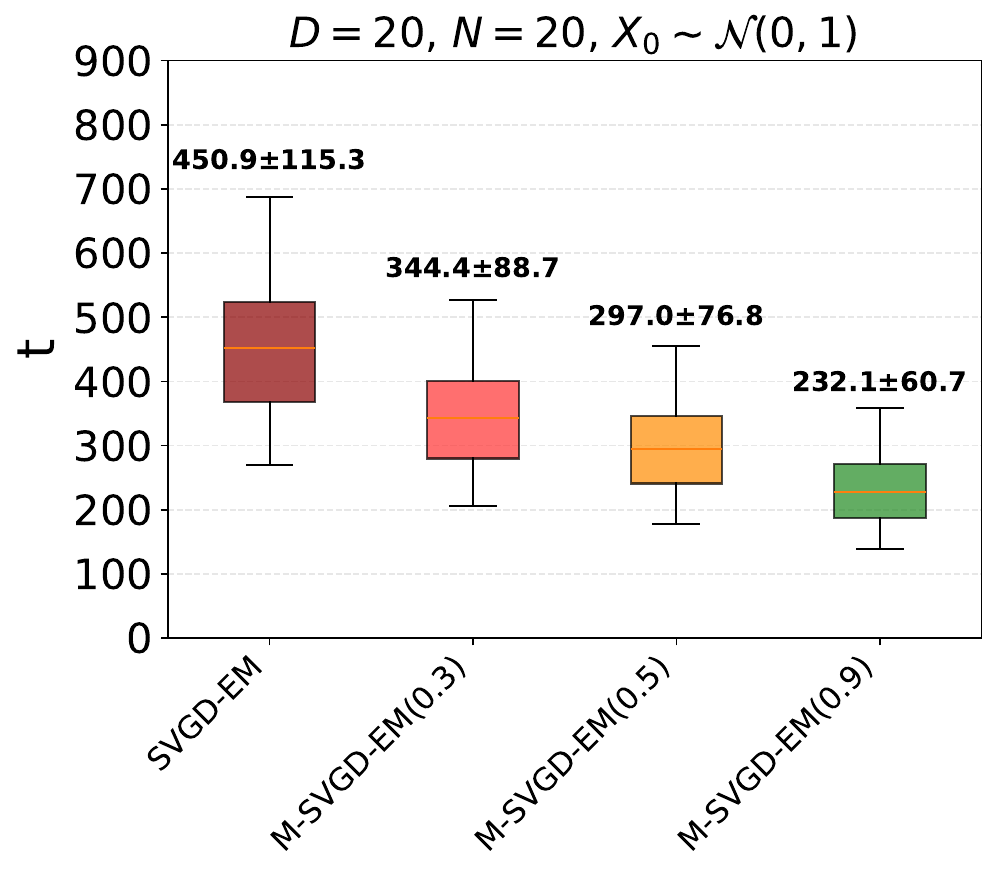}
    \caption{Average iterations}
    \label{fig:toy_boxplot_paper}
  \end{subfigure}
  \caption{Comparison of \gls*{M-SVGD-EM} with the \gls*{SVGD-EM}. Part (a) shows the performance on parameter estimation on a single iteration on the ToyHM(10, 12) with different acceleration parameters $\alpha_{\theta}=\alpha_{X}=(0.3,0.5,0.9)$.(b) displays MSE with respect to the empirical minimizer for each algorithm. (c) shows the average number of iterations until convergence, defined as being within a threshold of 0.05 from the empirical minimizer. The results are averaged over 20 independent trials.}
  \label{fig:toy_paper}
\end{figure*}

\begin{figure*}[h!]
  \centering
  \begin{subfigure}[h]{0.32\textwidth}
    \includegraphics[height=3.5cm]{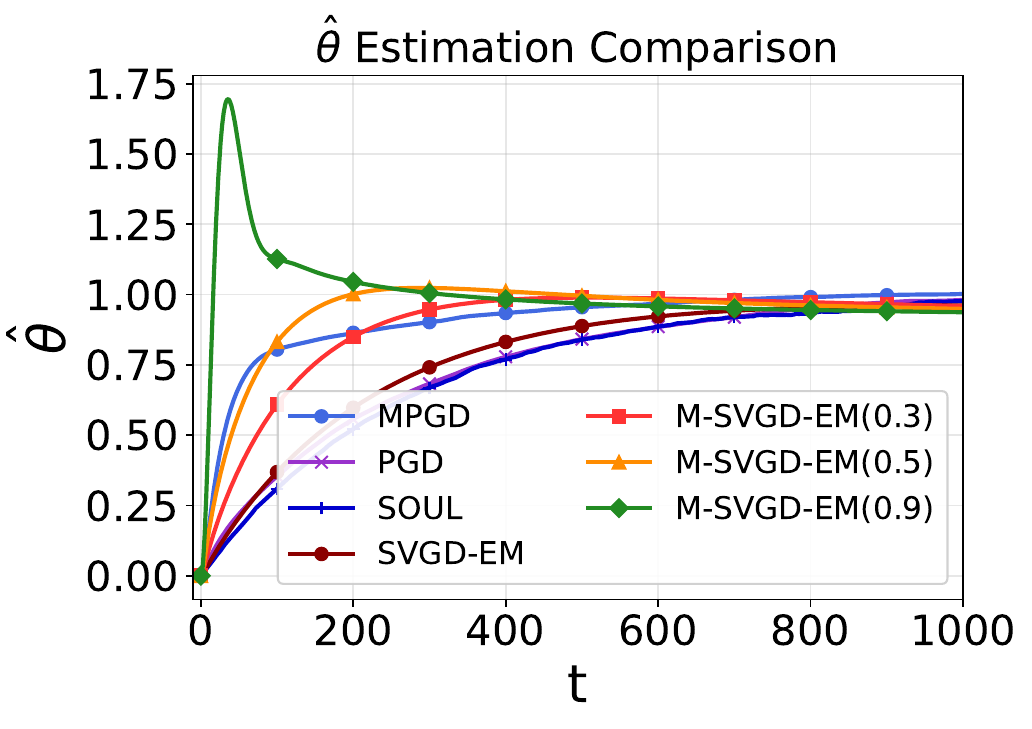}
    \caption{Parameter Estimation}
    \label{fig:blr_para_paper}
  \end{subfigure}
  \hfill
  \begin{subfigure}[h!]{0.32\textwidth}
    \includegraphics[height=3.5cm]{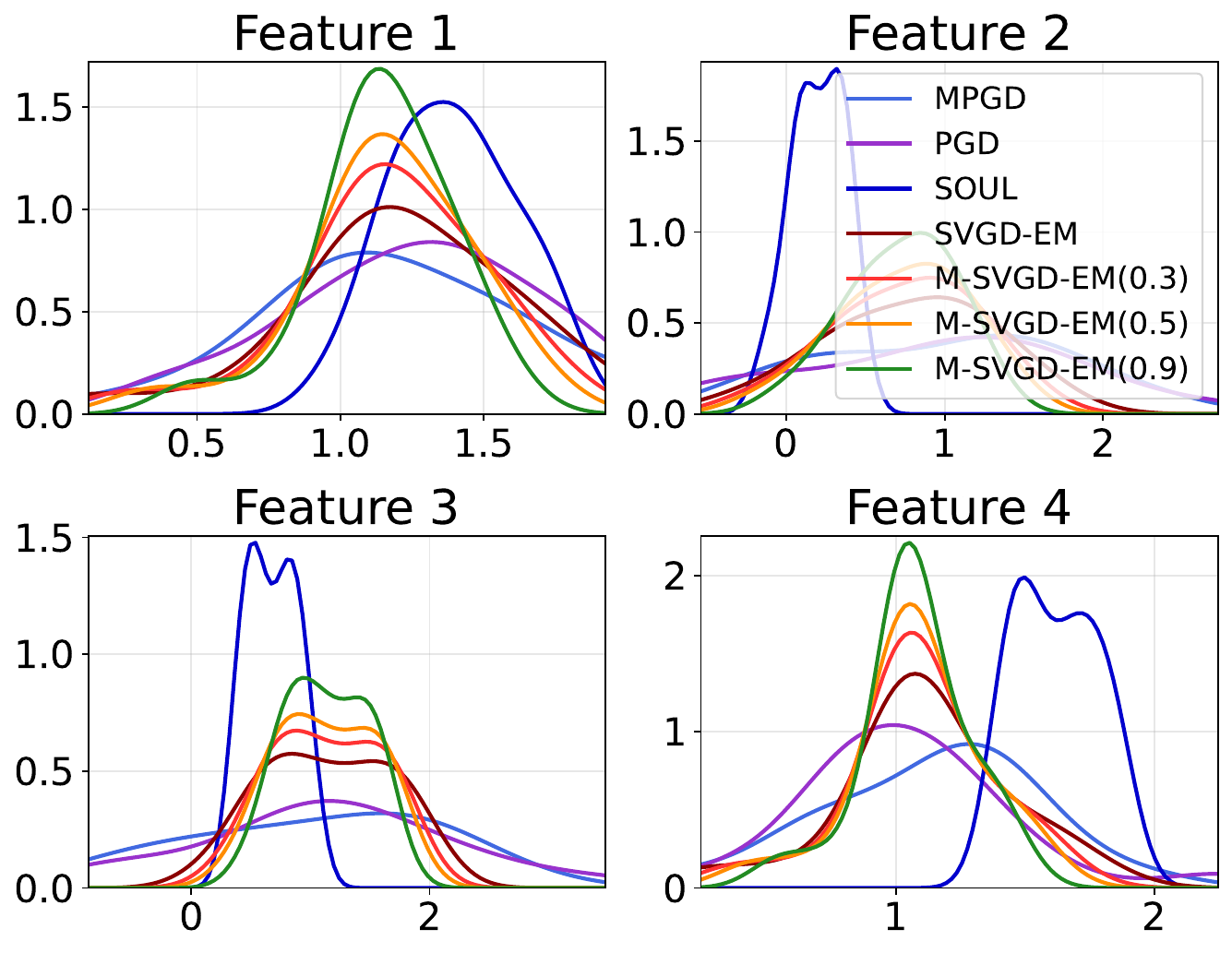}
    \caption{Posterior Estimates}
    \label{fig:blr_features_paper}
  \end{subfigure}
  \hfill
  \begin{subfigure}[h!]{0.32\textwidth}
    \includegraphics[height=3.5cm]{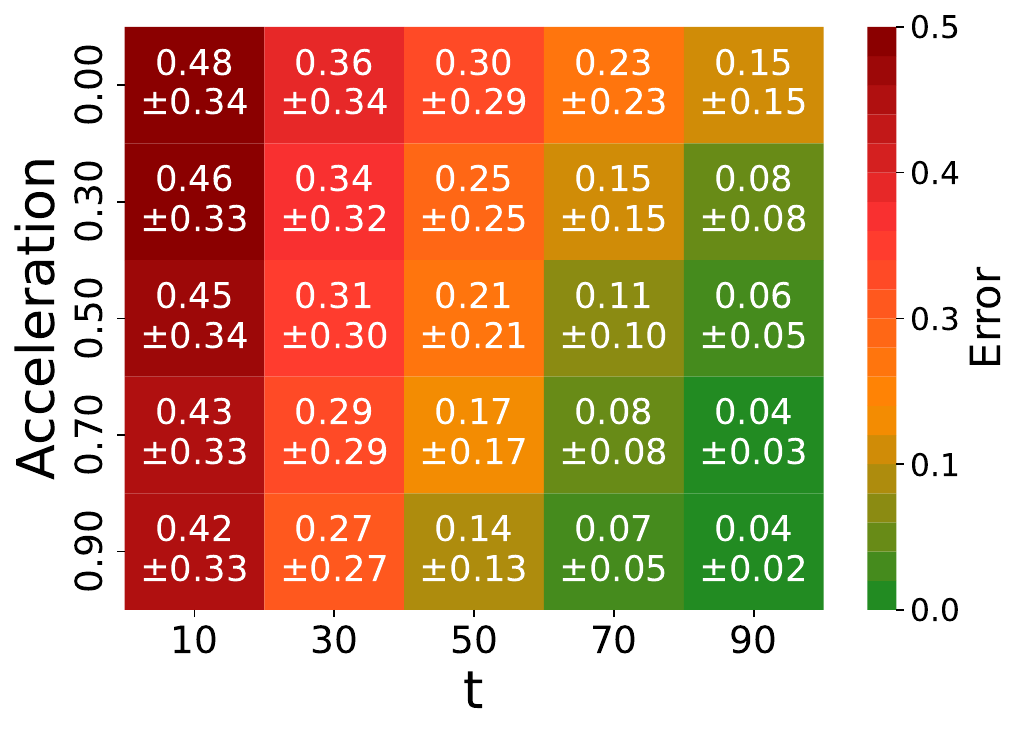}
    \caption{Test error Heatmap}
    \label{fig:blr_heat_paper}
  \end{subfigure}
  \caption{(a) Parameter estimation convergence on BLR varying acceleration. (b) Posterior distributions across test features, showing acceleration improves convergence. (c) Test error vs. acceleration for different $t$ values, showing error reduction with increased acceleration.}
  \label{fig:blr_paper}
\end{figure*}
 All algorithms are initialized with parameter $\theta_0 \sim \mathcal{U}(-3,3)$. To determine the optimal learning rates $\gamma$, we conduct a fine grid of 100 logarithmically spaced values in $\gamma\in[10^{-3}, 10^2]$. For each $\gamma$ we run the algorithms for a fixed number of iterations $T=1000$ and select the $\gamma$ which minimizes the MSE for each algorithm over 10 different trials. In Figure \ref{fig:toy_paper} results displayed use standard gradient descent with $\gamma=0.01$ for \gls*{SOUL},$\gamma=0.2$ for both \gls*{PGD} and \gls*{MPGD},$\gamma=0.3$ for both \gls*{SVGD-EM} and \gls*{M-SVGD-EM}.
In Figure \ref{fig:toy_param_estimates_paper} we observe that \gls*{M-SVGD-EM} consistently outperforms the \gls*{SVGD-EM} across different acceleration parameters $\alpha = \alpha_{\theta}=\alpha_{X}$. Figure \ref{fig:toy_mse_paper} shows faster drop on the MSE for all accelerations and in particular acceleration $\alpha = 0.9$ achieves the same MSE as \gls*{SVGD-EM} in approximately $50\% $ of its iterations. Figure \ref{fig:toy_boxplot_paper} demonstrates substantial computational savings, with \gls*{M-SVGD-EM}(0.9) requiring only $232\pm60.7$ iterations to converge compared to $450.9\pm115.1$ for \gls*{SVGD-EM}, representing almost $50\%$ number of iterations on average and standard error. These results are averaged over 20 seeds indicating \gls*{M-SVGD-EM} consistent improvement from \gls*{SVGD-EM} the ToyHM(10, 12) model.

\subsection{Bayesian Logistic Regression}
\label{sec:blr_experiments}
We now consider a Bayesian Logistic Regression Model on the \textit{Wisconsin Breast Cancer Dataset} \parencite{breast_cancer_wisconsin_17} with Gaussian priors on the latent variables represented by the regression weights. The prior on the weights is $\mathcal{N}(\theta \textbf{1}_{d_x}, 5I_{d_x})$ and we consider the labels $l$ are independent conditioned on the weights and features. The Bayesian Logistic regression model is used to predict the labels based on the data and latent variables, the model supposes $p_\theta(\mathcal{Y} |x) = \prod_{(f,l)\in \mathcal{Y}}s(f^{T}x)^l [1-s(f^{T}x)]^{1-l}$. The joint model density is then:
\begin{align*}
    p_\theta(x,\mathcal{Y}) &= p_\theta(x)p(\mathcal{Y}|x) \\
    &= \mathcal{N}(\theta \textbf{1}_{d_x}, 5I_{d_x}) \hspace{-1mm} \prod_{(f,l)\in \mathcal{Y}}\hspace{-1mm} s(f^{T}x)^l [1-s(f^{T}x)]^{1-l}, \nonumber
\end{align*}
where $f$  denotes the features, $l$ the labels, $s$ is the sigmoid function $s(x) = (1 + e^{-x})^{-1}$ and $\mathcal{Y}$ is the data set. In Figure \ref{fig:blr_paper}, we present the results using standard gradient descent with $N=20$ particles, a step-size of $\gamma=0.01$ for all methods determined by a fine-grid search over $100$ logarithmically spaced ranged from $\gamma\in[10^{-7}, 10^2]$ and averaged over 5 distinct seeds for a fixed number of $T=500$ iterations.

We notice in Figure~\ref{fig:blr_para_paper}.(a) that for all $\alpha$ (acceleration) parameters \gls*{M-SVGD-EM} outperforms the non-accelerated methods \gls*{SVGD-EM}, \gls*{SOUL}, and \gls*{PGD}. However, the accelerated method \gls*{MPGD} is a competitive approach for which the number of time-steps required to converge is similar to \gls*{M-SVGD-EM}.

Figure \ref{fig:blr_features_paper} presents the kernel density estimates (KDEs) of their final particle distributions showing the distribution of the first 4 features. We see that as $\alpha$ grows \gls*{M-SVGD-EM} the posterior estimates reach a higher peak and tighter distribution implying a smaller variance, showing that acceleration provides a more confident estimate rather than \gls*{SVGD-EM}. In Figure \ref{fig:blr_heat_paper}, we demonstrate the average test error over 20 independent $(80\%,20\%)$ train and test respectively splits using a step size $\gamma=10^{-5}$ which is again within the optimal range as shown in Appendix
\ref{appendix:Blr}.

We use $N=20$ particles and we vary the acceleration and the number of iterations $T$. The higher the acceleration the more rigorously the test error drops achieving a better performance with less iterations.

\begin{figure*}[ht!]
\centering
\begin{subfigure}{0.32\textwidth}
    \includegraphics[width=\linewidth, height=4cm]{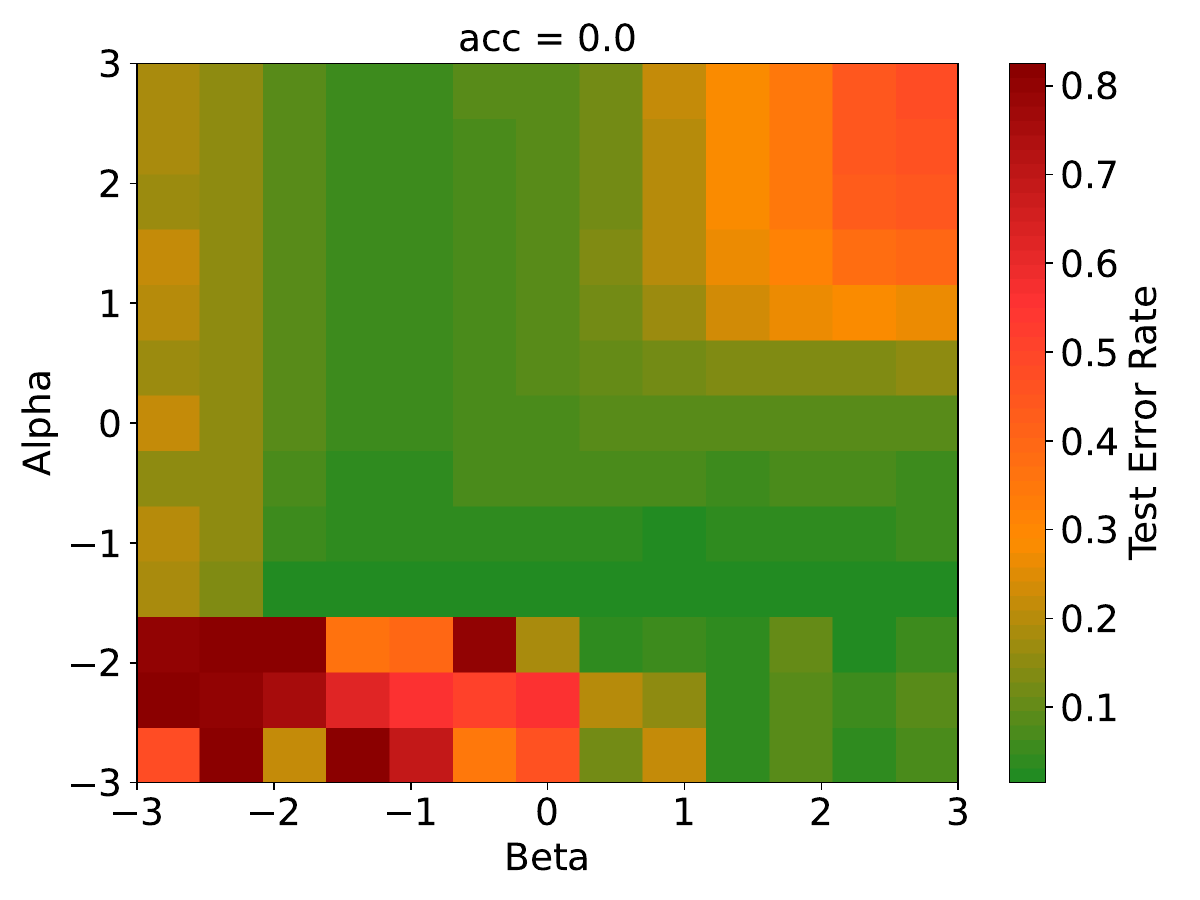}
    \caption{\gls*{SVGD-EM}}
    \label{fig:bnn_alpha_beta_acc0_appendix}
\end{subfigure}
\hfill
\begin{subfigure}{0.32\textwidth}
    \includegraphics[width=\linewidth, height=4cm]{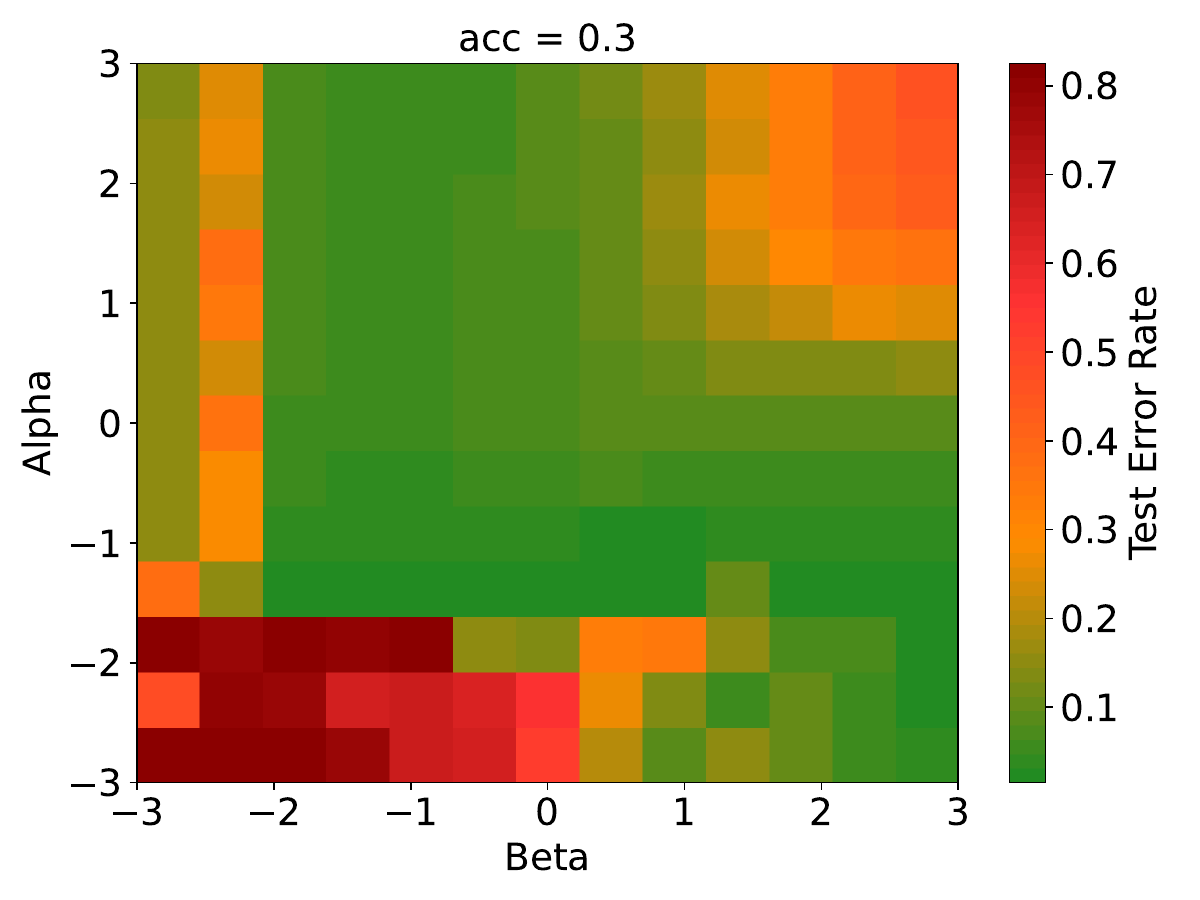}
    \caption{\gls*{M-SVGD-EM}$(0.3)$}
    \label{fig:bnn_alpha_beta_acc03_appendix}
\end{subfigure}
\hfill
\begin{subfigure}{0.32\textwidth}
    \includegraphics[width=\linewidth, height=4cm]{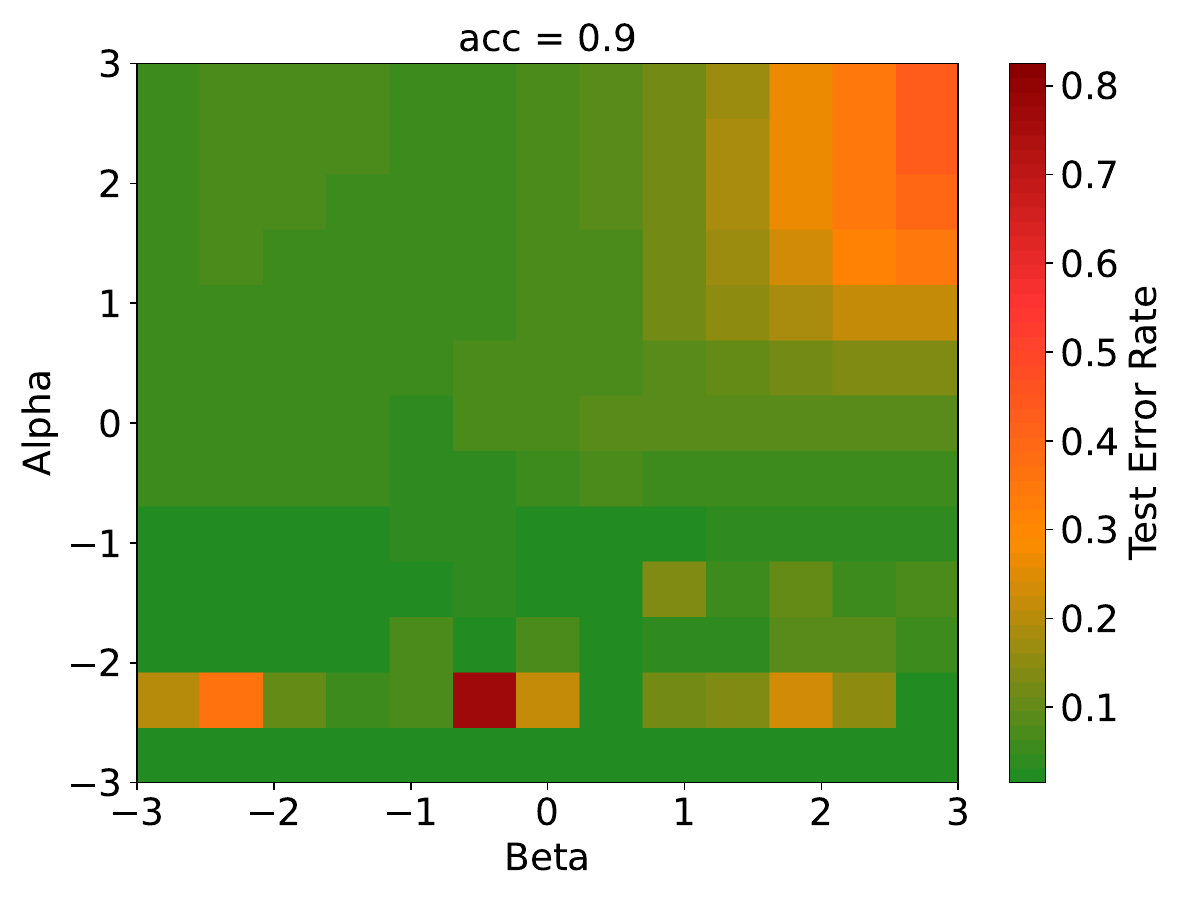}
    \caption{\gls*{M-SVGD-EM}$(0.9)$}
    \label{fig:bnn_alpha_beta_acc09_appendix}
\end{subfigure}
\caption{Different Initilisations of $\alpha$ and $\beta$ for accelerations $(0,0.3,0.9)$ with $N=10$ for $T=500$.}
\label{fig:bnn_alpha_beta_appendix}
\end{figure*}
\subsection{Bayesian Neural Network MNIST}
\label{sec:bnn_experiments}
We consider the following setup that is also described in \textcite{kuntz2023particle,sharrock2024tuning} for the binary classification task on the MNIST dataset between digits $4$ and $9$. The dataset can be described as
$\mathcal{Y} = \bigl\{(z_i, y_i)\bigr\}_{i=1}^{N_\mathcal{Y}},
z_i \in \mathbb{R}^{784},\; y_i \in \{0,1\},\; N_\mathcal{Y} = 1000$
where each feature meaning each pixel-dimension has been standardized across the dataset. The model parameters are $x = (w, v)$ where $w \in \mathbb{R}^{40\times784}$, $v \in \mathbb{R}^{2\times40}$. We assume the likelihood
\begin{align*}
p(y_i | z_i, x) {\propto} \exp\left( \sum_{j=1}^{40} v_{ij} \textnormal{tanh}\left(\sum_{k=1}^{784} w_{jk} z_{ik} \right) \right).
\end{align*}

\begin{figure}[h!]
  \centering
  \begin{subfigure}{0.45\textwidth}
    \includegraphics[height=4.5cm]{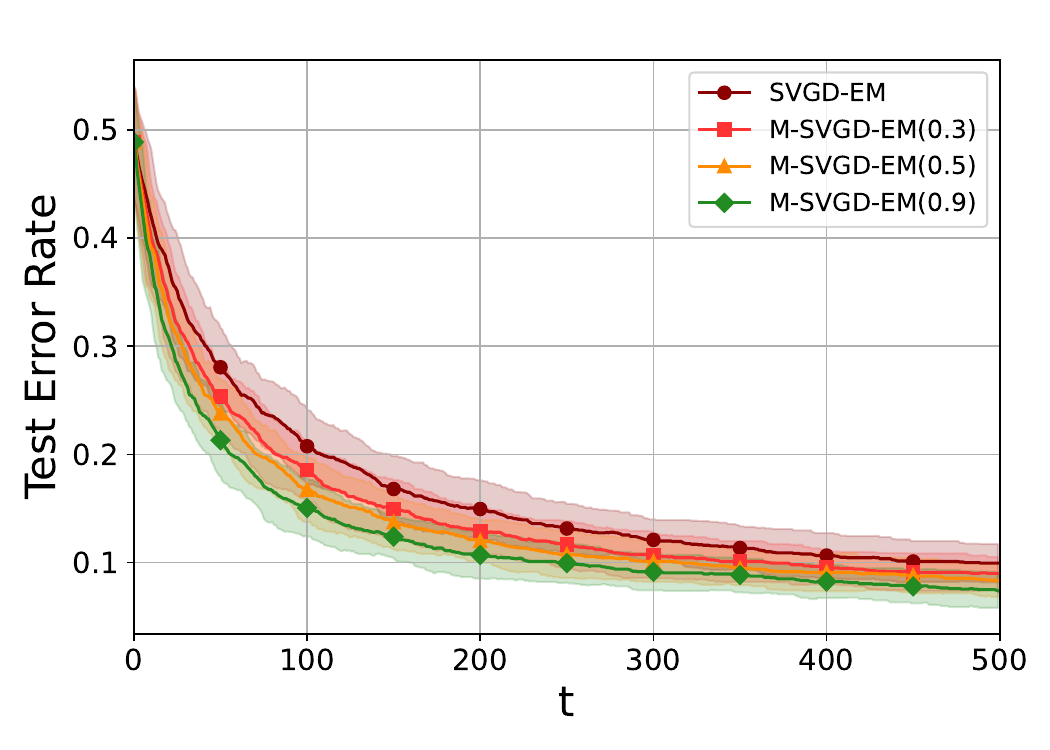}
    \caption{Test error, $(\alpha_0,\beta_0)=(0,0)$}
    \label{fig:bnn_test_0_paper}
  \end{subfigure}
  \hfill
  \begin{subfigure}{0.45\textwidth}
    \includegraphics[height=4.5cm]{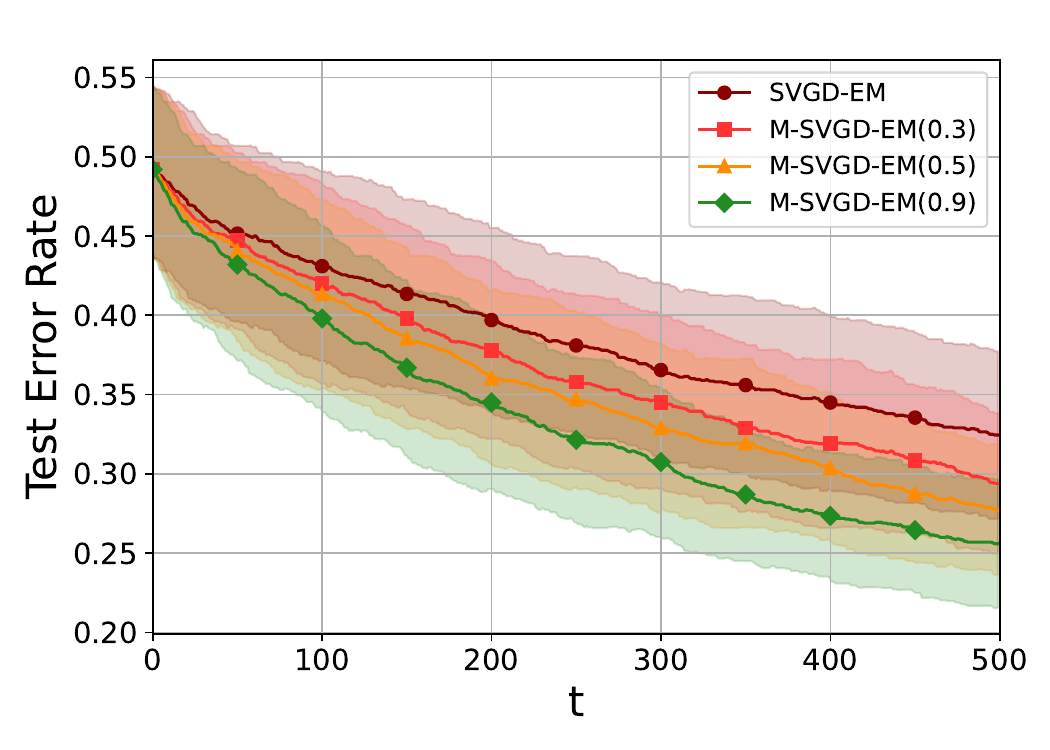}
    \caption{Test error, $(\alpha_0,\beta_0)=(2,2)$}
    \label{fig:bnn_test_2_paper}
  \end{subfigure}
  \caption{Test Error w.r.t iterations $t$ for the BNN on MNIST dataset}
  \label{fig:bnn_test_paper}
\end{figure}

\begin{figure}[h!]
  \centering
  \begin{subfigure}{0.45\textwidth}
    \includegraphics[height=4.5cm]{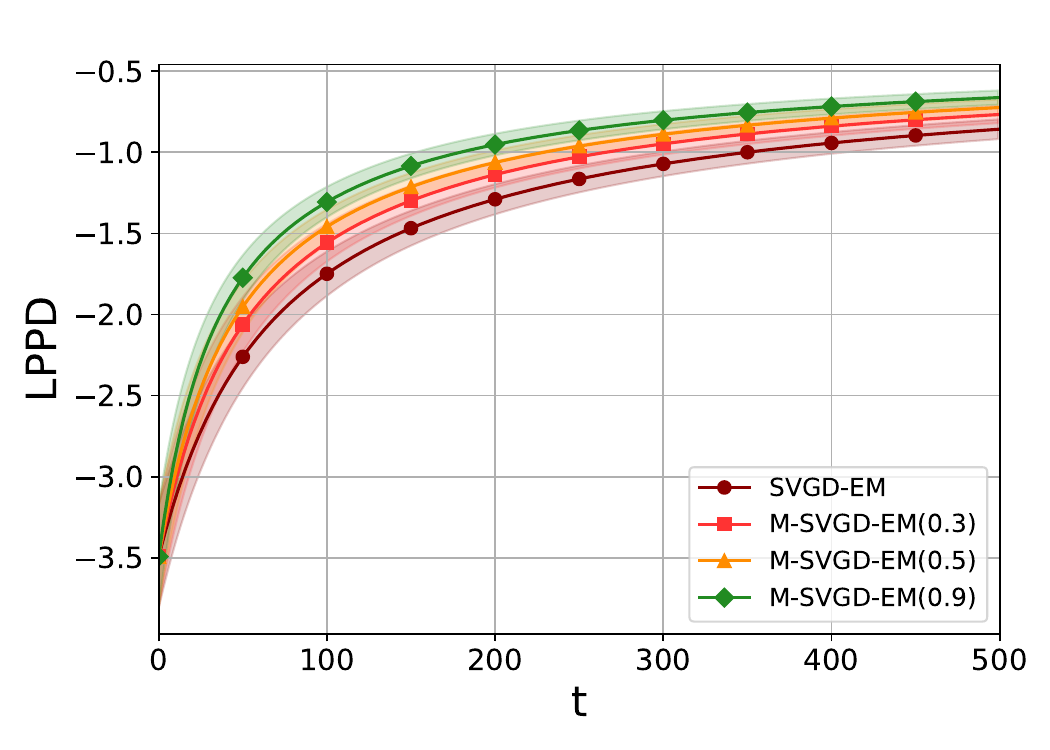}
    \caption{Lppd, $(\alpha_0,\beta_0)=(0,0)$}
    \label{fig:bnn_lppd_0_paper}
  \end{subfigure}
  \hfill
  \begin{subfigure}{0.45\textwidth}
    \includegraphics[height=4.5cm]{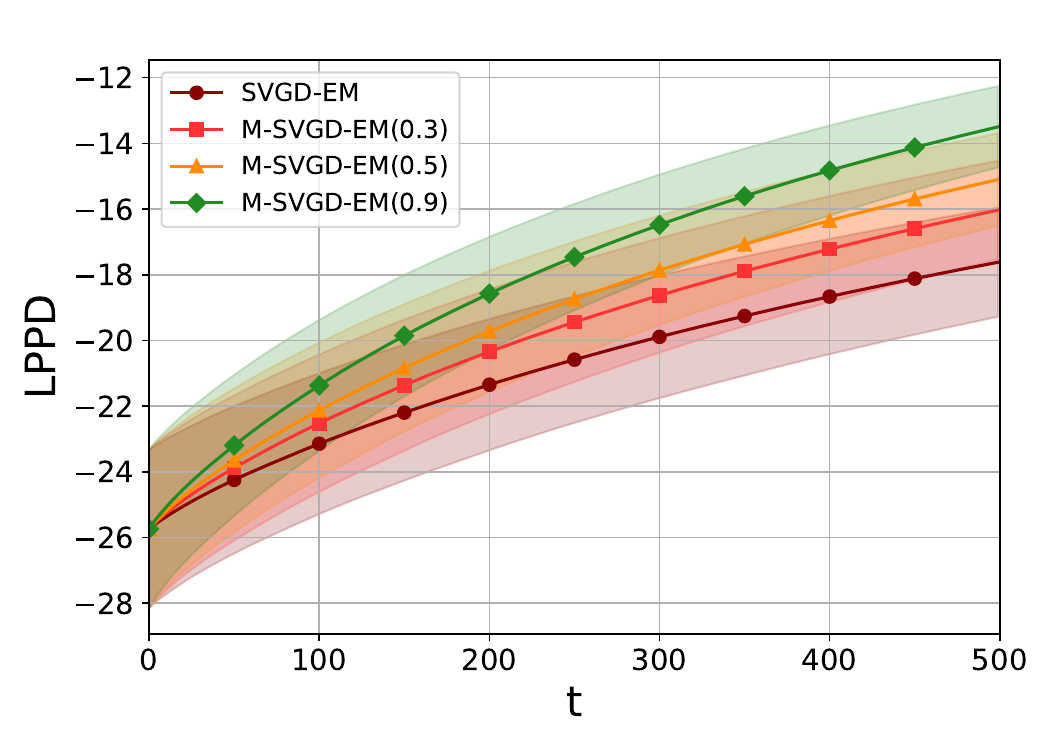}
    \caption{Lppd, $(\alpha_0,\beta_0)=(2,2)$}
    \label{fig:bnn_lppd_2_paper}
  \end{subfigure}
  \caption{Log predictive probability distribution vs iterations $t$ for the BNN on MNIST dataset}
  \label{fig:bnn_lppd_paper}
\end{figure}

We place independent Gaussian priors with log‐variances $\theta=(\alpha,\beta)$, i.e.
$
w_{jk} \sim \mathcal{N}\bigl(0, e^{2\alpha}\bigr)$ and $v_{ij} \sim \mathcal{N}\bigl(0, e^{2\beta}\bigr)$. Thus the joint density is
\begin{align*}
p_{\theta}(x, \mathcal{Y})
=
&\Bigl[\!\prod_{j=1}^{40}\prod_{k=1}^{784}\mathcal{N}(w_{jk};0,e^{2\alpha})\Bigr]  \\
\times&\Bigl[\!\prod_{c=0}^{1}\prod_{j=1}^{40}\mathcal{N}(v_{cj};0,e^{2\beta})\Bigr]
\,
\prod_{i=1}^N p(y_i \mid z_i, x).
\end{align*}
We run our method with $N = 10$. The optimizer used for the both updates in this experiment is AdaGrad \parencite{duchi2011adaptive} for the gradient descent step, with our additional momentum acceleration term applied post-update, effectively forming a Momentum-AdaGrad hybrid can be found in the Appendix \ref{alg:adagrad}. All methods are run for $T=500$ iterations and $\gamma=0.1$. We run the experiment for 10 independent trials and used the data split $(80\%,20\%)$ for train and test respectively. In Figure \ref{fig:bnn_test_paper}, we present the test error between \gls*{SVGD-EM} and \gls*{M-SVGD-EM} using two different initializations of $\theta$. In the case $\theta_0=(0,0)$ illustrated in Figure~\ref{fig:bnn_test_0_paper} \gls*{M-SVGD-EM} demonstrates better performances than \gls*{SVGD-EM}. This is also the case in Figure~\ref{fig:bnn_test_2_paper} where $\theta_0=(2,2)$. This last case is included in the study as we observed that the predictive performance is heavily affected by different initializations of $\theta_0=(\alpha_0,\beta_0)$, which can cause the algorithms to converge to local minima.
Similarly, in Figure~\ref{fig:bnn_lppd_paper}(a)--(b), we compare the log predictive probability density with \gls*{M-SVGD-EM} consistently outperforming \gls*{SVGD-EM} across all accelerations. More results can be found in the Appendix \ref{appendix:medianrbf} with $N\in\{5,10,20\}$ showing consistency of \gls*{M-SVGD-EM} to provide better predictive performance in different settings.

The heatmaps in Figure~\ref{fig:bnn_alpha_beta_appendix} show how acceleration behaves with different initializations of $\theta=(\alpha,\beta)$ for a fixed number of iterations $T=500$ and a learning rate $\gamma=0.1$ using Adagrad as an optimizer. In Figures~\ref{fig:bnn_alpha_beta_acc0_appendix}--\ref{fig:bnn_alpha_beta_acc03_appendix}, where acceleration is 0.3 we can see minimal difference between the test errors in \gls*{SVGD-EM} and \gls*{M-SVGD-EM}(0.3), whereas for a high acceleration value \gls*{M-SVGD-EM}(0.9) remains stable and dominates providing a much lower test-error for almost all combinations of $\theta=(\alpha,\beta)$, demonstrating the potential of using a high-acceleration to escape local minima on the parameter estimation.

\section{Conclusion}\label{sec:conclusion}
We have proposed a new method for accelerated \gls*{MMLE} using accelerated \gls*{SVGD} and optimization methods. The resulting method, \gls*{M-SVGD-EM}, results in a significantly faster method in terms of the number of iterations required for convergence. On a number of examples, we have demonstrated the speedup provided by \gls*{M-SVGD-EM} over the standard \gls*{SVGD-EM} method, as well as other state-of-the-art methods such as \gls*{PGD} and \gls*{SOUL}.Furthermore, we compared \gls*{M-SVGD-EM} with the accelerated algorithm \gls*{MPGD} \parencite{lim2023momentum}. On the Toy Hierarchical Model \gls*{M-SVGD-EM} is slower but achieves a lower and more stable MSE, while on Bayesian Logistic Regression it performs competitively, converging in a similar number of iterations. All methods were compared after tuning.

\textbf{Limitations and future work.} Our method is derived using the acceleration ideas from optimization and notably in the space of measures $\Pac$. It accounts for the interactions between the particles, leading to $\mathcal{O}(N^2)$ operations to update the particle cloud. This fact limits the scalability of the method with respect to the number of particles. However, the proposed acceleration method partially addresses this issue by reducing the number of iterations required for convergence by up to 50\% in the conducted experiments. A limitation of the method is it is  based on an approximation introduced in \textcite{liu2019understanding} relying on a heuristic and the theoretical explanation of the performance of our method hence is not provided here. We believe that investigating potential applications of the work in \textcite{stein2025accelerated} to our setting can overcome this problem.

We also highlight potential applications of \gls*{M-SVGD-EM} beyond those considered in this paper. By leveraging existing interacting particle methods and their formulations, our approach can be adapted to accelerate inverse problem solvers \parencite{glyn2025statistical}, to train energy-based generative models \parencite{marks2025learning}, and to train latent diffusion models \parencite{wang2025training}. These directions represent promising avenues for future work.

\textbf{Broader impact.} The proposed method is a novel \gls*{MMLE} scheme for \glspl*{LVM} which find applications in all areas of science. A potential positive societal impact of our method is it provides a significant speedup over existing methods, which can be beneficial in many applications as the computation times could be slashed, resulting in significant savings in terms of computational resources. Like many other training methods for probabilistic machine learning models, our method can be used to train classifiers or generative models which can be used for harmful purposes. However, this risk is not amplified here by our work and is the same as standard algorithms to train probabilistic models.

\section*{Acknowledgements}
Part of this work was completed while A.~R. was visiting O.~D.~A.'s research group in the Department of Mathematics at Imperial College London as an intern. R.~A. is supported by the EPSRC through the Modern Statistics and Statistical Machine Learning (StatML) CDT programme under grant EP/S023151/1. O.~D.~A. is grateful to Department of Statistics, London School of Economics and Political Science (LSE) for the hospitality during the preparation of this work.

\printbibliography

@article{liu1994ecme,
  title={The {ECME} algorithm: a simple extension of {EM} and ECM with faster monotone convergence},
  author={Liu, Chuanhai and Rubin, Donald B},
  journal={Biometrika},
  volume={81},
  number={4},
  pages={633--648},
  year={1994},
  publisher={Oxford University Press}
}

@article{duchi2011adaptive,
  title={Adaptive subgradient methods for online learning and stochastic optimization.},
  author={Duchi, John and Hazan, Elad and Singer, Yoram},
  journal={Journal of machine learning research},
  volume={12},
  number={7},
  year={2011}
}

@article{marks2025learning,
  title={Learning latent energy-based models via interacting particle {L}angevin dynamics},
  author={Marks, Joanna and Wang, Tim YJ and Akyildiz, O Deniz},
  journal={arXiv preprint arXiv:2510.12311},
  year={2025}
}

@misc{breast_cancer_wisconsin_17,
  author = {Wolberg, William and Mangasarian, Olvi and Street, Nick and Street, W.},
  title = {Breast Cancer Wisconsin (Diagnostic)},
  year = {1993},
  howpublished = {UCI Machine Learning Repository},
}

@article{nesterov2005smooth,
  title={Smooth minimization of non-smooth functions},
  author={Nesterov, Yu},
  journal={Mathematical programming},
  volume={103},
  pages={127--152},
  year={2005},
  publisher={Springer}
}

@book{santambrogio2015optimal,
  title={Optimal transport for applied mathematicians},
  author={Santambrogio, Filippo},
  volume={87},
  year={2015},
  publisher={Springer}
}

@inproceedings{xie2020fast,
  title={A fast proximal point method for computing exact wasserstein distance},
  author={Xie, Yujia and Wang, Xiangfeng and Wang, Ruijia and Zha, Hongyuan},
  booktitle={Uncertainty in artificial intelligence},
  pages={433--453},
  year={2020},
  organization={PMLR}
}

@article{cuturi2013sinkhorn,
  title={Sinkhorn distances: Lightspeed computation of optimal transport},
  author={Cuturi, Marco},
  journal={Advances in neural information processing systems},
  volume={26},
  year={2013}
}

@inproceedings{pele2009fast,
  title={Fast and robust earth mover's distances},
  author={Pele, Ofir and Werman, Michael},
  booktitle={2009 IEEE 12th international conference on computer vision},
  pages={460--467},
  year={2009},
  organization={IEEE}
}

@misc{jax2018github,
  author = {Bradbury, James and Frostig, Roy and Hawkins, Peter and Johnson, Matthew James
            and Leary, Chris and Maclaurin, Dougal and Necula, George and Paszke, Adam
            and VanderPlas, Jake and Wanderman-Milne, Skye and Zhang, Qiao},
  title = {{JAX}: Composable transformations of {P}ython+{N}um{P}y programs},
  note = {Version 0.3.13},
  year = {2018}
}

@book{nesterov2013introductory,
  title={Introductory lectures on convex optimization: A basic course},
  author={Nesterov, Yurii},
  volume={87},
  year={2013},
  publisher={Springer Science \& Business Media}
}

@article{nesterov1988approach,
  title={On an approach to the construction of optimal methods of minimization of smooth convex functions},
  author={Nesterov, Yurii},
  journal={Ekonomika i Mateaticheskie Metody},
  volume={24},
  number={3},
  pages={509--517},
  year={1988}
}

@article{de2021efficient,
  title={Efficient stochastic optimisation by unadjusted {L}angevin {M}onte {C}arlo: {A}pplication to maximum marginal likelihood and empirical {B}ayesian estimation},
  author={De Bortoli, Valentin and Durmus, Alain and Pereyra, Marcelo and Vidal, Ana F},
  journal={Statistics and Computing},
  volume={31},
  pages={1--18},
  year={2021},
  publisher={Springer}
}

@article{wei1990monte,
  title={A {M}onte {C}arlo implementation of the {EM} algorithm and the poor man's data augmentation algorithms},
  author={Wei, Greg CG and Tanner, Martin A},
  journal={Journal of the American statistical Association},
  volume={85},
  number={411},
  pages={699--704},
  year={1990},
  publisher={Taylor \& Francis}
}

@article{meng1993maximum,
  title={Maximum likelihood estimation via the {ECM} algorithm: A general framework},
  author={Meng, Xiao-Li and Rubin, Donald B},
  journal={Biometrika},
  volume={80},
  number={2},
  pages={267--278},
  year={1993},
  publisher={Oxford University Press}
}

@article{lange1995gradient,
  title={A gradient algorithm locally equivalent to the {EM} algorithm},
  author={Lange, Kenneth},
  journal={Journal of the Royal Statistical Society: Series B (Methodological)},
  volume={57},
  number={2},
  pages={425--437},
  year={1995},
  publisher={Wiley Online Library}
}

@inproceedings{nesterov1983method,
  title={A method for unconstrained convex minimization problem with the rate of convergence ${O}(1/k^2)$},
  author={Nesterov, Yurii},
  booktitle={Dokl. Akad. Nauk. SSSR},
  volume={269},
  pages={543},
  year={1983}
}

@inproceedings{zhang2018estimate,
  title={An estimate sequence for geodesically convex optimization},
  author={Zhang, Hongyi and Sra, Suvrit},
  booktitle={Conference On Learning Theory},
  pages={1703--1723},
  year={2018},
  organization={PMLR}
}

@article{whiteley2022statistical,
 title={Statistical exploration of the Manifold Hypothesis},
  author={Whiteley, Nick and Gray, Annie and Rubin-Delanchy, Patrick},
  journal={Journal of the Royal Statistical Society: Series B},
  year={2025}
}

@inproceedings{lim2023momentum,
  title={Momentum particle maximum likelihood},
  author={Lim, JN and Kuntz, J and Power, S and Johansen, Adam M},
  booktitle={Proceedings of 41st International Conference on Machine Learning (ICML)},
  volume={235},
  pages={29816--29871},
  year={2024}
}

@inproceedings{kuntz2023particle,
  title={Particle algorithms for maximum likelihood training of latent variable models},
  author={Kuntz, Juan and Lim, Jen Ning and Johansen, Adam M},
  booktitle={International Conference on Artificial Intelligence and Statistics},
  pages={5134--5180},
  year={2023},
  organization={PMLR}
}

@incollection{neal1998view,
  title={A view of the {EM} algorithm that justifies incremental, sparse, and other variants},
  author={Neal, Radford M and Hinton, Geoffrey E},
  booktitle={Learning in graphical models},
  pages={355--368},
  year={1998},
  publisher={Springer}
}

@article{oliva2024kinetic,
  title={Kinetic interacting particle {L}angevin {M}onte {C}arlo},
  author={Oliva, Paul Felix Valsecchi and Akyildiz, O Deniz},
  journal={arXiv preprint arXiv:2407.05790},
  year={2024}
}

@article{caprio2024error,
  title={Error bounds for particle gradient descent, and extensions of the log-Sobolev and Talagrand inequalities},
  author={Caprio, Rocco and Kuntz, Juan and Power, Samuel and Johansen, Adam M},
  journal={Journal of Machine Learning Research},
  volume={26},
  number={103},
  pages={1--38},
  year={2025}
}

@article{encinar2025proximal,
  title={Proximal Interacting Particle {L}angevin Algorithms},
  author={Encinar, Paula Cordero and Crucinio, Francesca R and Akyildiz, O Deniz},
  journal={Uncertainty in Artificial Intelligence (UAI)},
  year={2025}
}

@article{akyildiz2025interacting,
  title={Interacting Particle {L}angevin Algorithm for Maximum Marginal Likelihood Estimation},
  author={Akyildiz, {\"O}. Deniz and Crucinio, Francesca Romana and Girolami, Mark and Johnston, Tim and Sabanis, Sotirios},
  journal={ESAIM: Probability and Statistics},
  year={2025}
}

@inproceedings{sharrock2024tuning,
  title={Tuning-Free Maximum Likelihood Training of Latent Variable Models via Coin Betting},
  author={Sharrock, Louis and Dodd, Daniel and Nemeth, Christopher},
  booktitle={International Conference on Artificial Intelligence and Statistics},
  pages={1810--1818},
  year={2024},
  organization={PMLR}
}

@article{dempster1977maximum,
  title={Maximum likelihood from incomplete data via the {EM} algorithm},
  author={Dempster, Arthur P and Laird, Nan M and Rubin, Donald B},
  journal={Journal of the Royal Statistical Society: Series B (Methodological)},
  volume={39},
  number={1},
  pages={1--22},
  year={1977},
  publisher={Wiley Online Library}
}

@inproceedings{liu2019understanding,
  title={Understanding and accelerating particle-based variational inference},
  author={Liu, Chang and Zhuo, Jingwei and Cheng, Pengyu and Zhang, Ruiyi and Zhu, Jun},
  booktitle={International Conference on Machine Learning},
  pages={4082--4092},
  year={2019},
  organization={PMLR}
}

@article{liu2016stein,
  title={Stein variational gradient descent: A general purpose bayesian inference algorithm},
  author={Liu, Qiang and Wang, Dilin},
  journal={Advances in neural information processing systems},
  volume={29},
  year={2016}
}

@book{villani2009optimal,
  title={Optimal transport: old and new},
  author={Villani, C{\'e}dric and others},
  volume={338},
  year={2009},
  publisher={Springer}
}

@article{glyn2025statistical,
  title={Statistical finite elements via interacting particle Langevin dynamics},
  author={Glyn-Davies, Alex and Duffin, Connor and Kazlauskaite, Ieva and Girolami, Mark and Akyildiz, {\"O} Deniz},
  journal={SIAM/ASA Journal on Uncertainty Quantification},
  volume={13},
  number={3},
  pages={1200--1227},
  year={2025},
  publisher={SIAM}
}

@article{wang2025training,
  title={Training Latent Diffusion Models with Interacting Particle Algorithms},
  author={Wang, Tim YJ and Kuntz, Juan and Akyildiz, O Deniz},
  journal={arXiv preprint arXiv:2505.12412},
  year={2025}
}

@book{ambrosio2008gradient,
  title={Gradient flows: in metric spaces and in the space of probability measures},
  author={Ambrosio, Luigi and Gigli, Nicola and Savar{\'e}, Giuseppe},''
  year={2008},
  publisher={Springer Science \& Business Media}
}

@article{otto2001geometry,
  title={The geometry of dissipative evolution equations: the porous medium equation},
  author={Otto, Felix},
  year={2001},
  publisher={Taylor \& Francis}
}

@article{attia2021algorithmic,
  title={Algorithmic instabilities of accelerated gradient descent},
  author={Attia, Amit and Koren, Tomer},
  journal={Advances in Neural Information Processing Systems},
  volume={34},
  pages={1204--1214},
  year={2021}
}

@inproceedings{stein2025accelerated,
  title={Accelerated Stein variational gradient flow},
  author={Stein, Viktor and Li, Wuchen},
  booktitle={International Conference on Geometric Science of Information},
  pages={80--90},
  year={2025},
  organization={Springer}
}

\clearpage
\appendix
\thispagestyle{empty}
\onecolumn
\aistatstitle{Momentum SVGD-EM for Accelerated Maximum Marginal Likelihood Estimation \\
Supplementary Material}
\section{Derivation of SVGD-WNes from SVGD-RAGD}
\label{appendix: SVGD-WNes}
\subsection{Practical tools for the derivation}
In the following, we derive \gls*{SVGD-WNes} from the \gls*{SVGD-RAGD}. To facilitate this derivation, we introduce a set of practical tools that make the exponential map and its inverse tractable and composable:
\begin{itemize}
    \item Let $\{ \tilde{x}^{(i)}\}_{1 \leq i \leq N}$  be a set of particles sampled from  $\tilde{q}\in \Pac$, then the particle approximation of the exponential map at $q$ defined by $\text{Exp}_{\tilde{q}}(v) := (\text{id} + v)_{\#}\,\tilde{q}$ where $v \in \mathcal{T}_{\tilde{q}}\Pac$ is the application of the map $x \mapsto x + v(x)$ on each particle \parencite{liu2019understanding}, thus 
    \begin{equation}
        \label{eq:exp_particles}
        \text{Exp}_{\tilde{q}}(v)(\tilde{x}^{(i)}) = \tilde{x}^{(i)} + v(\tilde{x}^{(i)}).
    \end{equation}
    \item Consider the density $q \approx \sum_i^N \delta_{x^{(i)}}$ and the momentum density $\tilde{q} \approx \sum_i^N\delta_{\tilde{x}^{(i)}}$. The pairwise closeness assumption between $\{ x^{(i)} \}_{1\leq i \leq N} \text{ and } \{ \tilde{x}^{(i)} \}_{1\leq i \leq N}$ is $d(x^{(i)}, \tilde{x}^{(i)}) \ll \min \left\{ \min_{j \ne i} d(x^{(i)}, x^{(j)}), 
    \min_{j \ne i} d(\tilde{x}^{(i)}, \tilde{x}^{(j)}) \right\}$ where $d$ is the euclidean distance in $\mathcal{X}$. As shown in  \textcite{liu2019understanding}, this leads to 
    \begin{equation}
        \label{eq:closness_assump_appendix}
        \text{Exp}_{\tilde{q}}^{-1}(q)(\tilde{x}^{(i)}) \approx x^{(i)} - \tilde{x}^{(i)}  \; \text{with} \; 1 \leq i \leq N,
    \end{equation}
    where $x^{(i)} - \tilde{x}^{(i)} = v(\tilde{x}^{(i)})$ with $v(x) := x^{(i)} - x$. In the derivation, we use the assumption \eqref{eq:closness_assump_appendix} on the particles sampled from the pairs $(\tilde{q}_{t-1}, q_t)$ and  $(\tilde{q}_{t-1}, q_{t-1})$.
    \item Finally we will also encounter expressions of the general form $\text{Exp}_{q} (v(x^{(i)}))$ where the approximation resulting from the usage of \eqref{eq:closness_assump_appendix} serves as the input of the exponential map. In these cases we simply use \eqref{eq:exp_particles} and deduce 
    \begin{equation}
        \label{eq:notation_trick}
        \text{Exp}_{q}(v(x^{(i)})) = x^{(i)} + v(x^{(i)}).
    \end{equation}
\end{itemize}
Before proceeding with the derivation, we highlight that the particle-level versions of the exponential map and its inverse yield an analogue of the identity $\text{Exp}_{\tilde{q}}(\text{Exp}_{\tilde{q}}^{-1}(q)) = q$. Specifically, with $\; 1 \leq i \leq N \quad$ we have:
\begin{align}
    \text{Exp}_{\tilde{q}}(\text{Exp}_{\tilde{q}}^{-1}(q)(\tilde{x}^{(i)})) &= \tilde{x}^{(i)} + \text{Exp}_{\tilde{q}}^{-1}(q)(\tilde{x}^{(i)}) \quad \text{using} \; \eqref{eq:exp_particles} \nonumber\\
    & \approx \tilde{x}^{(i)} + x^{(i)} - \tilde{x}^{(i)}  \; \text{using} \;  \eqref{eq:closness_assump_appendix}  \nonumber \\
    &= x^{(i)}, \nonumber 
\end{align}
where we emphasize that the inverse exponential map is taken at $\tilde{q}$ and thus evaluated at the particle $\tilde{x}^{(i)}$.
\subsection{Derivation}
We are now equipped to derive one iteration of \gls*{SVGD-WNes} from the corresponding iteration of \gls*{SVGD-RAGD} given by:
\begin{align*}
    q_{t+1} &= \text{Exp}_{\tilde{q}_{t}}(-\gamma \WG_{H_k} \mathcal{F}(\theta, \tilde{q}_{t})) \\
    \tilde{q}_{t+1} &= \text{Exp}_{q_{t+1}} \{ c_1 \text{Exp}_{q_{t+1}}^{-1} [ \text{Exp}_{\tilde{q}_{t}} ( \text{Exp}_{\tilde{q}_{t}}^{-1}(q_{t+1}) +(c_2-1)(\text{Exp}_{\tilde{q}_{t}}^{-1}(q_{t+1}) - \text{Exp}_{\tilde{q}_{t}}^{-1}(q_{t})))] \}.
\end{align*}
Letting $q_t \approx \sum_i^N \delta_{x_t^{(i)}}$ and $\tilde{q}_t \approx \sum_i^N \delta_{\tilde{x}_t^{(i)}}$, we obtain the particle-level formulation of \gls*{SVGD-RAGD} :
\begin{align}
    x_{t+1}^{(i)} &= \text{Exp}_{\tilde{q}_{t}}(-\gamma \WG_{H_k} \mathcal{F}(\theta, \tilde{q}_{t}))(\tilde{x}_t^{(i)}) \label{eq:RAGD_1_particle}\\
    \tilde{x}_{t+1}^{(i)} &= \text{Exp}_{q_{t+1}} \{ c_1 \text{Exp}_{q_{t+1}}^{-1} [\text{Exp}_{\tilde{q}_{t}} ( \text{Exp}_{\tilde{q}_{t}}^{-1}(q_{t+1})(\tilde{x}_t^{(i)}) + (c_2-1)(\text{Exp}_{\tilde{q}_{t}}^{-1}(q_{t+1})(\tilde{x}_t^{(i)}) -\text{Exp}_{\tilde{q}_{t}}^{-1}(q_{t})(\tilde{x}_t^{(i)}))))] \}. \label{eq:RAGD_2_particle}
\end{align}
Recalling \eqref{eq:exp_particles}, we can rewrite \eqref{eq:RAGD_1_particle} as:
\begin{align*}
    x_{t+1}^{(i)} &= \tilde{x}_t^{(i)} - \gamma \WG_{H_k} \mathcal{F}(\theta, \tilde{q}_{t})(\tilde{x}_t^{(i)}) \\
    &= \tilde{x}_{t}^{(i)} + \gamma \frac{1}{N}\sum_{j=1}^{N}\left[ k(\tilde{x}_{t}^{(j)},\tilde{x}_{t}^{(i)})\nabla_x \ell(\theta,\tilde{x}_{t}^{(j)}) +\nabla_{1}k(\tilde{x}_t^{(j)},\tilde{x}_{t}^{(i)})\right],
\end{align*}
which corresponds to \eqref{eq:svgd-wnes-x}. Next we use \eqref{eq:closness_assump_appendix} in \eqref{eq:RAGD_2_particle}, yielding:
\begin{align*}
    \tilde{x}_{t+1}^{(i)} = \text{Exp}_{q_{t+1}} \{ c_1 \text{Exp}_{q_{t+1}}^{-1} [\text{Exp}_{\tilde{q}_{t}} ( x^{(i)}_{t+1} - \tilde{x}^{(i)}_{t} + (c_2-1)(x^{(i)}_{t+1} - x^{(i)}_{t}))]\}.
\end{align*}
This corresponds to the setting in \eqref{eq:notation_trick}, where the velocity vector field is given by $v_1(x)=x^{(i)}_{t+1} -x + (c_2-1)(x^{(i)}_{t+1} - x^{(i)}_{t}) $. Hence, we obtain:
\begin{equation}
    \tilde{x}_{t+1}^{(i)} = \text{Exp}_{q_{t+1}} \{ c_1 \text{Exp}_{q_{t+1}}^{-1} [ x^{(i)}_{t+1} + (c_2-1)(x^{(i)}_{t+1} - x^{(i)}_{t})]\} \label{eq:last_tool_2}
\end{equation}
Applying \eqref{eq:closness_assump_appendix} in \eqref{eq:last_tool_2} and then \eqref{eq:notation_trick} in \eqref{eq:end_derivation}, we further simplify: 
\begin{align}
    \tilde{x}_{t+1}^{(i)} &= \text{Exp}_{q_{t+1}} \{ c_1 [x^{(i)}_{t+1} + (c_2-1)(x^{(i)}_{t+1} - x^{(i)}_{t}) - x^{(i)}_{t+1}]\} \nonumber\\
    &= \text{Exp}_{q_{t+1}} \{ c_1(c_2-1)(x^{(i)}_{t+1} - x^{(i)}_{t})\} \label{eq:end_derivation}\\
    &= x^{(i)}_{t+1}+ c_1(c_2-1)(x^{(i)}_{t+1} - x^{(i)}_{t}) , \nonumber
\end{align}
where \eqref{eq:notation_trick} is applied with the velocity vector field $v_2(x) = c_1(c_2-1)(x - x^{(i)}_{t})$. We thus obtain the final \gls*{SVGD-WNes} update:
\begin{align*}
x_{t+1}^{(i)} &= \tilde{x}_{t}^{(i)} + \gamma \frac{1}{N}\sum_{j=1}^{N} k(\tilde{x}_{t}^{(j)},\tilde{x}_{t}^{(i)})\nabla_x \ell(\theta, \tilde{x}_{t}^{(j)}) + \nabla_{1}k(\tilde{x}_t^{(j)},\tilde{x}_{t}^{(i)})\\
\tilde{x}_{t+1}^{(i)} &= x_{t+1}^{(i)} + c_1 (c_2 - 1) (x_{t+1}^{(i)} - x_{t}^{(i)})
\end{align*}
\section{Results with MedianRBF}
We reproduce the same results presented in the main paper using the MedianRBF kernel instead of the AutoRBF kernel.
\label{appendix:medianrbf}
\subsection{Toy Hierarchical Example MedianRBF}
\begin{figure}[h!]
  \centering
  \begin{subfigure}[h]{0.32\textwidth}
    \includegraphics[height=3.2cm]{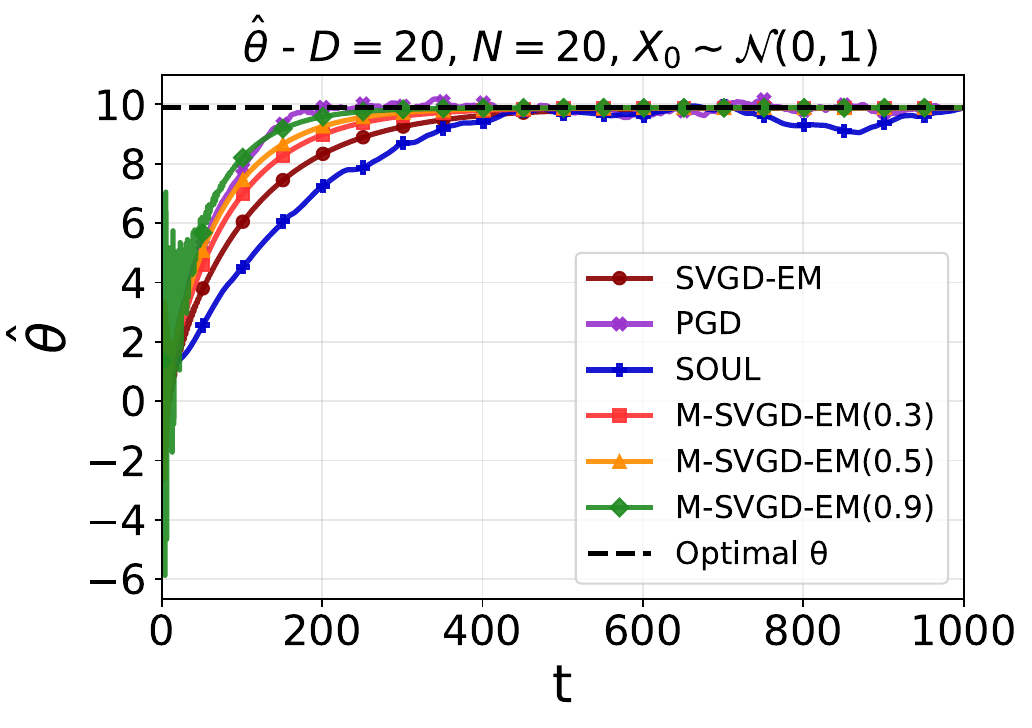}
    \caption{Parameter Estimation}
    \label{fig:toy_param_estimates_appendix_medianrbf}
  \end{subfigure}
  \hfill
  \begin{subfigure}[h]{0.32\textwidth}
    \includegraphics[height=3.2cm]{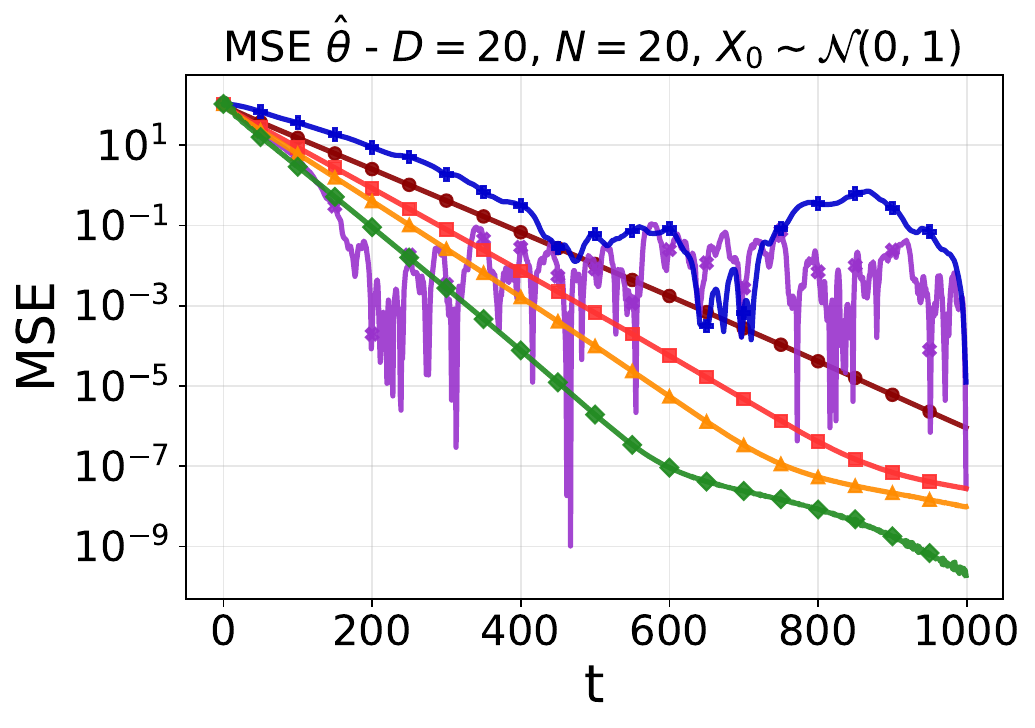}
    \caption{MSE($\hat{\theta})$ vs t}
    \label{fig:toy_mse_appendix_medianrbf}
  \end{subfigure}
  \hfill
  \begin{subfigure}[h]{0.32\textwidth}
    \includegraphics[height=3.2cm]{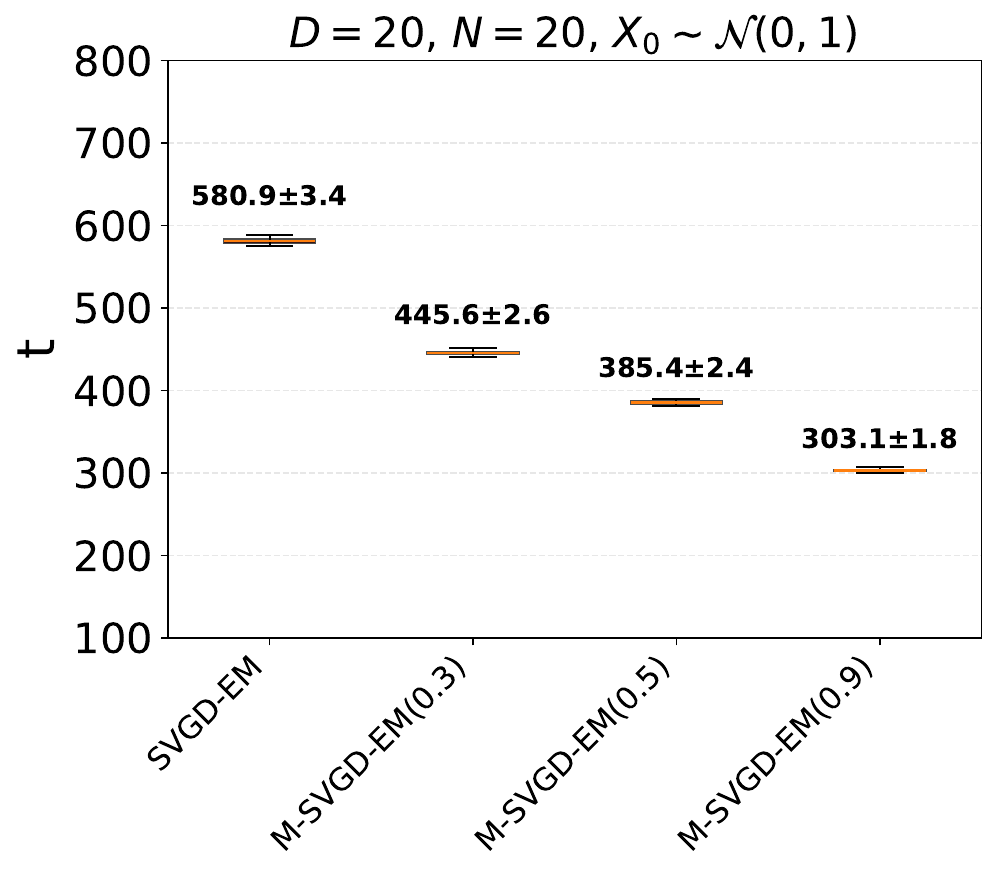}
    \caption{Average Iterations}
    \label{fig:toy_boxplot_appendix_medianrbf}
  \end{subfigure}
  \caption{Comparison between \gls*{M-SVGD-EM} and \gls*{SVGD-EM}. Part (a) shows the performance on parameter estimation on a single run on the ToyHM(10, 12) with different acceleration parameters $\alpha_{\theta}=\alpha_{X}=(0.3,0.5,0.9)$. Part (b) displays MSE relative to the empirical minimizer for each algorithm. (c) shows the average number of iterations until convergence, defined as being within a threshold of 0.05 from the empirical minimizer. The results are averaged over 20 independent trials.}
  \label{fig:four_in_a_row}
\end{figure}

\begin{figure}[h]
    \begin{subfigure}{0.47\textwidth}
    
            \begin{flushleft}
                \includegraphics[width=\textwidth]{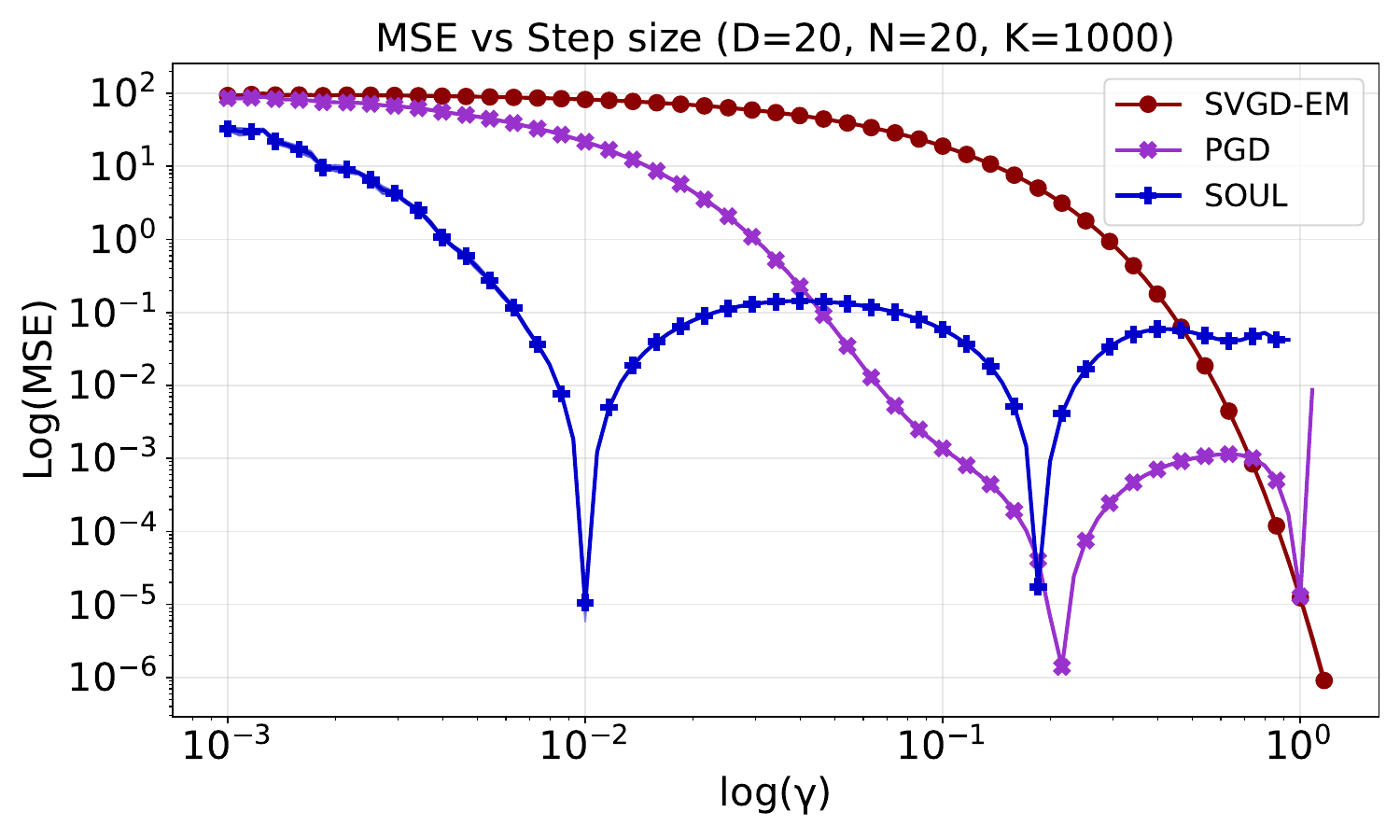}
                \caption{Log-Log Plot MSE vs LR}
                \label{fig:LR_TOY_1}
            \end{flushleft}
    \end{subfigure}
        \hfill
    \begin{subfigure}{0.47\textwidth}
    
            \begin{flushright}
                \includegraphics[width=\textwidth]{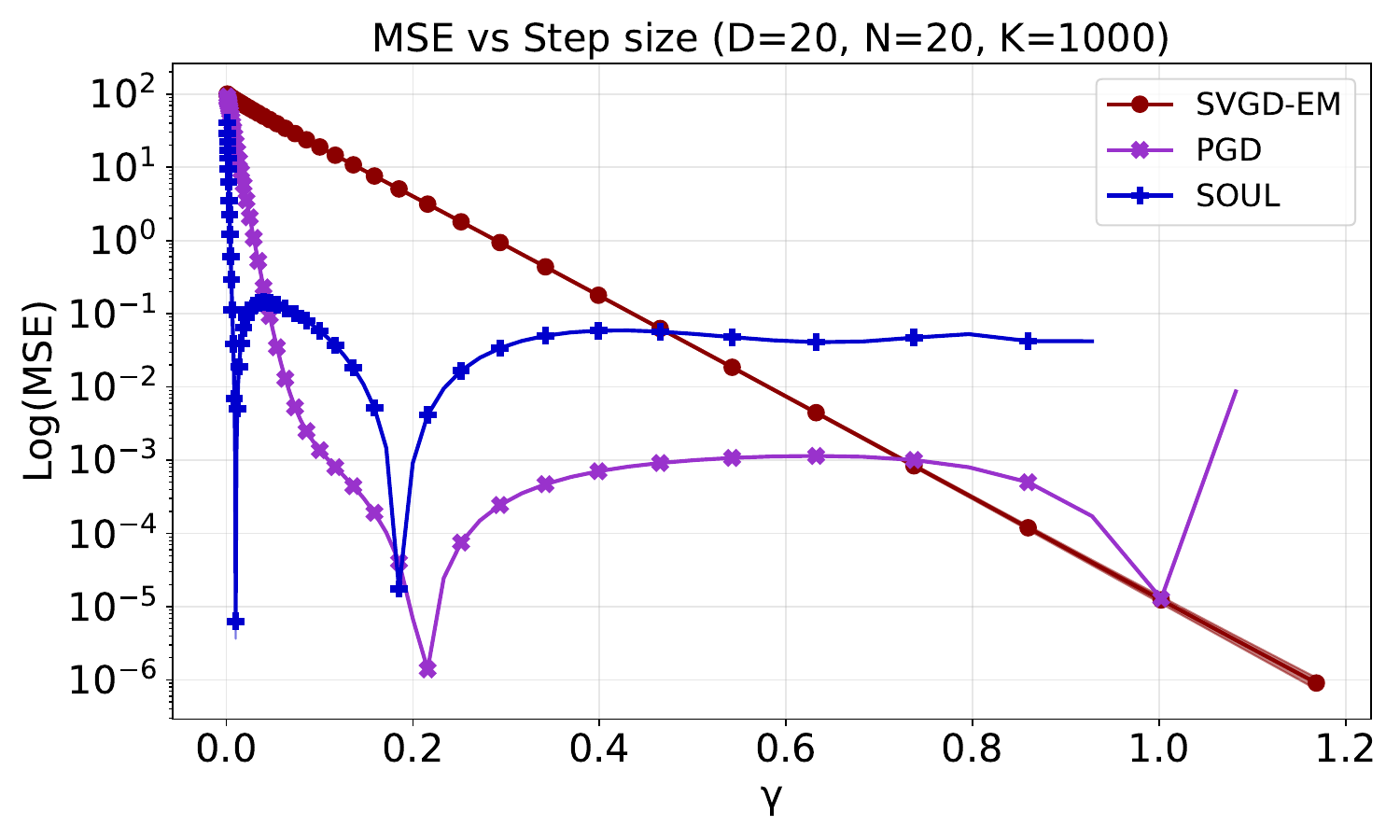}
                \caption{Log-Log Plot MSE vs LR}
                \label{fig:LR_TOY_2}
            \end{flushright}
    \end{subfigure}
    \caption{MSE vs Learning Rate}
\end{figure}

In Figure~\ref{fig:toy_param_estimates_appendix_medianrbf}, we observe that \gls*{M-SVGD-EM} consistently outperforms \gls*{SVGD-EM} across different acceleration parameters $\alpha = \alpha_{\theta}=\alpha_{X}$. As shown in Figure~\ref{fig:toy_mse_appendix_medianrbf}, the mean squared error (MSE) decreases for all values of $\alpha$, with $\alpha = 0.9$ achieving the lowest MSE within $T = 1000$ iterations. Similarly, Figure \ref{fig:toy_boxplot_appendix_medianrbf} demonstrates substantial computational savings with $\alpha = 0.9$ achieving again  achieving nearly a 50\% reduction in iterations. Although the average number of iterations is higher than in Figure~\ref{fig:toy_boxplot_paper}, the variance is significantly lower. These results are averaged over 20 independent trials, using standard gradient descent. For \gls{SOUL} and \gls*{PGD}, we use optimal learning rates $\gamma = (0.01, 0.2)$ (see Figures~\ref{fig:LR_TOY_1}--\ref{fig:LR_TOY_2}), while for both \gls*{SVGD-EM} and \gls*{M-SVGD-EM}, we set $\gamma = 1.15$.

\clearpage
\subsection{Bayesian Logistic Regression MedianRBF}
\label{sec:appendx_blr_medianrbf}
\begin{figure}[H]
  \centering
  \begin{subfigure}[h]{0.32\textwidth}
    \includegraphics[height=3.3cm]{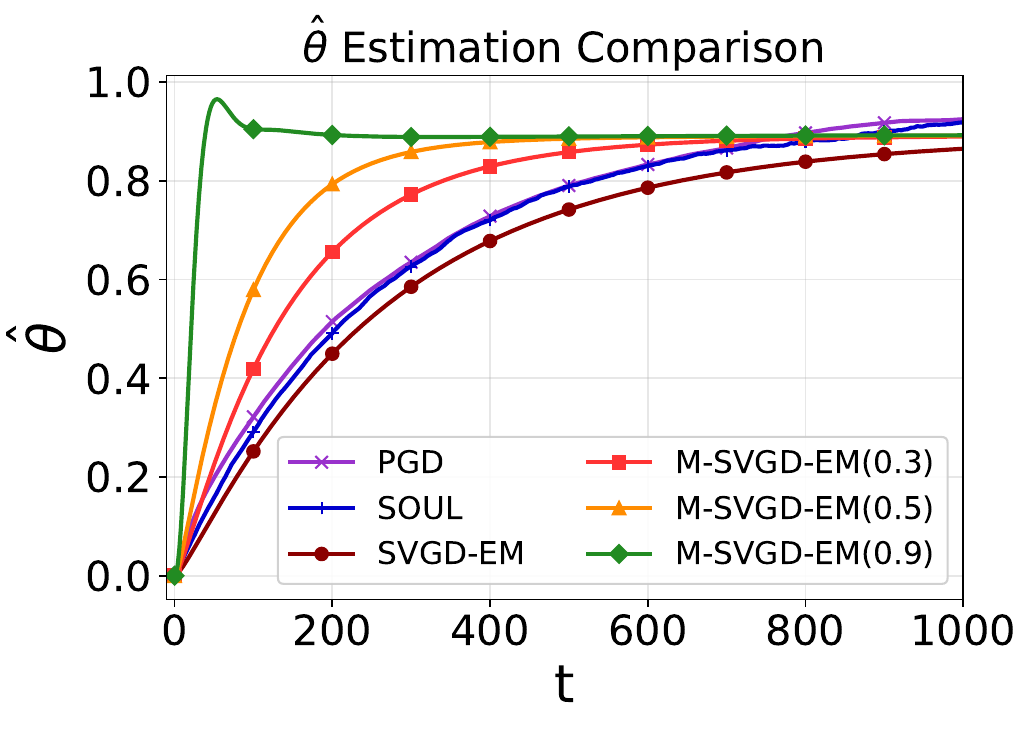}
    \caption{Parameter Estimation}
    \label{fig:blr_para_appendix_medianrbf}
  \end{subfigure}
  \hfill
  \begin{subfigure}[h]{0.32\textwidth}
    \includegraphics[height=3.3cm]{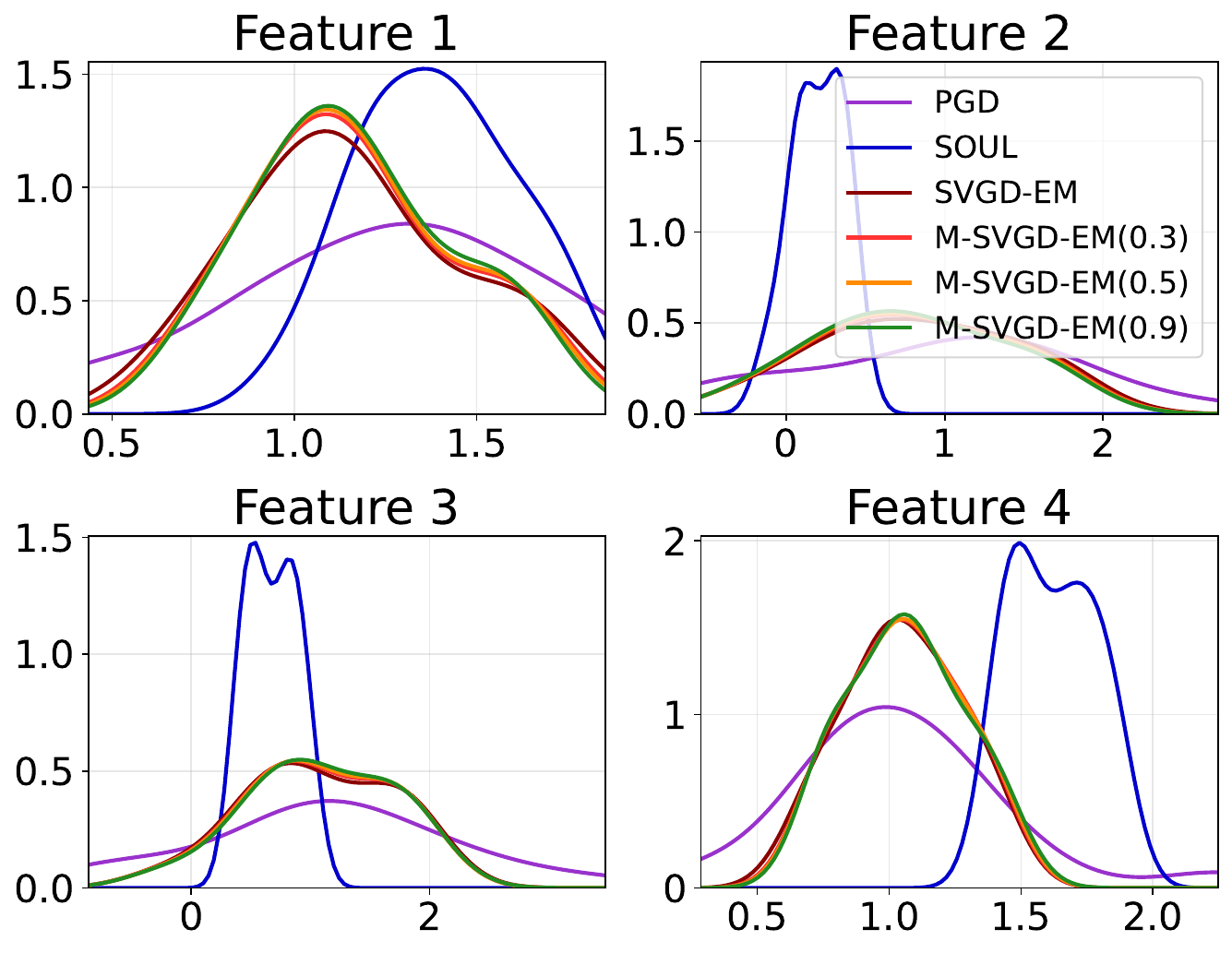}
    \caption{Posterior Estimates}
    \label{fig:blr_features_appendix_medianrbf}
  \end{subfigure}
  \hfill
  \begin{subfigure}[h]{0.32\textwidth}
    \includegraphics[height=3.3cm]{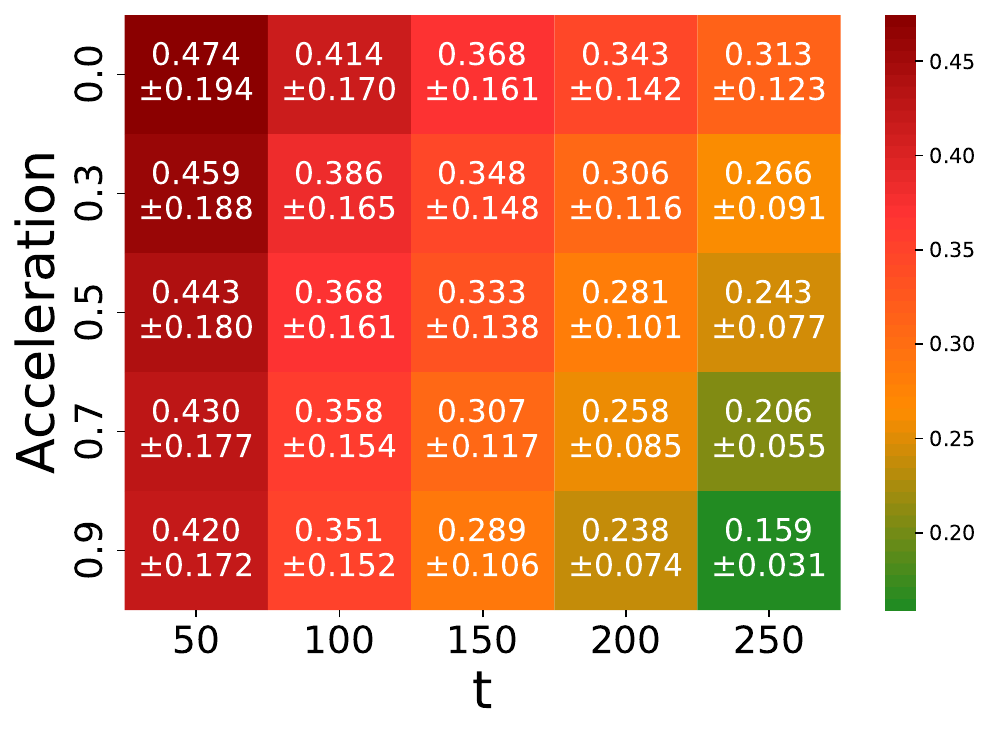}
    \caption{Test Error Heatmap}
    \label{fig:blr_heat_appendix_medianrbf}
  \end{subfigure}
    \caption{(a) Convergence of parameter estimates on BLR under varying acceleration parameters. (b) Posterior distributions across 4 test features. (c) Test error as a function of the acceleration parameter for different values of $t$, showing consistent error reduction with increased acceleration.}
  \label{fig:blr_appendix_medianrbf}
\end{figure}

\begin{figure}[hb]
\centering
\begin{subfigure}{0.32\textwidth}
    \includegraphics[width=\linewidth, height=3.3cm]{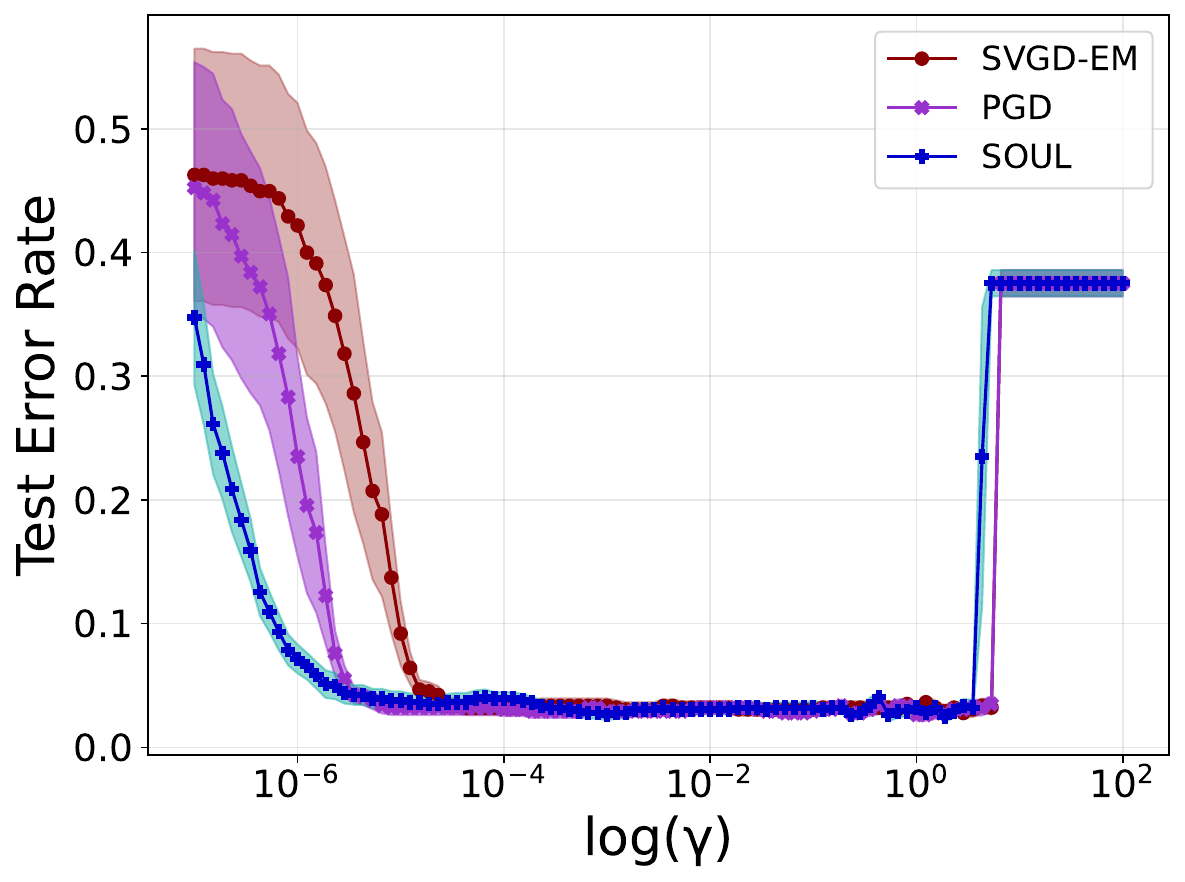}
    \caption{$N=10$}
    \label{fig:blr_test_N10_appendix_medianrbf}
\end{subfigure}
\hfill
\begin{subfigure}{0.32\textwidth}
    \includegraphics[width=\linewidth, height=3.3cm]{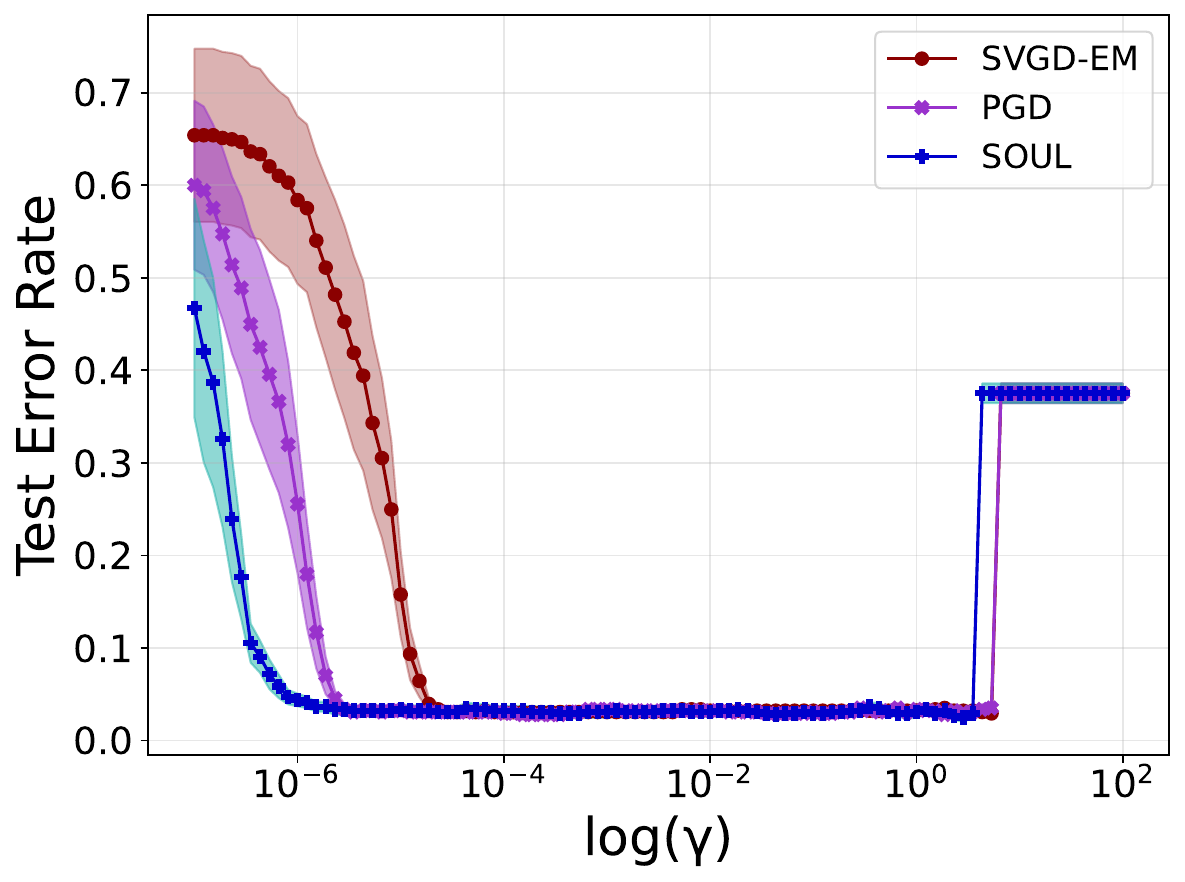}
    \caption{$N=20$}
    \label{fig:blr_test_N20_appendix_medianrbf}
\end{subfigure}
\hfill
\begin{subfigure}{0.32\textwidth}
    \includegraphics[width=\linewidth, height=3.3cm]{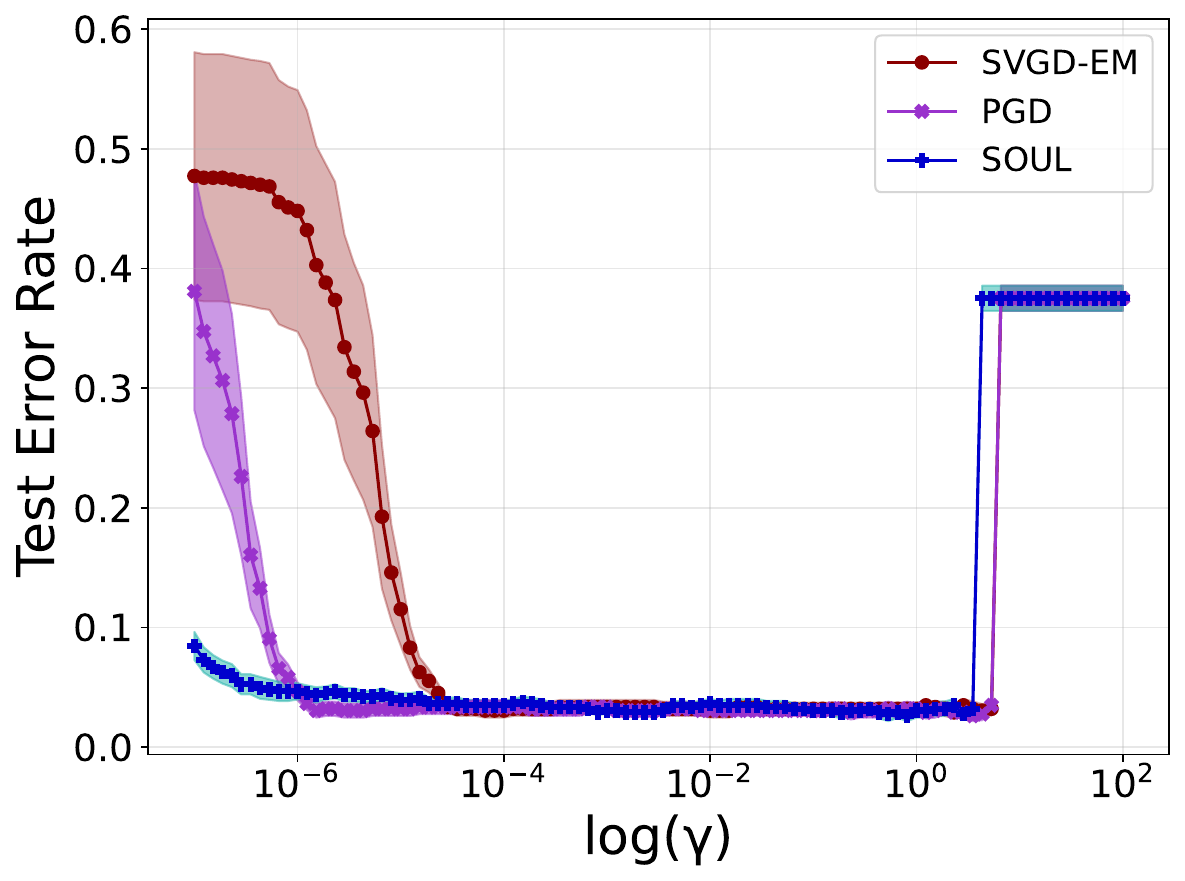}
    \caption{$N=100$}
    \label{fig:blr_test_N100_appendix_medianrbf}
\end{subfigure}
\caption{Test Error (Accuracy) for Bayesian Logistic Regression with different numbers of particles.}
\label{fig:blr_test_appendix_medianrbf}
\end{figure}

In Figure~\ref{fig:blr_appendix_medianrbf}, we again present the results on BLR using standard gradient descent with $N=20$ particles, a step-size of $\gamma=0.01$ for all methods determined by a fine-grid search over $100$ logarithmically spaced ranged from $\gamma\in[10^{-7}, 10^2]$ and averaged over 5 distinct seeds for a fixed number of iterations $T=500$. As shown in Figure ~\ref{fig:blr_test_N20_appendix_medianrbf} the value of $\gamma$ was chosen based on test error performance. Figure~\ref{fig:blr_features_appendix_medianrbf} presents the kernel density estimates (KDEs) of their final particle distributions for the the first 4 features. In comparison to using the AutoRBF in Figure~\ref{fig:blr_features_paper}, the posterior estimates here overlap between accelerations and are not significantly different from each other. In Figure~\ref{fig:blr_heat_appendix_medianrbf}, we demonstrate the average test error over 20 independent train $(80\%)$ and test $(20\%)$ splits using a step size $\gamma=10^{-5}$ which is again within the optimal range as shown in Figure~\ref{fig:blr_test_N20_appendix_medianrbf}. Using MedianRBF takes longer (Figure~\ref{fig:blr_heat_appendix_medianrbf}) to reach a low test error compared to using the AutoRBF, as shown in Figure~\ref{fig:blr_heat_paper}. This highlights also the effect of kernels for convergence. Overall, we still see acceleration outperforms standard \gls*{SVGD-EM} when considering $N=20$ particles, over $T=500$ iterations. The higher the acceleration the more rigorously the test error drops achieving a better performance with less iterations.

\subsection{Bayesian Neural Network}
\label{sec:appendix_bnn_medianrbf}
\begin{figure}[H]
  \centering
  \begin{subfigure}{0.45\textwidth}
    \includegraphics[height=5cm]{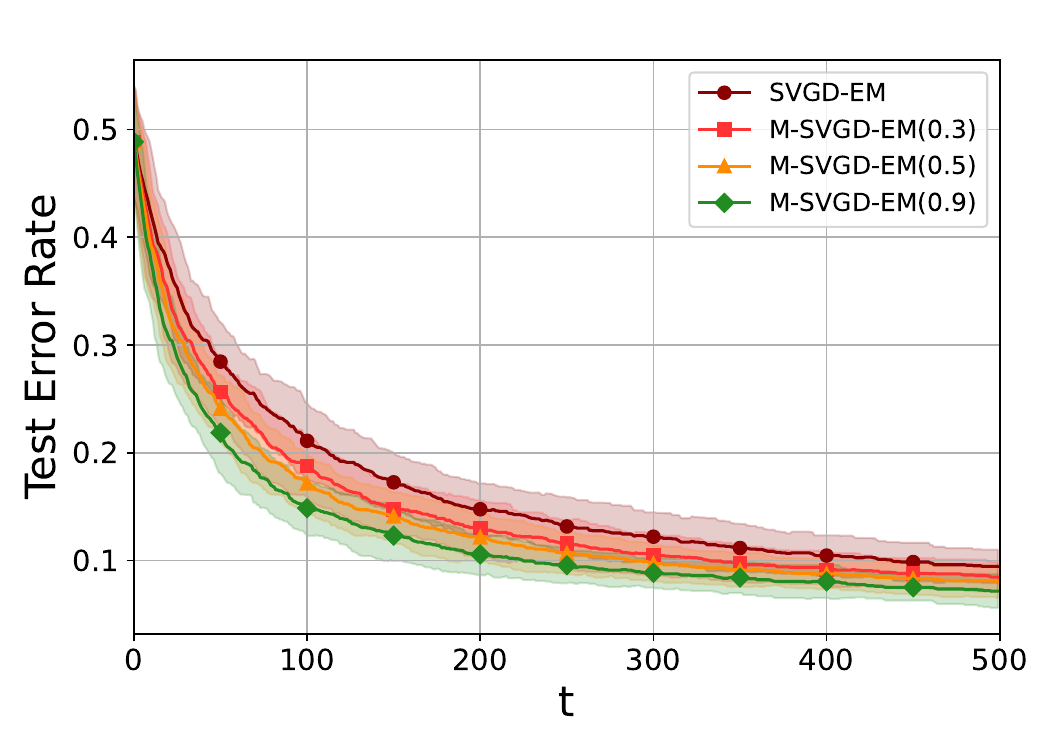}
    \caption{Test Error vs. iterations $t$}
    \label{fig:bnn_test_appendix_medianrbf}
  \end{subfigure}
  \hfill
  \begin{subfigure}{0.45\textwidth}
    \includegraphics[height=5cm]{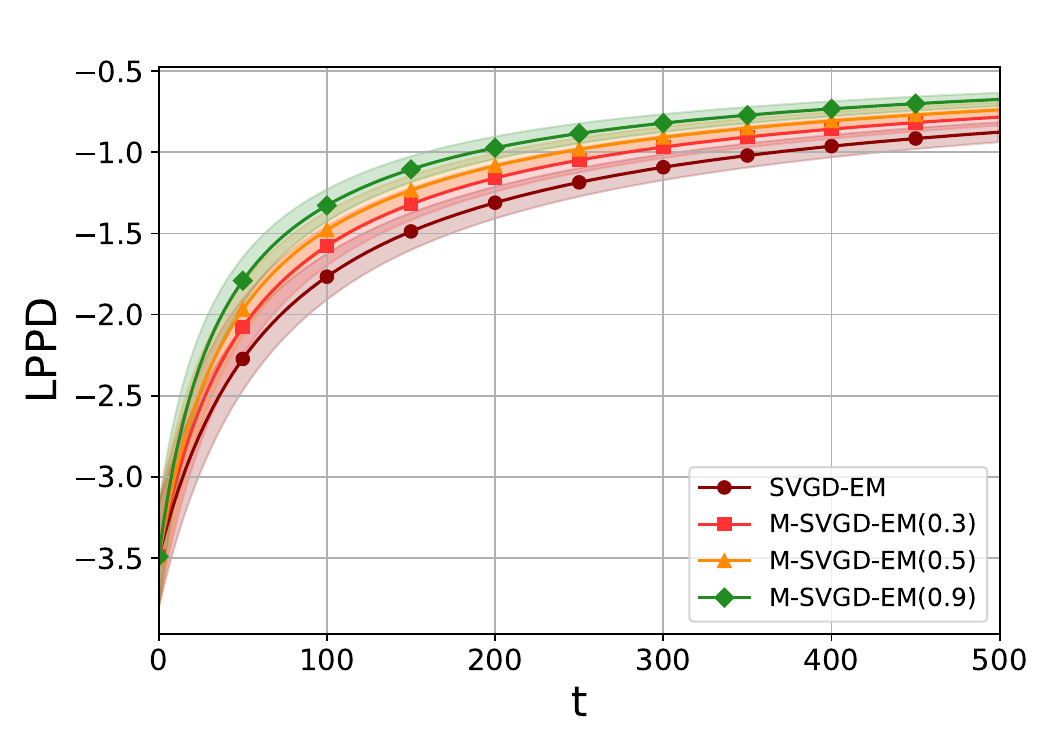}
    \caption{Log Predictive Probability Density (LPPD)}
    \label{fig:bnn_lppd_appendix_medianrbf}
  \end{subfigure}
  \caption{Performance metrics for the BNN on MNIST dataset: (a) Test Error across iterations $t$, and (b) Log Predictive Probability Density (LPPD).}
  \label{fig:BNN}
\end{figure}

\begin{figure}[h!]
\centering
    \includegraphics[width=\linewidth, height=4.0cm]{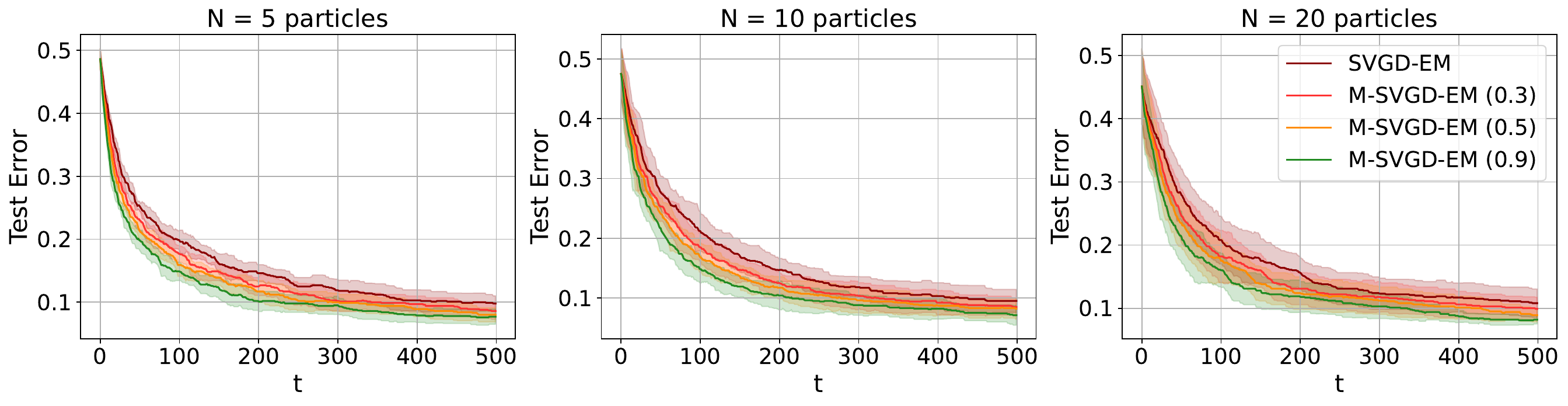}
    \caption{Test Error using $N={5,10,20}$ with initial $\alpha=\beta=0.0$.}
    \label{fig:bnn_5_10_20_test_medianrbf_0}
\end{figure}

\begin{figure}[h!]
\centering
    \includegraphics[width=\linewidth, height=4.0cm]{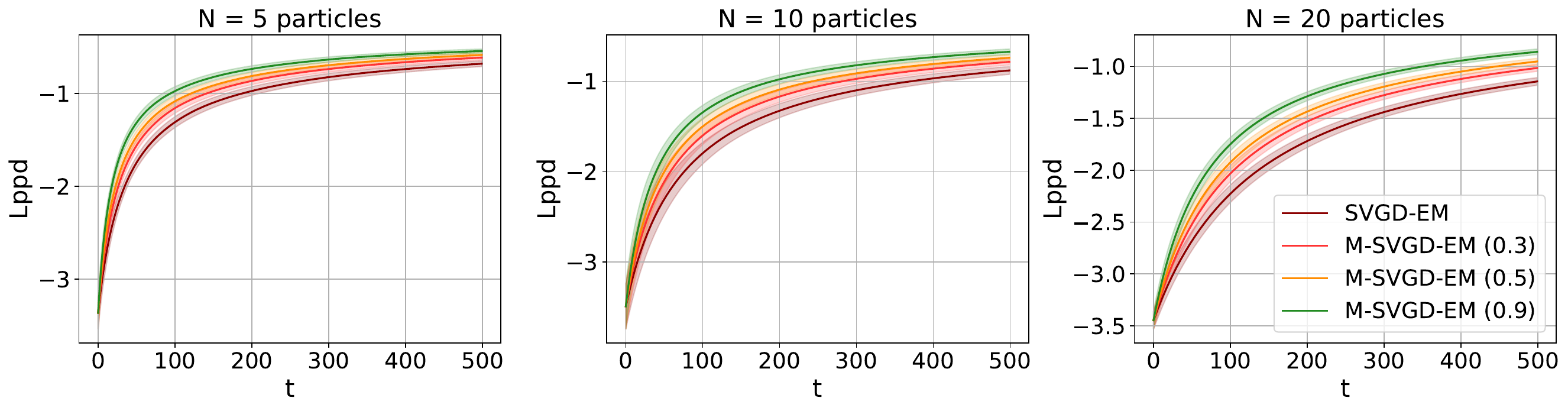}
    \caption{Log Predictive Probability Density using $N={5,10,20}$ with initial $\alpha=\beta=0.0$.}
    \label{fig:bnn_5_10_20_lppd_medianrbf_0}
\end{figure}

\begin{figure}[H]
\centering
    \includegraphics[width=\linewidth, height=4.0cm]{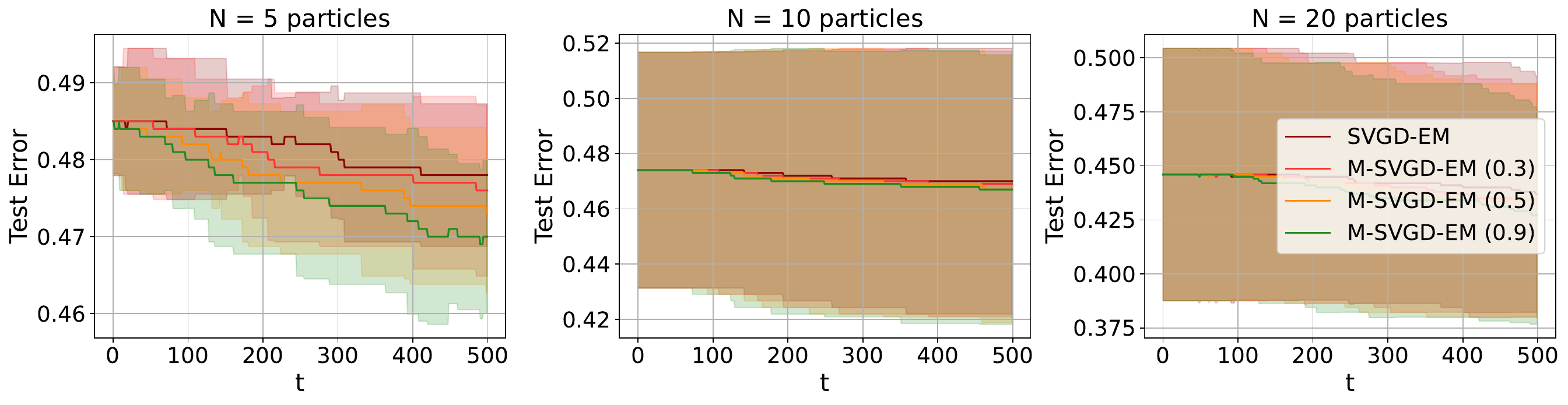}
    \caption{Test Error using $N={5,10,20}$ with initial $\alpha=\beta=2.0$.}
    \label{fig:bnn_5_10_20_test_medianrbf_2}
\end{figure}

\begin{figure}[h!]
\centering
    \includegraphics[width=\linewidth, height=4.0cm]{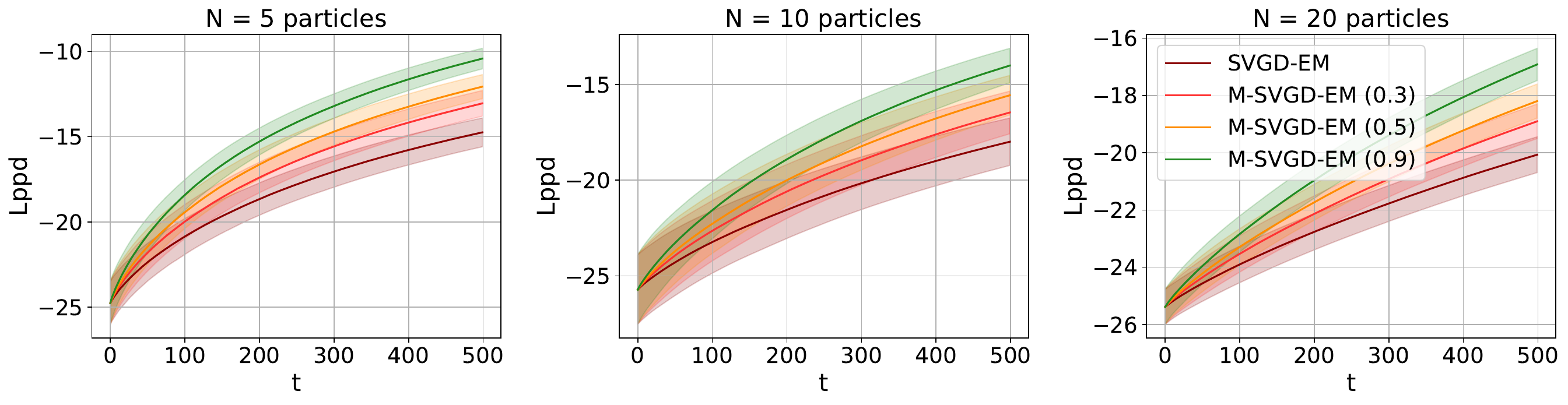}
    \caption{Log Predictive Probability Density using $N={5,10,20}$ with initial $\alpha=\beta=2.0$.}
    \label{fig:bnn_5_10_20_lppd_medianrbf_2}
\end{figure}

We use in this section a modification of our method, provided in Algorithm~\ref{alg:adagrad}.
\begin{algorithm}
\caption{M-SVGD-EM with AdaGrad}
\label{alg:adagrad}
\begin{algorithmic}[1]
\STATE {\bfseries Input:} momentum constants $\alpha_X, \alpha_\theta \in \mathbb{R}$, initial particles $\{x_0^{(i)}\}_{i=1}^{N} \sim q_0$, initial parameter $\theta_0$, joint log-likelihood $\ell$, kernel $k$, learning rate $\eta$, stability constant $\epsilon$.
\STATE {\bfseries Initialize:} $\theta_0 = \tilde{\theta}_0$, $\{x_0^{(i)}\}_{i=1}^{N} = \{\tilde{x}_0^{(i)}\}_{i=1}^{N}$, $G_\theta^{(0)} = 0$, $G_x^{(i,0)} = 0$ for $i=1,\ldots,N$
\FOR{$t = 0,1,\dots, T-1$}
    \STATE \textcolor{red}{Update $\theta$ with AdaGrad}
    \STATE $g_\theta^{(t)} = \frac{1}{N}\sum_{j=1}^{N} \nabla_{\theta} \ell(\tilde{\theta}_t, x_{t}^{(j)})$
    \STATE $G_\theta^{(t)} = G_\theta^{(t-1)} + (g_\theta^{(t)})^2$
    \STATE $\theta_{t+1} = \tilde{\theta}_t + \eta \cdot \frac{g_\theta^{(t)}}{\sqrt{G_\theta^{(t)} + \epsilon}}$
    \STATE $\tilde{\theta}_{t+1} = \theta_{t+1} + \alpha_\theta(\theta_{t+1} - \theta_{t})$
    \STATE \textcolor{red}{Update particles with AdaGrad}
    \FOR{$i = 1,2,\dots, N$}
        \STATE $g_x^{(i,t)} = \frac{1}{N}\sum_{j=1}^{N} k(\tilde{x}_{t}^{(j)}, \tilde{x}_{t}^{(i)}) \nabla_{x} \ell(\theta_{t+1}, \tilde{x}_{t}^{(j)})$
        \STATE \hspace{1.44cm} $+ \nabla_{1}k(\tilde{x}_{t}^{(j)}, \tilde{x}_{t}^{(i)})$
        \STATE $G_x^{(i,t)} = G_x^{(i,t-1)} + (g_x^{(i,t)})^2$
        \STATE $x_{t+1}^{(i)} = \tilde{x}_{t}^{(i)} + \eta \cdot \frac{g_x^{(i,t)}}{\sqrt{G_x^{(i,t)}} + \epsilon}$
        \STATE $\tilde{x}_{t+1}^{(i)} = x_{t+1}^{(i)} + \alpha_X (x_{t+1}^{(i)} - x_{t}^{(i)})$
    \ENDFOR
\ENDFOR
\end{algorithmic}
\end{algorithm}
\\
Below we present the results using a MedianRBF kernel on the BNN Example. Figures~\ref{fig:bnn_test_appendix_medianrbf}--\ref{fig:bnn_lppd_appendix_medianrbf} display the test error and log-predictive-probability-density w.r.t $t$ for the initialization $\alpha=\beta=0.0$ using $N=10$. The performance is comparable to that obtained with the AutoRBF kernel (Figures~\ref{fig:bnn_test_0_paper}--\ref{fig:bnn_lppd_0_paper}), where \gls*{M-SVGD-EM} outperforms \gls*{SVGD-EM} across both kernels. We present the same experiment varying the number of particles $N=5,10,20$ for $\alpha=\beta=0.0$ and $\alpha=\beta=2.0$ (see Figures~\ref{fig:bnn_5_10_20_test_medianrbf_0}--\ref{fig:bnn_5_10_20_lppd_medianrbf_2}), it is clear that increasing the number of particles with MedianRBF does not offer a better performance especially when the initializations are poor. In Figures~\ref{fig:bnn_5_10_20_test_medianrbf_2}--\ref{fig:bnn_5_10_20_lppd_medianrbf_2} for $N=10$ results are heavily affected by poor initialization showing that the model is as good as a coin-flip, whereas in AutoRBF Figures~\ref{fig:bnn_test_2_paper}--\ref{fig:bnn_lppd_2_paper}, even though there still is a decline in the overall performance due to the poor initialization, AutoRBF is more robust providing a lower test error and a immediate decreasing on its trend.

\section{Additional results on Toy Hierarchical Model}
\begin{figure}[h]
    \centering
    \begin{subfigure}[t]{0.47\textwidth}
        \centering
        \includegraphics[width=\textwidth]{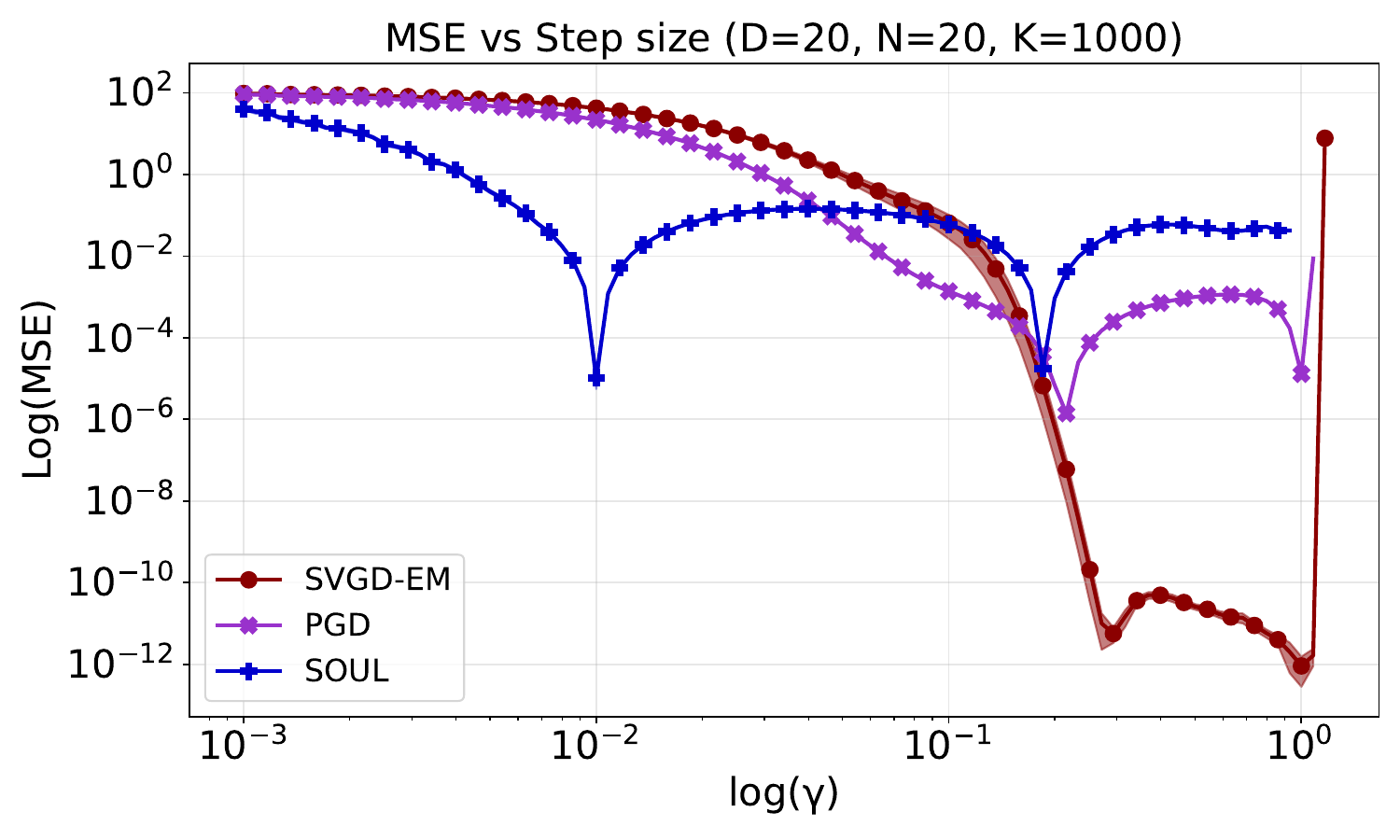}
        \caption{Log-Log Plot MSE vs log($\gamma$)}
        \label{fig:toy_lr_appendix_a}
    \end{subfigure}
    \hfill
    \begin{subfigure}[t]{0.47\textwidth}
        \centering
        \includegraphics[width=\textwidth]{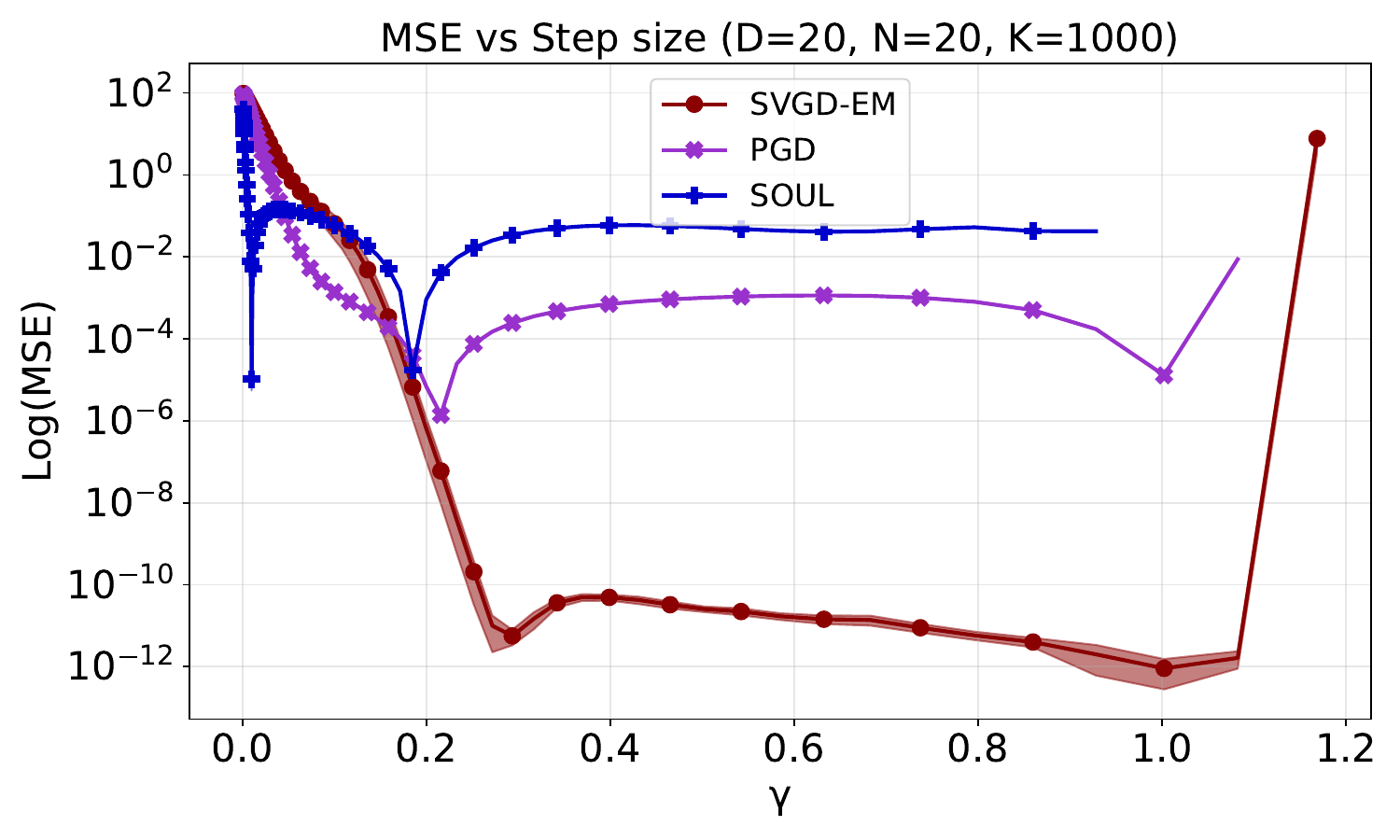}
        \caption{Log(MSE) vs $\gamma$}
        \label{fig:toy_lr_appendix_b}
    \end{subfigure}
    \caption{MSE w.r.t Learning Rate over 10 independent trials.}
    \label{fig:toy_lr_appendix}
\end{figure}

Figures~\ref{fig:toy_lr_appendix_a}--\ref{fig:toy_lr_appendix_b} show the selection of the optimal learning rates with respect to the lowest MSE for the results in section~\ref{sec:toy}. Each method is run for a fixed number of iterations $T=1000$ and averaged over 5 independent trials.

\subsection{Accelerations in \texorpdfstring{$\Theta$ vs. in $\Pac$}{}}
\begin{figure}[h!]
\centering
\begin{subfigure}{0.32\textwidth}
    \includegraphics[width=\linewidth, height=3.5cm]{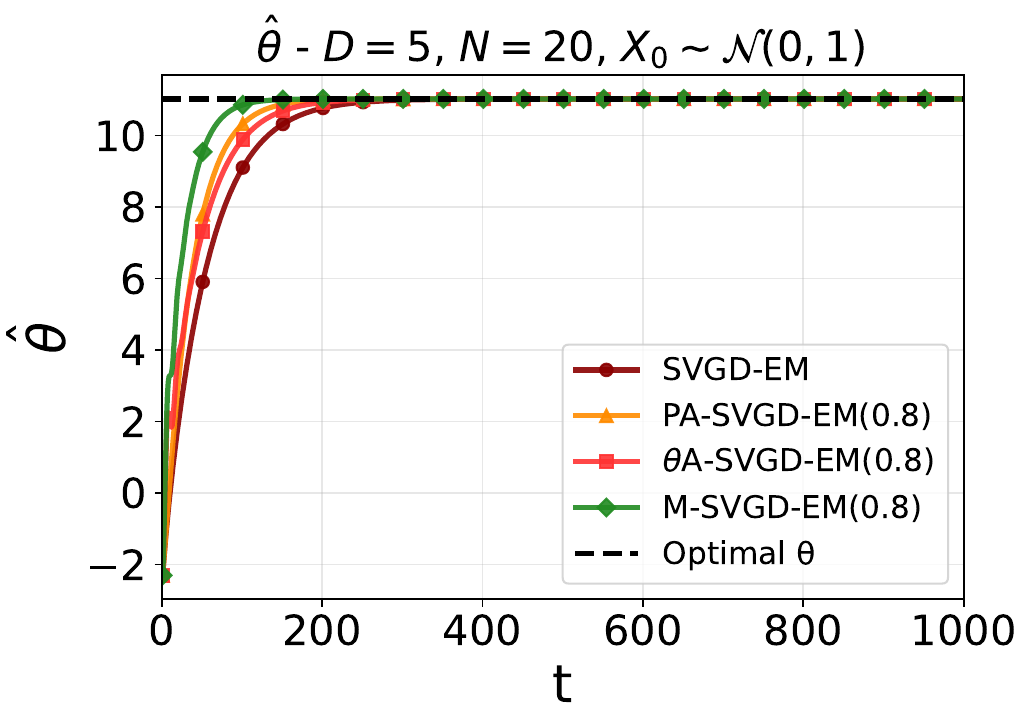}
    \caption{$D=5$}
    \label{fig:Toy_Extra_D5}
\end{subfigure}
\hfill
\begin{subfigure}{0.32\textwidth}
    \includegraphics[width=\linewidth, height=3.5cm]{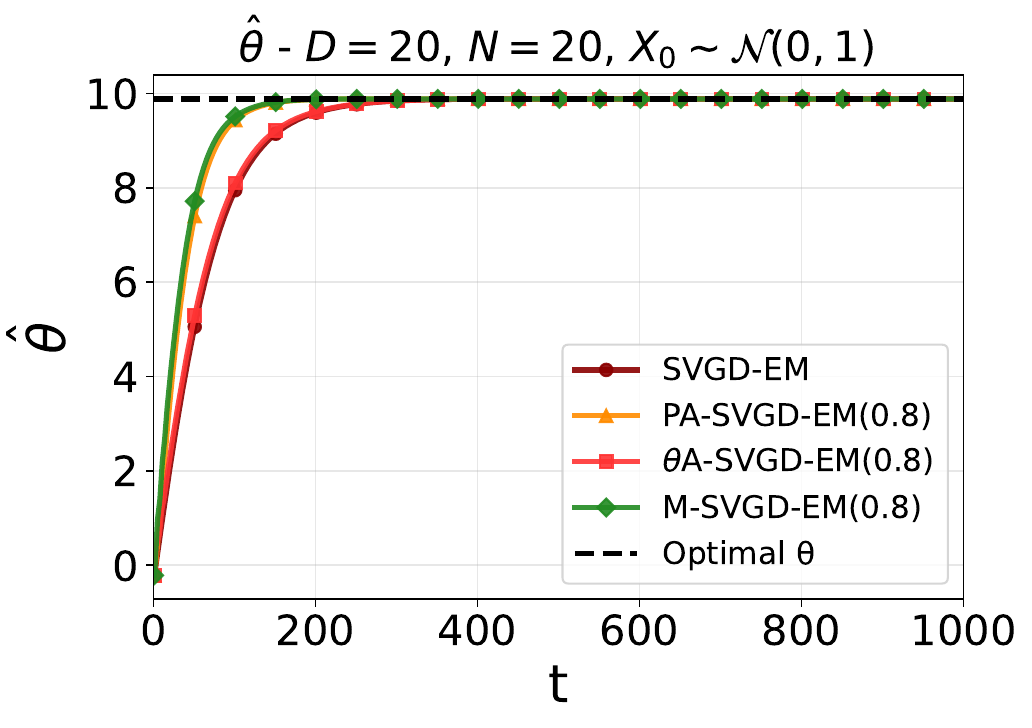}
    \caption{$D=20$}
    \label{fig:Toy_Extra_D20}
\end{subfigure}
\hfill
\begin{subfigure}{0.32\textwidth}
    \includegraphics[width=\linewidth, height=3.5cm]{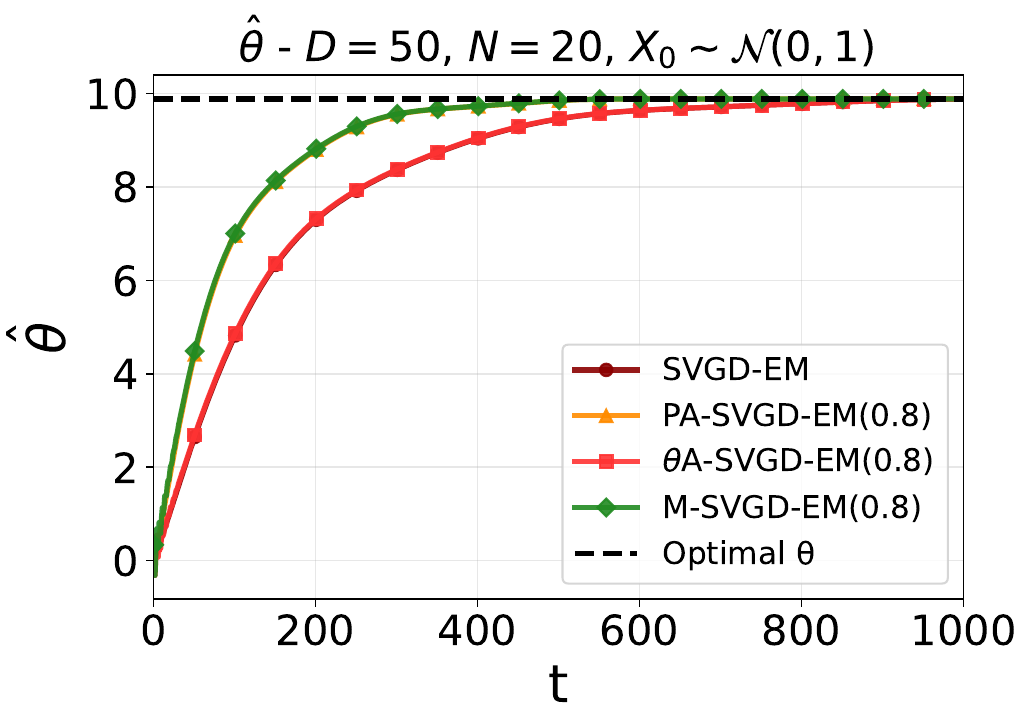}
    \caption{$D=50$}
    \label{fig:Toy_Extra_D50}
\end{subfigure}
\caption{Parameter estimation of a single run varying dimensions $D=5,20,50$.}
\label{fig:toy_para_appendix}
\end{figure}

\begin{figure}[h!]
\centering
\begin{subfigure}{0.33\textwidth}
    \includegraphics[width=\linewidth, height=3.5cm]{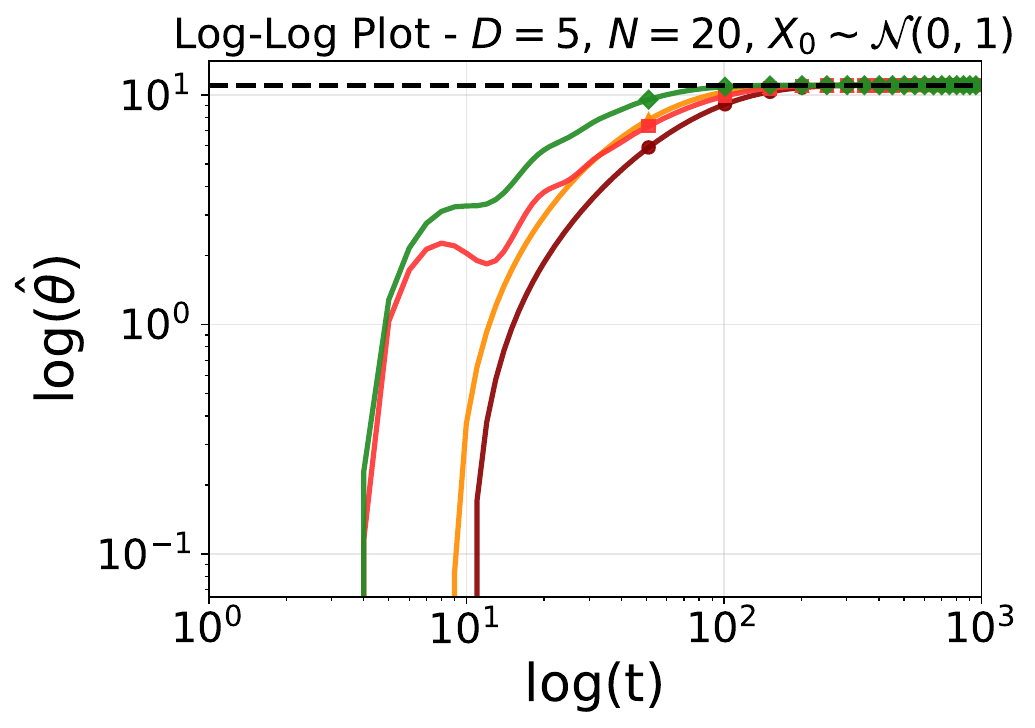}
    \caption{$D=5$}
    \label{fig:Toy_Extra_D5_log}
\end{subfigure}
\hfill
\begin{subfigure}{0.32\textwidth}
    \includegraphics[width=\linewidth, height=3.5cm]{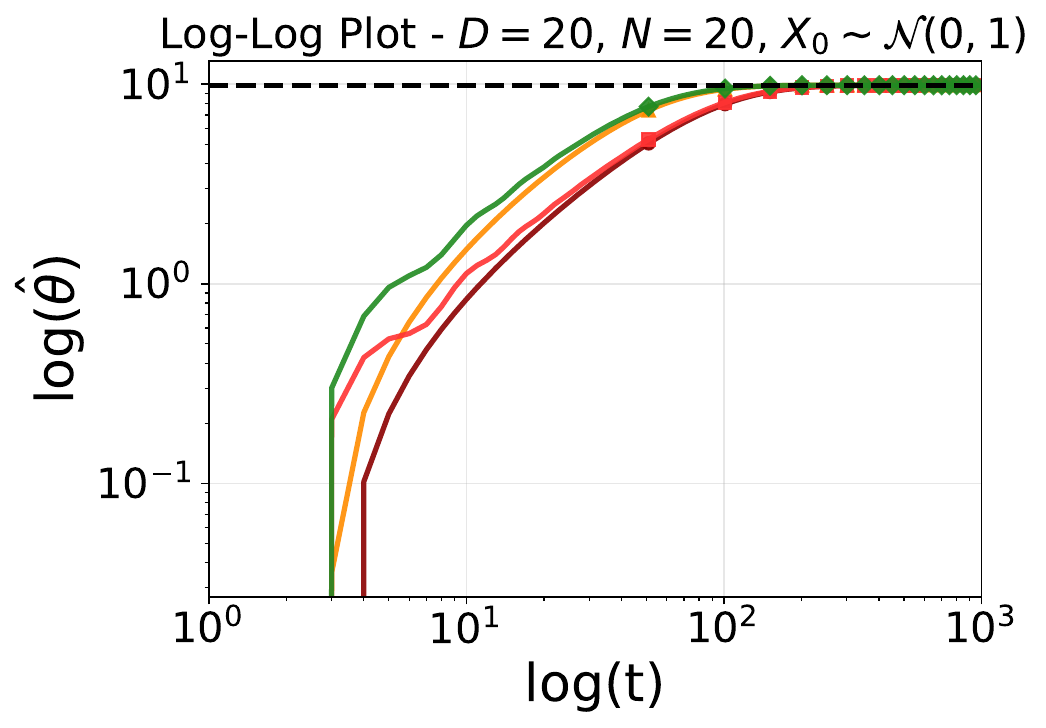}
    \caption{$D=20$}
    \label{fig:Toy_Extra_D20_log}
\end{subfigure}
\hfill
\begin{subfigure}{0.33\textwidth}
    \includegraphics[width=\linewidth, height=3.5cm]{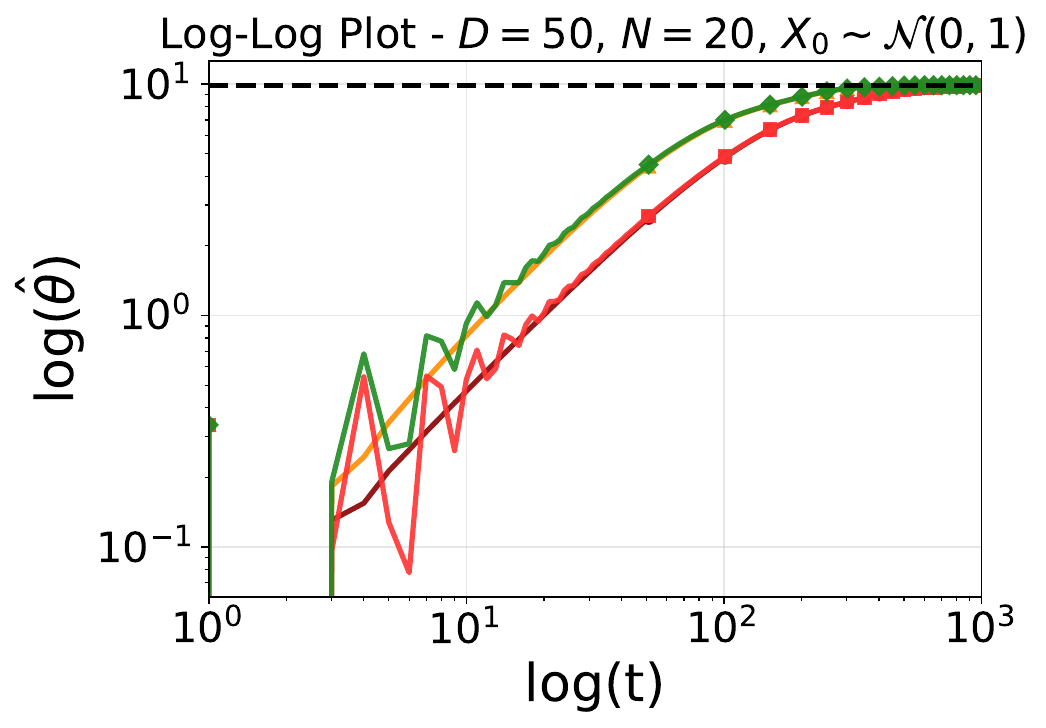}
    \caption{$D=50$}
    \label{fig:Toy_Extra_D50_log}
\end{subfigure}
\caption{Log-Log Parameter estimation of a single run varying dimensions $D=5,20,50$.}
\label{fig:loglog_toy_extra}
\end{figure}

\begin{figure}[h!]
\centering
\begin{subfigure}{0.32\textwidth}
    \includegraphics[width=\linewidth, height=3.5cm]{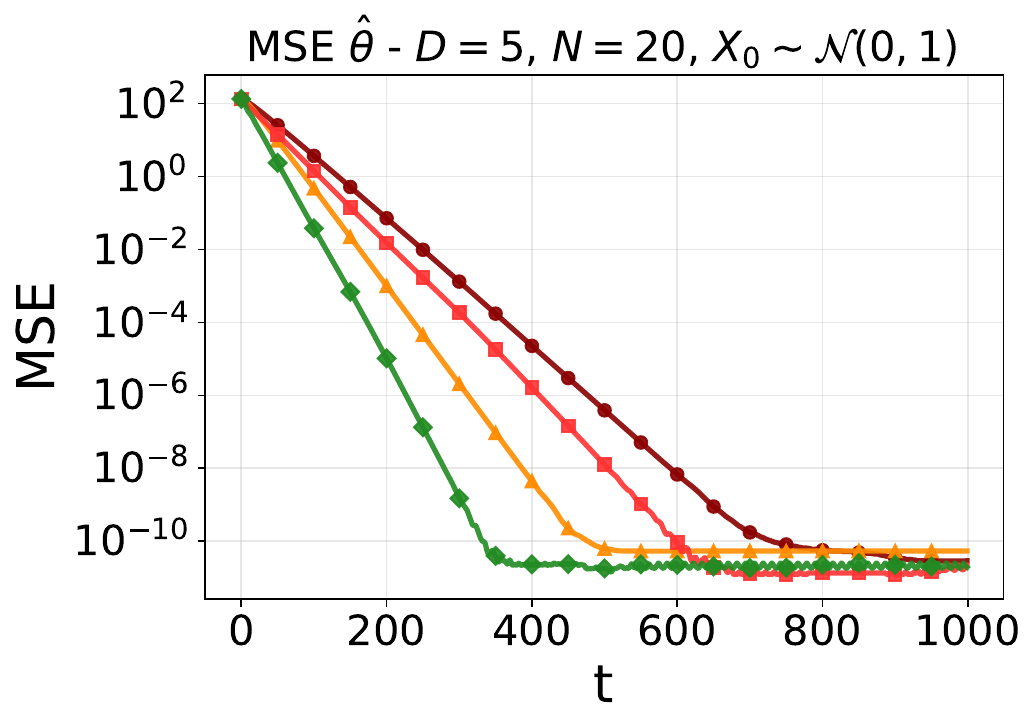}
    \caption{$D=5$}
    \label{fig:MSE_plots_toy_extra_5}
\end{subfigure}
\hfill
\begin{subfigure}{0.32\textwidth}
    \includegraphics[width=\linewidth, height=3.5cm]{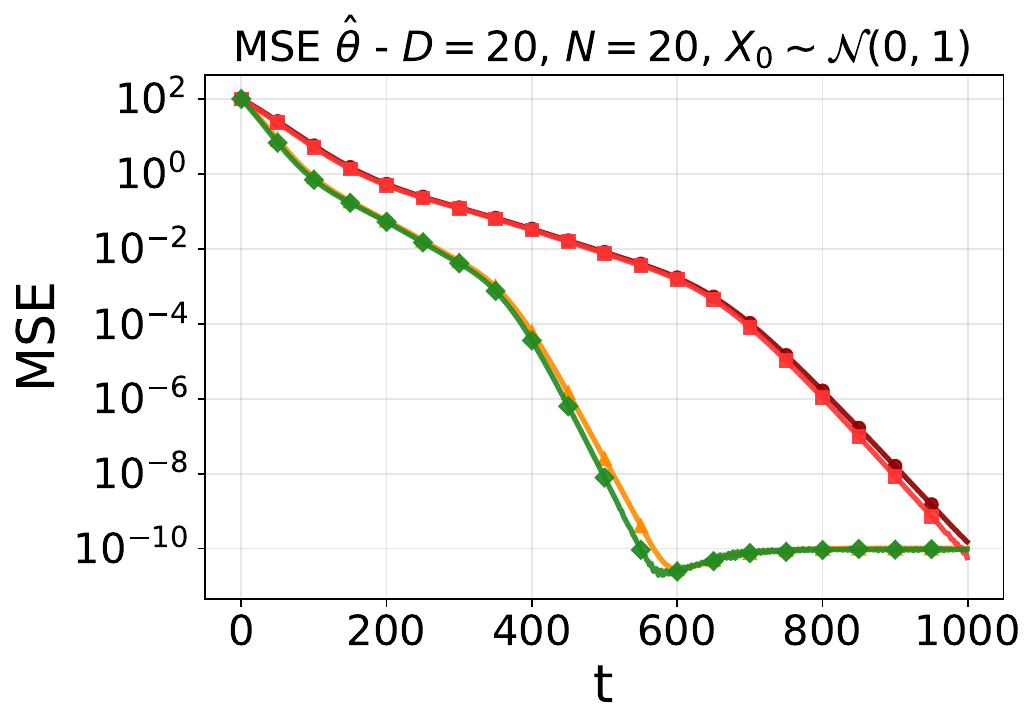}
    \caption{$D=20$}
    \label{fig:MSE_plots_toy_extra_20}
\end{subfigure}
\hfill
\begin{subfigure}{0.32\textwidth}
    \includegraphics[width=\linewidth, height=3.5cm]{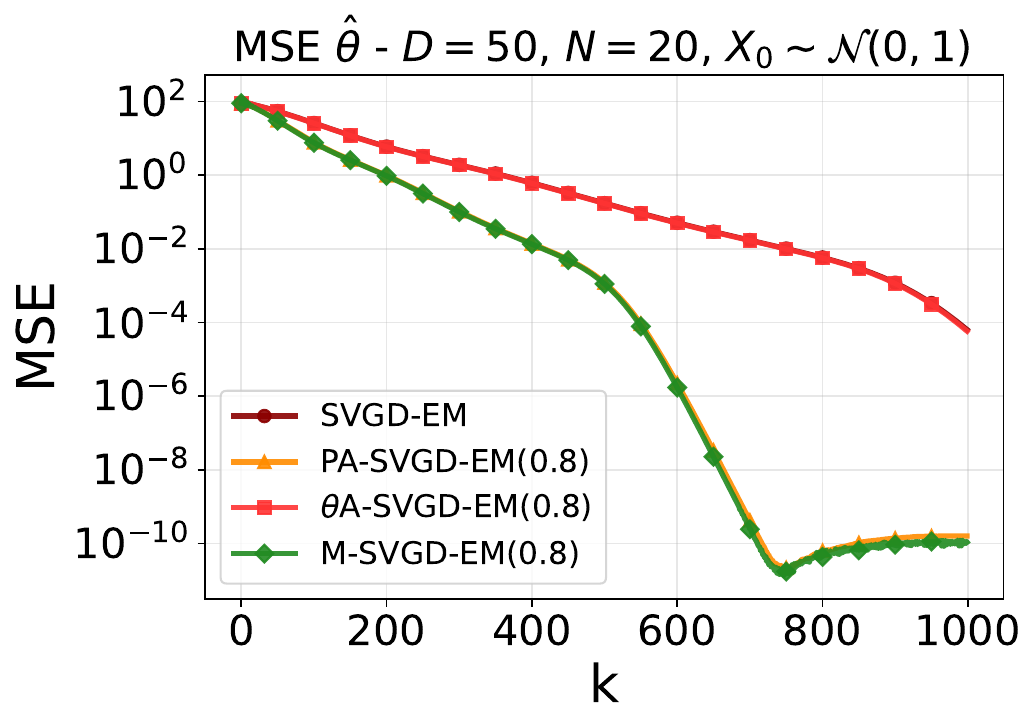}
    \caption{$D=50$}
    \label{fig:MSE_plots_toy_extra_50}
\end{subfigure}
\caption{MSE vs iterations calculated w.r.t the empirical minimizer $D=5,20,50$.}
\label{fig:MSE_plots_toy_extra}
\end{figure}

\begin{figure}[h]
\centering
\begin{subfigure}{0.32\textwidth}
    \includegraphics[width=\linewidth, height=3.5cm]{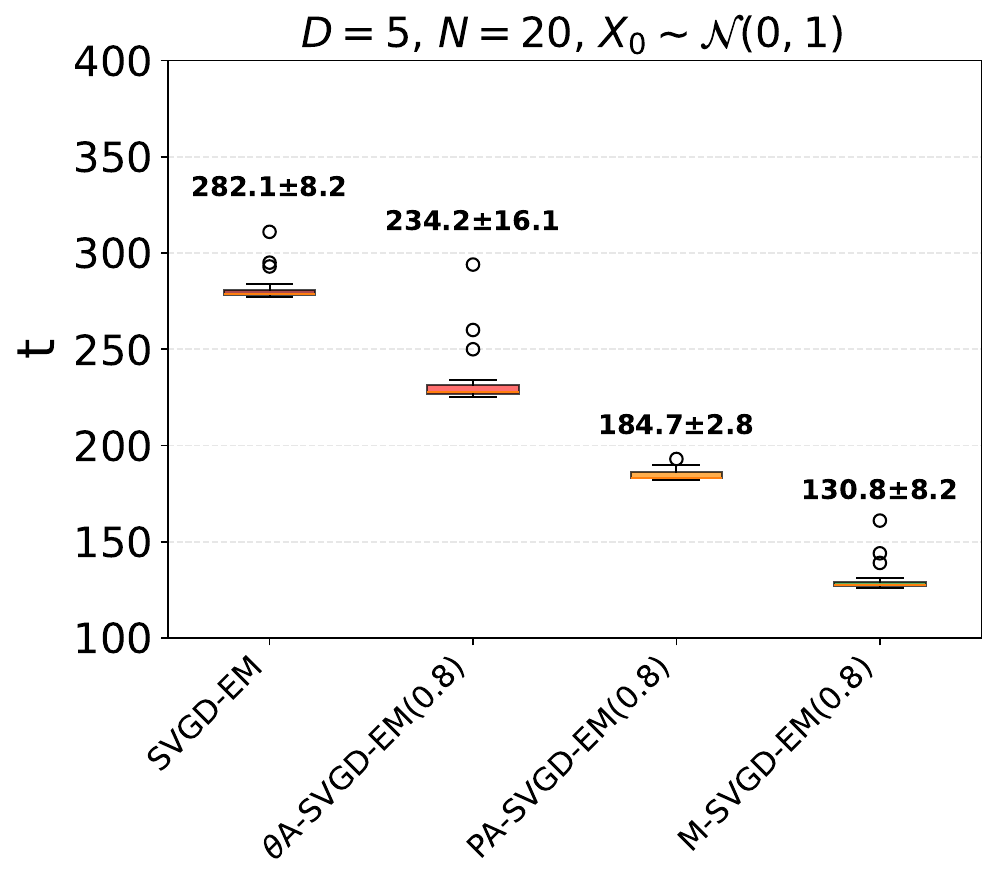}
    \caption{$D=5$}
    \label{fig:toy_boxplot_appendix_5}
\end{subfigure}
\hfill
\begin{subfigure}{0.32\textwidth}
    \includegraphics[width=\linewidth, height=3.5cm]{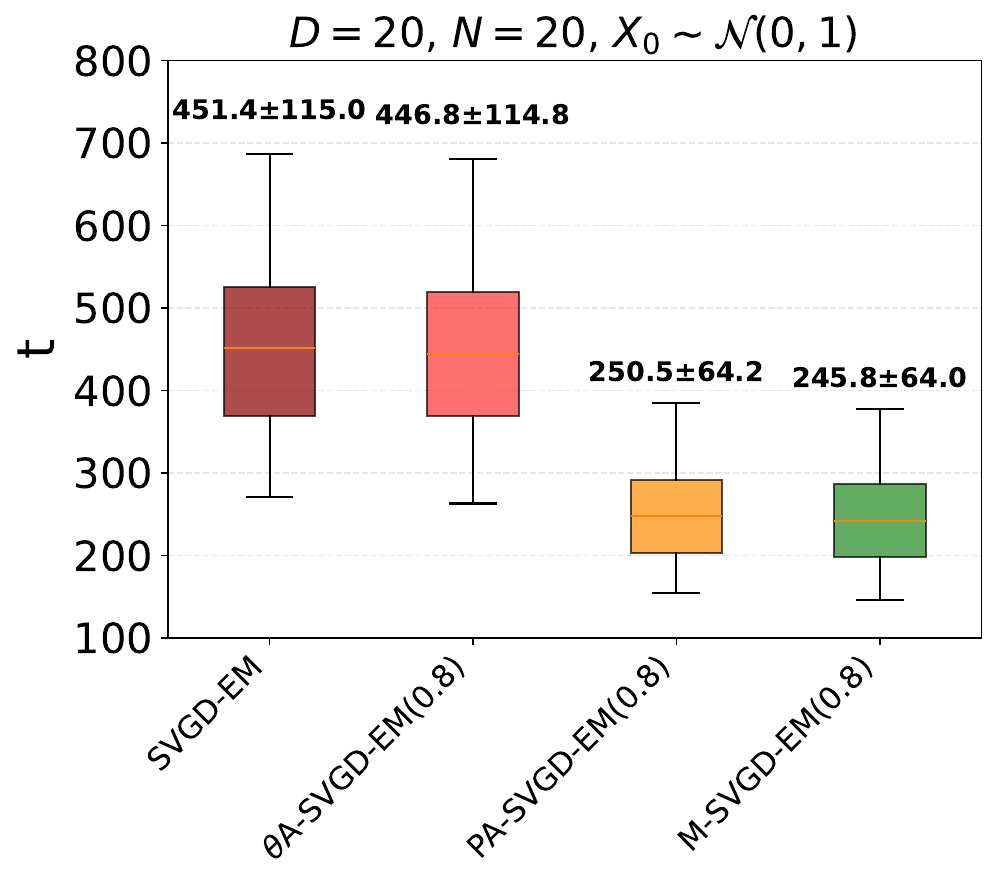}
    \caption{$D=20$}
    \label{fig:toy_boxplot_appendix_20}
\end{subfigure}
\hfill
\begin{subfigure}{0.32\textwidth}
    \includegraphics[width=\linewidth, height=3.5cm]{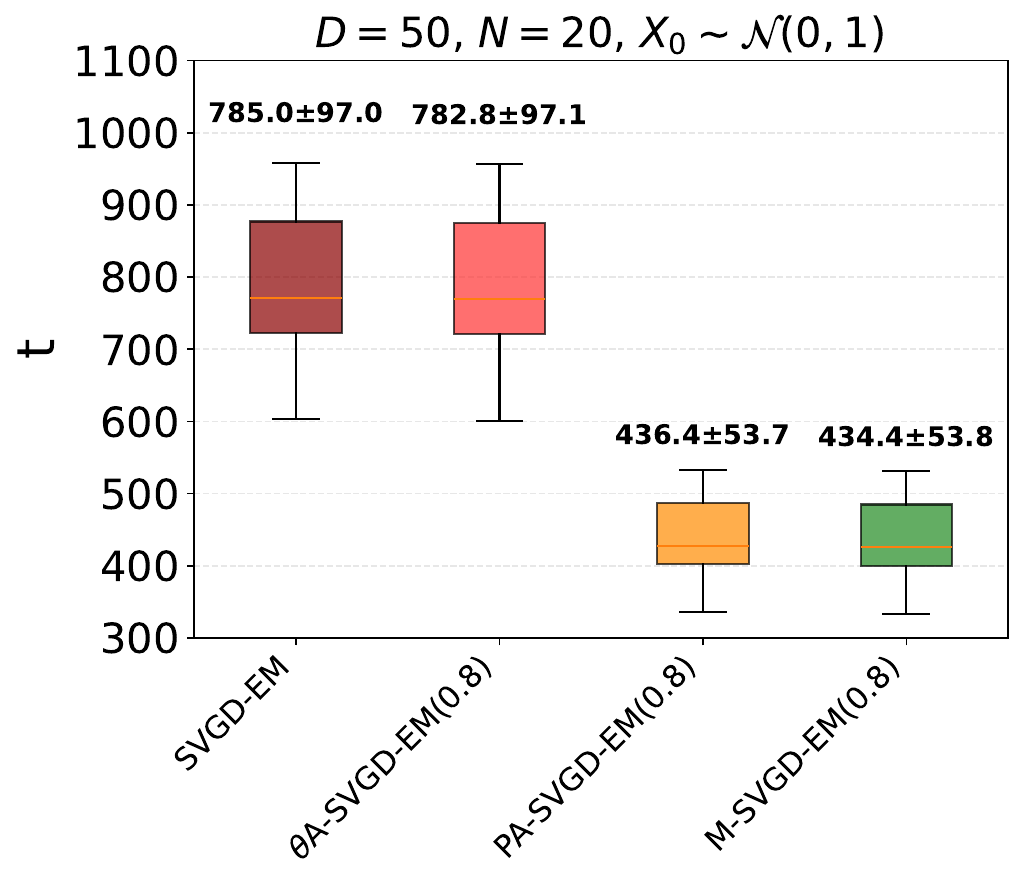}
    \caption{$D=50$}
\label{fig:toy_boxplot_appendix_50}
\end{subfigure}
\caption{Average number of iterations $\pm$se for 20 independent trials $D=5,20,50$.}
\label{fig:toy_boxplot_appendix}
\end{figure}

Here we present the results on the Toy Hierarchical model when accelerating only one part of \gls*{SVGD-EM}, namely the acceleration on the particles ($PA$) or the acceleration on the parameter ($\theta A$). We vary the dimensionality of the Toy Hierarchical Model with $D=5,20,50$ to better assess and understand the effect of different acceleration mechanisms. In Figures~\ref{fig:toy_para_appendix}--\ref{fig:loglog_toy_extra}, we display the parameter estimation against $t$ iterations on linear and loglog scales respectively. Figure~\ref{fig:MSE_plots_toy_extra} shows the MSE with respect to the empirical minimizer for each method and finally in Figure~\ref{fig:toy_boxplot_appendix} we present the average iterations to converge with a threshold of $0.05$ for 20 independent runs. All figures show that as we increase the dimensions of the problem, $\theta A$ offers no improvement to \gls*{SVGD-EM} whereas $PA$ is the only one responsible for accelerating the method.

\clearpage

\section{Additional results on Bayesian Logistic Regression}\label{appendix:Blr}
We present here how we choose the optimal learning rate $\gamma$ within the optimal range that minimizes the test error for different number of particles $N=10,20,100$.
\begin{figure}[h!]
\centering
\begin{subfigure}{0.32\textwidth}
    \includegraphics[width=\linewidth, height=3.5cm]{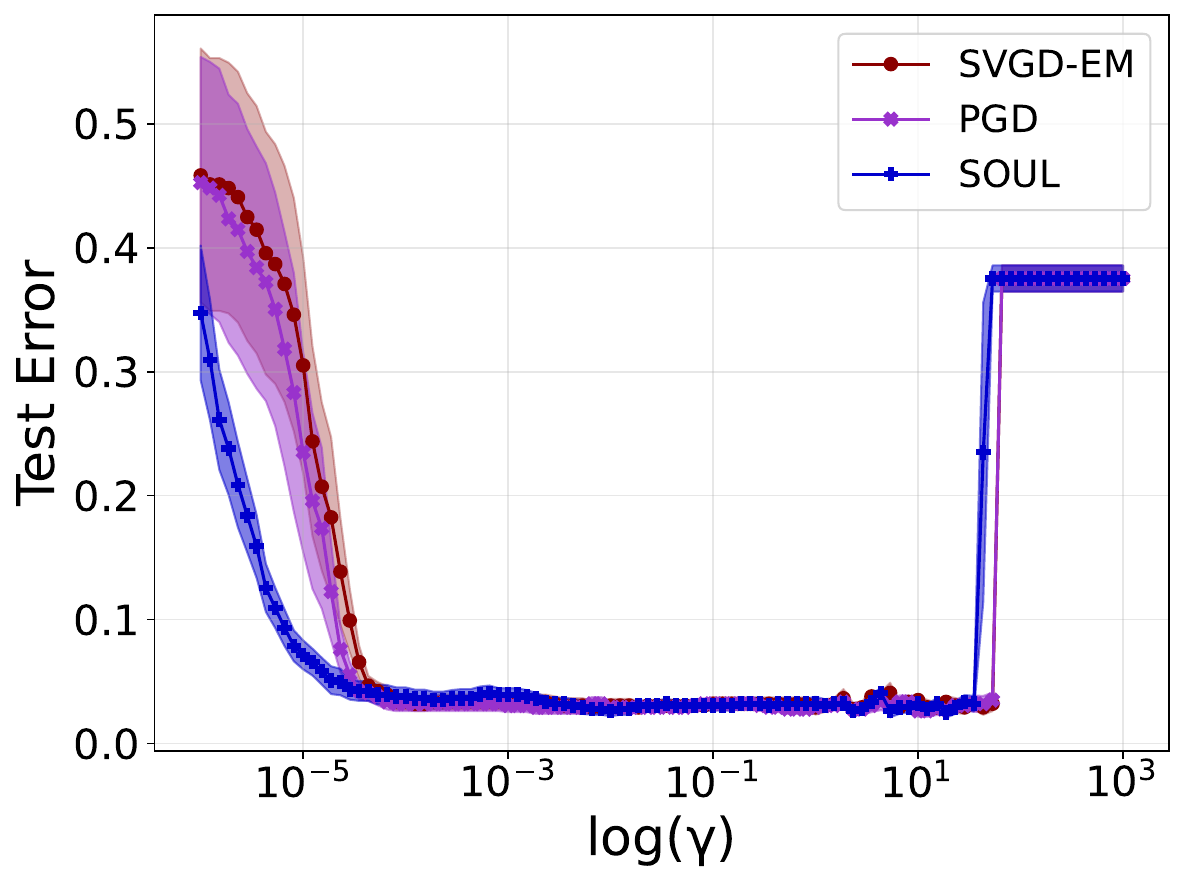}
    \caption{$N=10$}
    \label{fig:blr_test_N10_appendix}
\end{subfigure}
\hfill
\begin{subfigure}{0.32\textwidth}
    \includegraphics[width=\linewidth, height=3.5cm]{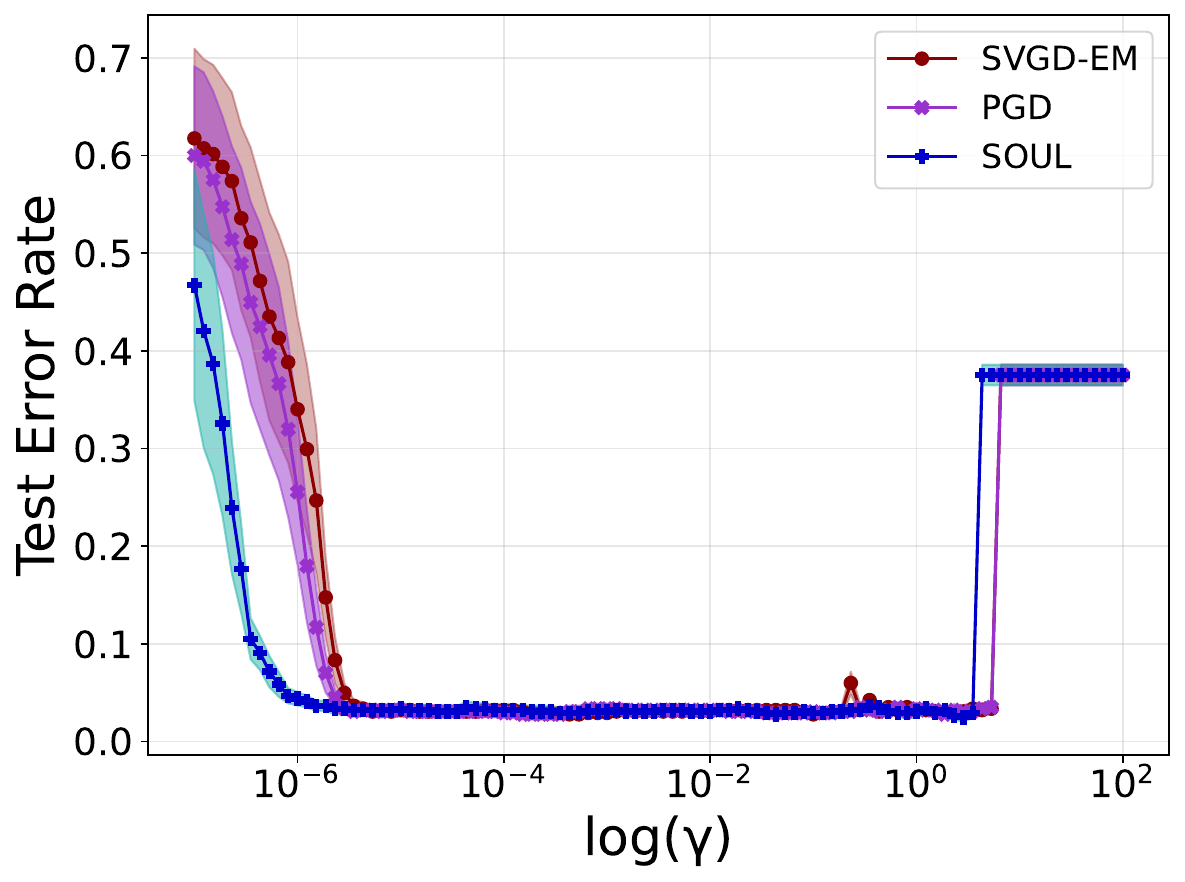}
    \caption{$N=20$}
    \label{fig:blr_test_N20_appendix}
\end{subfigure}
\hfill
\begin{subfigure}{0.32\textwidth}
    \includegraphics[width=\linewidth, height=3.5cm]{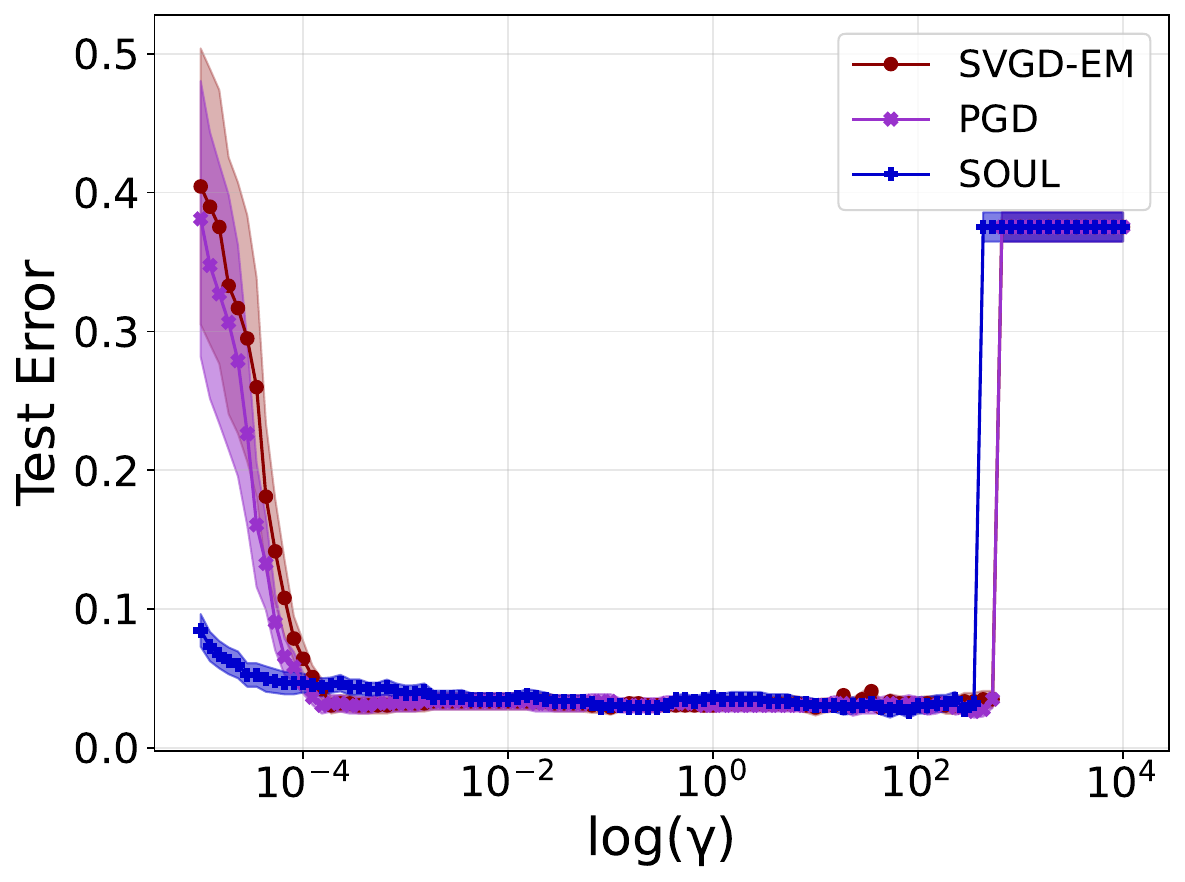}
    \caption{$N=100$}
    \label{fig:blr_test_N100_appendix}
\end{subfigure}
\caption{Test Error for Bayesian Logistic Regression with different numbers of particles.}
\label{fig:blr_test_N20_appendix111}
\end{figure}

\section{MPGD Tuning}\label{appendix:MPGD}
Following the approach found in the original implementation
\parencite{lim2023momentum}.
We do a grid search over 10 independent runs with the hyper-parameters $\varepsilon$ damping, and $\eta$ mass.

\parencite{lim2023momentum} We set the joint effect to be captured by momentum
$\mu = 1 - \varepsilon\gamma\eta$, which is the velocity kept per update.

We fix $(\mu,\varepsilon)$ and choosing a familiar $\gamma$ this sets
$
    \mu = e^{-\gamma\eta\varepsilon}
\rightarrow
\eta =\frac{1-\mu}{\gamma\varepsilon}
$

\begin{figure}[h!]
\centering
    \includegraphics[width=0.63\linewidth, height=4.5cm]{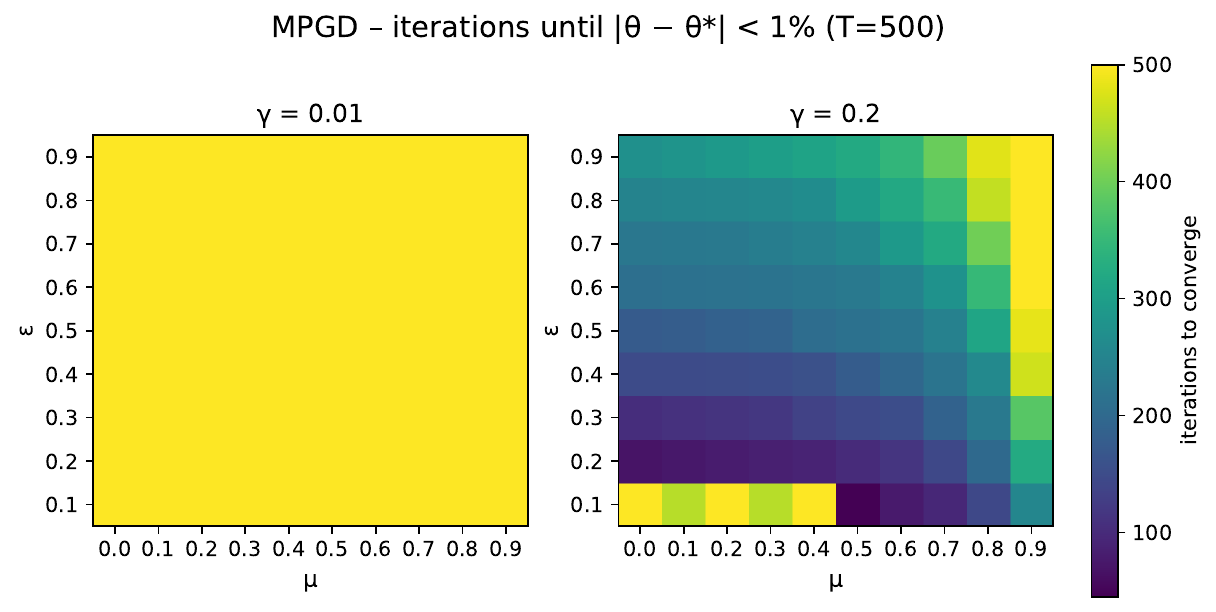}
    \caption{Heatmap Toy Hierarchical Model $\gamma$ vs $\mu$,with the darker color giving the least iterations needed until convergence}
    \label{fig:Heatmap_mpgd}

\end{figure}
\clearpage
 For the sake of consistency we fix the optimal learning rate found in \ref{fig:LR_TOY_1} for \gls*{PGD} $\gamma=(0.01,0.2)$ to provide a direct comparison against accelerated algorithm \gls*{MPGD}. In Figure \ref{fig:Heatmap_mpgd} displays a Heatmap for each learning rate $\gamma$ and the grid of   $
  \varepsilon \in \{0.1k : k = 1,\dots,9\}, 
  \mu \in \{0.1k : k = 0,\dots,9\},
$ the darker the color the less iterations required for \gls*{MPGD} to converge. In Figure \ref{fig:toy_paper} hyper-parameters used were $(\epsilon,\mu)=(0.1,0.2)$ or equivalently $(\epsilon,\eta)=(0.1,20)$.
 
 Similarly, in the BLR example Figure \ref{fig:blr_paper} we consider the optimal learning rate $\gamma=0.01$ found in Figure \ref{fig:blr_test_N20_appendix111} for \gls*{PGD} to provide a fair comparison against accelerated algorithm \gls*{MPGD}.

\begin{figure}[h!]
\centering
    \includegraphics[width=0.33\linewidth, height=4.5cm]{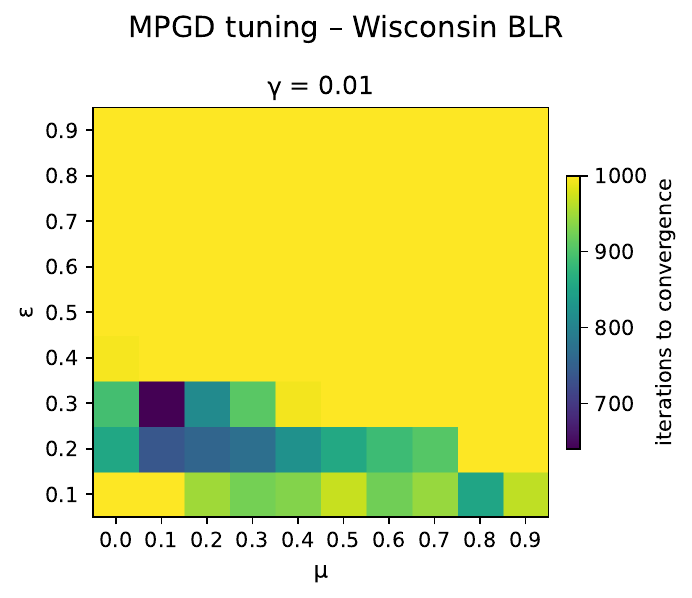}
    \caption{Heatmap Wisconsin Bayesian Logistic Regression Model
 $\gamma$ vs $\mu$,with the darker color giving the least iterations needed until convergence}
    \label{fig:Heatmap_mpgd_blr}

\end{figure}

In Figure \ref{fig:Heatmap_mpgd_blr} we present the hyper-parameters tuning where we chose the algorithm with the least iterations to converge over 10 independent trials and $T=1000$ steps, namely $(\epsilon,\mu)=(0.3,0.1)$ or equivalently $(\epsilon,\eta)=(0.3,300)$.

\section{Experimental Details}
We report the runtime required to obtain the main results presented in the paper, averaged over five independent runs for each experiment. All experiments were conducted on a machine with a 12th Gen Intel(R) Core(TM) i7-12700H CPU @ 2.30 GHz and 32 GB of RAM.

\begin{table}[h!]
\centering
\caption{Average Runtime for each model's notebook}
\label{tab:time_table}
\begin{tabular}{|l|l|}
\hline
\textbf{Notebook}    & \textbf{Time (s)} \\ \hline
Toy Hierarchical Model       & $497.09$           \\ \hline
Bayesian Logistic Regression & $423.35$           \\ \hline
Bayesian Neural Network      & $578.71$           \\ \hline
\end{tabular}
\end{table}

\clearpage

\end{document}